%% file: main.tex
\documentclass[10pt,twocolumn,letterpaper]{article}

\usepackage[pagenumbers]{cvpr} %

\input{preamble}

\definecolor{cvprblue}{rgb}{0.21,0.49,0.74}
\usepackage[pagebackref,breaklinks,colorlinks,allcolors=cvprblue]{hyperref}

\def\confName{CVPR}
\def\confYear{2027}

\title{Semantic Forcing: Unfolding Semantic Evolution\\
for Streaming Interactive Video Generation}

\author{%
\fontsize{11}{13}\selectfont Xu He$^{1,*,\dagger}$ \enspace
Runqi Wang$^{2,\dagger}$ \enspace
Hejia Chen$^{2}$ \enspace
Jiaming Zhang$^{3}$ \enspace
Haoxian Zhang$^{2,\ddagger}$ \enspace
Songlin Tang$^{2}$ \\
\fontsize{11}{13}\selectfont Changyuan Zheng$^{1}$ \,%
Yifan Xu$^{3}$ \,%
Minghong Cai$^{4}$ \,%
Liyang Chen$^{1}$ \,%
Yiran Qiao$^{5}$ \,%
Pengfei Wan$^{2}$ \,%
Zhiyong Wu$^{1,4,\textrm{\Letter}}$ \\[1mm]
{\normalsize $^{1}$Tsinghua University \enspace
$^{2}$KlingAI Research \enspace
$^{3}$Nanjing University \enspace
$^{4}$The Chinese University of Hong Kong \enspace
$^{5}$UCAS} \\[1mm]
{\normalsize $^{\dagger}$Equal contribution \enspace
$^{\ddagger}$Project lead \enspace
$^{\textrm{\Letter}}$Corresponding author} \\[1mm]
{\ttfamily\normalsize hexu18@mails.tsinghua.edu.cn \enspace zhanghaoxian@kuaishou.com \enspace zywu@sz.tsinghua.edu.cn} \\
{\normalsize\url{https://hjrphoebus.github.io/Semantic-Forcing/}}
}

\begin{document}

\twocolumn[{
\maketitle
\begin{center}
    \vspace{-.28in}
    \captionsetup{type=figure}
    \includegraphics[width=\linewidth]{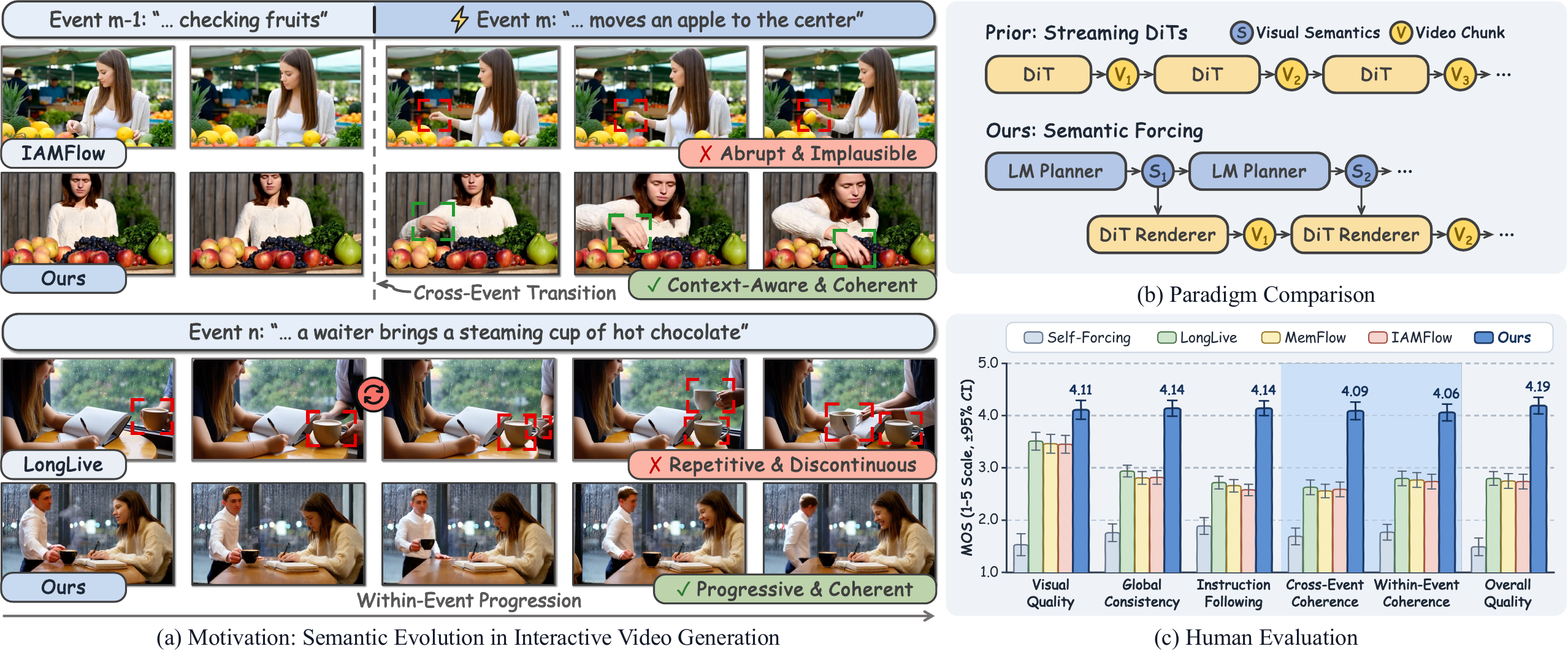}
    \vspace{-.28in}
    \captionof{figure}{
    \textbf{\methodname{}: Toward coherent semantic evolution in streaming interactive video generation.}
    (a) Existing methods suffer from abrupt cross-event transitions and repetitive within-event behaviors, whereas our method yields coherent semantic evolution.
    (b) Our framework couples online autoregressive semantic planning with streaming DiT rendering in a parallel pipeline.
    (c) Human evaluation reports MOS with 95\% confidence intervals, highlighting improvements
    in perceptual quality and narrative coherence.
    }
    \label{fig:teaser}
    \vspace{-.3em}
\end{center}
}]

\begingroup
\renewcommand{\thefootnote}{*}
\footnotetext[1]{\footnotesize This work was conducted during the author's internship at KlingAI.}
\endgroup

\input{sec/0_abstract}

\input{sec/1_body}

{
    \small
    \bibliographystyle{ieeenat_fullname}
    \bibliography{main}
}

\input{sec/X_suppl}

\end{document}

%% file: preamble.tex
\usepackage{microtype}

\usepackage{xspace}

\usepackage{multirow}
\usepackage{marvosym} %

\input{new_colors}

\input{math_commands.tex}

\newcommand{\methodname}{Semantic Forcing\xspace}
\newcommand{\datasetname}{EventVid\xspace}
\newcommand{\myparagraph}[1]{\vspace{.15em}\noindent\textbf{#1}}

\makeatletter
\DeclareRobustCommand{\supp}{%
  \@ifnextchar.%
    {\textit{Supp.\ Mat}}%
    {\textit{Supp.\ Mat.}\xspace}%
}
\makeatother

\definecolor{plannerblue}{HTML}{1B4F72}   %
\definecolor{rendereramber}{HTML}{9A6300} %
\definecolor{cotpurple}{HTML}{5B2C6F}     %
\definecolor{membrown}{HTML}{7E4009}      %

\usepackage{colortbl}
\definecolor{GroupGray}{RGB}{240,242,245}
\definecolor{OursBlue}{RGB}{237,244,251}

\usepackage[normalem]{ulem}

\usepackage{etoolbox}
\patchcmd{\abstract}
  {\vspace*{12pt}}
  {\vspace*{3pt}}
  {}
  {\PackageError{abstract-spacing}{Patch failed}{}}

\usepackage[most]{tcolorbox}

\usepackage{algorithm}
\usepackage{algpseudocode}

\usepackage{float}

\usepackage{titletoc}

\definecolor{calloutbg}{RGB}{232, 242, 254}
\definecolor{calloutborder}{RGB}{145, 188, 235}
\definecolor{callouttitle}{RGB}{18, 62, 130}

\usepackage{placeins}

%% file: new_colors.tex
\definecolor{ao}{rgb}{0.0, 0.0, 1.0}
\definecolor{airforceblue}{rgb}{0.36, 0.54, 0.66}
\definecolor{ceruleanblue}{rgb}{0.16, 0.32, 0.75}
\definecolor{cerulean}{rgb}{0.0, 0.48, 0.65}
\definecolor{celestialblue}{rgb}{0.29, 0.59, 0.82}
\definecolor{azure(colorwheel)}{rgb}{0.0, 0.5, 1.0}
\definecolor{babyblue}{rgb}{0.54, 0.81, 0.94}
\definecolor{babyblueeyes}{rgb}{0.63, 0.79, 0.95}
\definecolor{ballblue}{rgb}{0.13, 0.67, 0.8}

\definecolor{asparagus}{rgb}{0.53, 0.66, 0.42}
\definecolor{ao(english)}{rgb}{0.0, 0.5, 0.0}
\definecolor{applegreen}{rgb}{0.55, 0.71, 0.0}
\definecolor{armygreen}{rgb}{0.29, 0.33, 0.13}
\definecolor{gray-asparagus}{rgb}{0.27, 0.35, 0.27}
\definecolor{green(ryb)}{rgb}{0.4, 0.69, 0.2}

\definecolor{amethyst}{rgb}{0.6, 0.4, 0.8}
\definecolor{antiquefuchsia}{rgb}{0.57, 0.36, 0.51}
\definecolor{blue-violet}{rgb}{0.54, 0.17, 0.89}
\definecolor{brightlavender}{rgb}{0.75, 0.58, 0.89}
\definecolor{brightube}{rgb}{0.82, 0.62, 0.91}
\definecolor{brilliantlavender}{rgb}{0.96, 0.73, 1.0}

\definecolor{amber}{rgb}{1.0, 0.75, 0.0}
\definecolor{amber(sae/ece)}{rgb}{1.0, 0.49, 0.0}
\definecolor{atomictangerine}{rgb}{1.0, 0.6, 0.4}
\definecolor{burntorange}{rgb}{0.8, 0.33, 0.0}
\definecolor{burntsienna}{rgb}{0.91, 0.45, 0.32}
\definecolor{cadmiumorange}{rgb}{0.93, 0.53, 0.18}
\definecolor{carrotorange}{rgb}{0.93, 0.57, 0.13}
\definecolor{chocolate(web)}{rgb}{0.82, 0.41, 0.12}
\definecolor{chromeyellow}{rgb}{1.0, 0.65, 0.0}
\definecolor{darkorange}{rgb}{1.0, 0.55, 0.0}
\definecolor{darktangerine}{rgb}{1.0, 0.66, 0.07}
\definecolor{deepcarrotorange}{rgb}{0.91, 0.41, 0.17}
\definecolor{deepsaffron}{rgb}{1.0, 0.6, 0.2}
\definecolor{fulvous}{rgb}{0.86, 0.52, 0.0}

%% file: math_commands.tex
\usepackage{amsmath,amsfonts,bm}

\def\eqref#1{equation~\ref{#1}}

\def\1{\bm{1}}

\DeclareMathAlphabet{\mathsfit}{\encodingdefault}{\sfdefault}{m}{sl}
\SetMathAlphabet{\mathsfit}{bold}{\encodingdefault}{\sfdefault}{bx}{n}

%% file: sec/0_abstract.tex
\begin{abstract}
Streaming interactive video generation enables users to dynamically guide evolving video narratives through successive text instructions.
However, existing causal video diffusion models often struggle with implausible cross-event transitions and repetitive within-event behaviors, lacking explicit semantic reasoning and long-horizon planning.
We introduce \methodname{}, a streaming framework that integrates high-level semantic evolution modeling into causal video synthesis.
An autoregressive language model planner continuously predicts spatiotemporally aligned visual semantic tokens, with context-aware chain-of-thought reasoning to bridge incoming instructions with evolving visual context.
In parallel, a causal diffusion transformer renderer synthesizes video chunks from streaming semantics, leveraging semantic-guided memory retrieval for long-range consistency.
To support training, we construct \datasetname{}, an event-centric dataset comprising 2.5M single-event and 245K multi-event videos with fine-grained annotations.
Experiments demonstrate superior semantic coherence, instruction following, and visual quality over existing streaming approaches.
\end{abstract}

%% file: sec/1_body.tex
\section{Introduction}
\label{sec:intro}

Diffusion transformers (DiTs)~\cite{peebles2023dit} have remarkably elevated video generation in visual fidelity and spatiotemporal consistency~\cite{brooks2024sora,kling2024,wan2025wan,seedance2026,minimaxh3}, opening up new possibilities for creative and cinematic applications~\cite{wu2026effectlearner,yuan2026edit,fang20263d,chen2026human,cai2026videocanvas,huang2026unityvideo,he2026from,bai2025recammaster,wang2026subjectanchor,team2026kling,wang2026disentangle,liu2026mozoo,chen2025cafe}.
Interactive long video generation consequently emerges as a compelling direction, allowing users to dynamically guide continuous multi-event narratives through streaming text prompts~\cite{villegas2023phenaki}.
Recent advances in autoregressive (AR) video generation~\cite{chen2024diffusion,teng2025magi,causvid,chen2025skyreels,huang2026self,cf,cfplus} offer a promising approach toward this goal: causal DiTs are employed to synthesize videos chunk-by-chunk conditioned on previously generated content and provided text prompts.
To accommodate prompt switching, common practice updates historical context at switch boundaries via KV-recache~\cite{longlive}, allowing the model to respond to newly provided instructions while preserving local visual smoothness across event transitions.
Subsequent works further explore incorporating memory mechanisms~\cite{ji2025memflow,iamflow,zhang2026tinyhistory} to enhance long-range content consistency.
Collectively, these advances successfully extend DiT-based video generation to streaming interactive synthesis for multi-event narratives.

However, these approaches primarily focus on maintaining local visual continuity and long-range appearance consistency, often overlooking high-level semantic coherence in evolving interactive narratives.
Fig.~\ref{fig:teaser}(a) illustrates two typical failures: 1) \textit{incoherent cross-event transition}, where new actions or objects arise without respecting the established visual context, leading to abrupt or implausible event shifts;
2) \textit{incoherent within-event progression}, where actions are repeatedly or implausibly executed rather than naturally progressing toward completion.
We argue that these limitations arise because causal video diffusion models tend to respond to text instructions locally within individual chunks, without either explicitly reasoning about how new textual guidance interacts with the evolving visual context, or planning future semantic evolution across extended temporal horizons.

To bridge this gap, we advocate introducing explicit semantic evolution modeling into causal video synthesis.
Specifically, we leverage autoregressive language models (LMs)~\cite{llm} to reason and plan in a visual semantic space~\cite{tatok},
generating spatially grounded semantic tokens temporally aligned with the evolving video chunks.
This formulation offers two key properties:
1) \textit{in-context reasoning}, where newly arriving instructions can be seamlessly interleaved with historical visual semantics, facilitating reasoning about plausible cross-event transitions; 
and 2) \textit{sequential planning}, 
where future visual semantics are predicted step-by-step along the fully unfolding trajectory, driving coherent within-event progression toward completion.

To this end, we propose \textbf{\methodname}, an interactive video generation framework combining autoregressive semantic reasoning and planning with causal video synthesis for coherent multi-event narratives.
At its core, \methodname orchestrates an autoregressive \textit{LM planner} and a causal video \textit{DiT renderer} in a parallel streaming pipeline:
the LM planner progressively predicts spatiotemporal visual semantic tokens~\cite{tatok} within an evolving context, 
while the causal DiT renderer incrementally synthesizes video chunks as the corresponding semantic tokens become available.
To enable explicit reasoning upon prompt switching, we design a dedicated \textit{context-aware chain-of-thought (CoT)} mechanism~\cite{wei2022chain} within the planner: when a new instruction arrives, the planner jointly reasons over the instruction and historical visual semantics to infer structured event changes, specifying how the new event should evolve from the current state.
These inferred cues remain in the planner's context to guide subsequent semantic prediction, enabling plausible cross-event transitions and coherent within-event progression.

Conditioned on these streaming semantic tokens for spatiotemporal dynamics alongside textual descriptions for detailed visual appearance, the causal DiT renderer synthesizes high-fidelity video chunks on the fly.
To further maintain visual consistency over extended durations, we introduce \textit{semantic-guided memory retrieval} for the renderer, which leverages the planned semantics of each upcoming video chunk to adaptively retrieve relevant historical visual context for rendering.
By coupling streaming semantic planning with memory-augmented causal rendering, \methodname generates interactive video streams with both high visual fidelity and coherent semantic evolution.

To support the training of our framework, we construct \textbf{\datasetname}, a large-scale event-centric video dataset featuring fine-grained event segmentation and annotations, comprising 2.5M single-event and 245K multi-event videos.
Beyond concise natural-language instructions and self-contained structured captions for each event, we further provide context-dependent event-change annotations, capturing how each event evolves from preceding visual context and providing explicit supervision for semantic reasoning and planning.

To summarize, our main contributions are as follows:
\begin{itemize}
    \item We propose \textbf{\methodname}, a streaming interactive video generation framework coupling an autoregressive LM planner with a causal DiT renderer to achieve coherent narrative evolution alongside high-fidelity synthesis.
    \item We design \textit{context-aware CoT} for explicit cross-event transition reasoning and long-horizon semantic planning, paired with \textit{semantic-guided memory retrieval} to preserve long-range visual consistency.
    \item We construct \textbf{\datasetname{}}, a large-scale event-centric dataset with fine-grained event-level annotations for multi-event interactive video generation.
    \item Extensive experiments show that \methodname{} significantly outperforms existing methods in both narrative semantic coherence and overall generation quality.
\end{itemize}

\begin{figure*}[t]
\centering
\includegraphics[width=.95\textwidth]{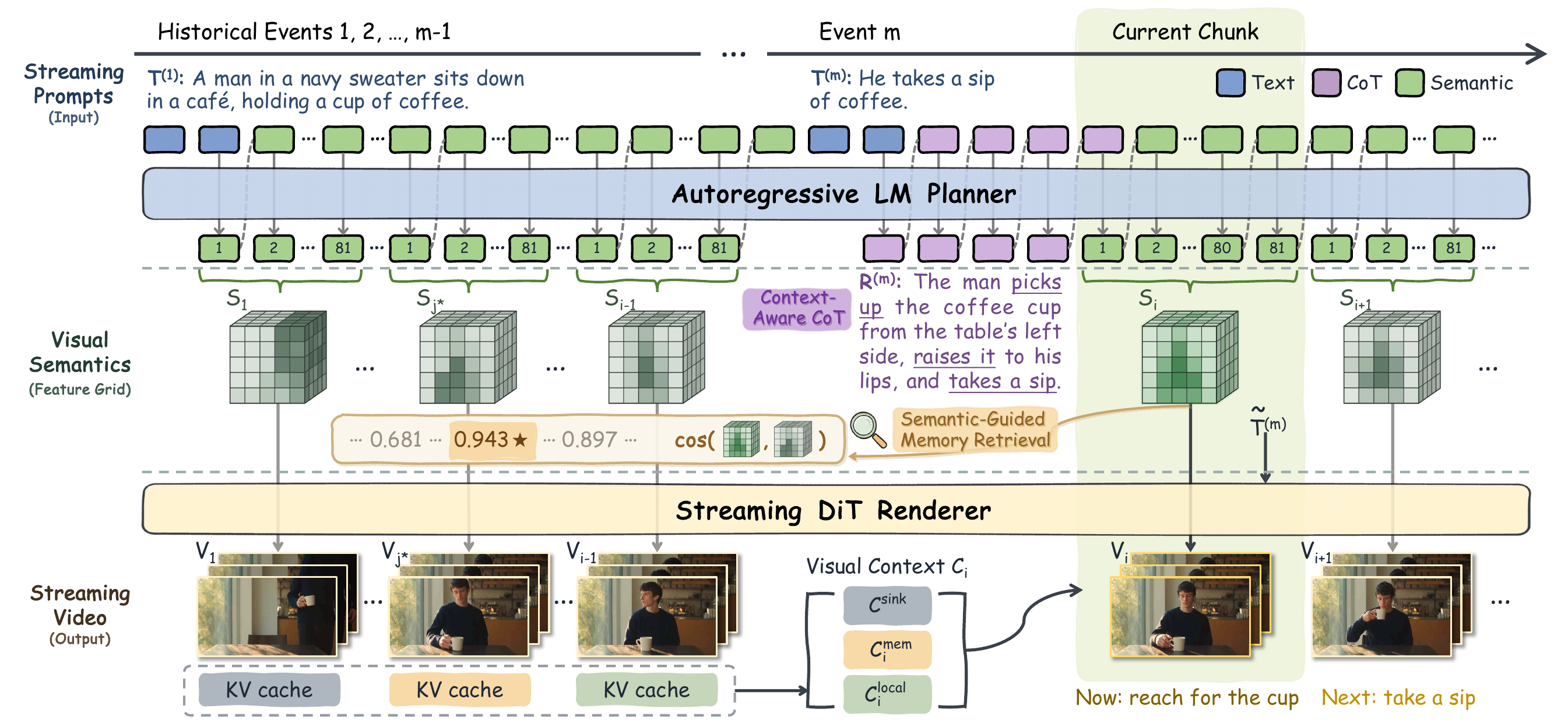}
\vspace{-.3em}
\caption{\textbf{The overall framework of \methodname{}} for streaming interactive video generation.
Our framework coordinates an \textcolor{plannerblue}{\textbf{autoregressive LM planner}} for online semantic planning and a \textcolor{rendereramber}{\textbf{streaming DiT renderer}} for real-time high-fidelity video synthesis.
Upon receiving a new instruction $T^{(m)}$, the planner employs \textcolor{cotpurple}{\textbf{context-aware CoT}} over preceding visual semantics to predict reasoning tokens $R^{(m)}$ describing event transitions and evolution, and sequentially outputs spatiotemporal visual semantic tokens $(S_i, S_{i+1}, \dots)$ to guide event progression.
In parallel, the renderer synthesizes video chunk $V_i$ from $S_i$ and visual context $C_i = [C^{\mathrm{sink}}, C_i^{\mathrm{mem}}, C_i^{\mathrm{local}}]$ with a tailored rewritten prompt $\widetilde{T}^{(m)}$ supplementing visual details, while leveraging $S_i$ via semantic-guided memory retrieval to recall the most relevant historical KV cache as visual memory $C_i^{\mathrm{mem}}$ for long-range consistency.}
\label{fig:pipeline}
\vspace{-1em}
\end{figure*}

\section{Related Work}
\label{sec:related_work}

\myparagraph{Autoregressive Video Generation.}
While bidirectional video diffusion transformers~\cite{peebles2023dit,brooks2024sora,wan2025wan,kling2024,seedance2026,minimaxh3} jointly denoise full sequences, autoregressive video models predict future frames from preceding context~\cite{cf,cfplus,zhuang2026self,helios2026,zhang2026scores}.
Early attempts rely on teacher forcing over clean history~\cite{jin2024pyramidal,gao2024vidgpt,kondratyuk2024videopoet,deng2024autoregressive,chen2026mindflow}, whereas Diffusion Forcing~\cite{chen2024diffusion} introduces independent per-frame noise levels, inspiring large-scale AR frameworks~\cite{teng2025magi,chen2025skyreels}.
To enable low-latency synthesis, CausVid~\cite{causvid} distills bidirectional teachers into few-step causal generators~\cite{dmd}, and Self-Forcing~\cite{huang2026self} incorporates self-generated rollouts to bridge the train-test gap.
Subsequent works~\cite{selfforcingpp2025,rollingforcing2025,deepforcing2025,xue2026ring,xia2026adversarial,su2026sgf+,zhang2026connected,wang2026liveact,wang2026liveedit} improve rollout stability, long-horizon extrapolation, and controllability.

A key application of AR video models is interactive multi-event generation~\cite{villegas2023phenaki,cai2025ditctrl}.
LongLive~\cite{longlive} extends Self-Forcing to streaming prompt switching via KV-recache and long tuning, while follow-up works~\cite{ji2025memflow,iamflow,anchorforcing2026,zhang2026tinyhistory,sun2025worldplay,wu2026echo,cai2026focused,chen2026pyramid,wang2026weave,zhang2026memory} introduce memory mechanisms to mitigate appearance drift.
While these methods primarily focus on visual continuity and long-range consistency, our work integrates online semantic planning into causal video generation, supporting coherent narrative evolution alongside high-fidelity synthesis.

\myparagraph{Semantic Modeling for Video Generation.}
Recent studies leverage multimodal language models (MLLMs) to enhance semantic understanding and reasoning in diffusion models.
Common strategies employ MLLMs for offline prompt enhancement~\cite{seedance2026,yang2025cogvideox,wan2025wan} or script planning~\cite{lin2023videodirectorgpt,zhuang2024vlogger}, extract conditioning embeddings via learnable queries or hidden states~\cite{pan2025metaqueries,omnivideo2,univideo}, or unify autoregressive text decoding with diffusion~\cite{zhou2024transfusion,deng2025bagel,liang2025mot,xie2024showo}.
Another line of work models intermediate visual semantic spaces to ease pixel-level synthesis~\cite{ding2026beyond}, initially via auxiliary diffusion models~\cite{semanticgen2025,tokensgen2025}.
Recently, Plan-X~\cite{planx} and Bernini~\cite{bernini2026} adopt autoregressive or masked models~\cite{li2024mar} with textual reasoning to predict semantic representations for offline synthesis.
In contrast, we target streaming interactive generation, performing online reasoning and semantic planning over dynamic instructions and evolving context alongside real-time rendering.

\section{\methodname{}}
\label{sec:method}

Interactive video generation aims to sequentially synthesize temporally continuous video chunks $\mathcal{V} = (V_1, V_2, \dots, V_N)$ in response to an initial prompt $T^{(1)}$ and subsequent text instructions $(T^{(2)}, T^{(3)}, \dots, T^{(M)})$ dynamically introduced during runtime.
This task requires both coherent narrative evolution and long-range visual consistency.

As illustrated in Fig.~\ref{fig:pipeline}, our proposed \textbf{\methodname{}} addresses these challenges by decoupling high-level semantic planning from pixel-level visual synthesis, coordinating an autoregressive LM planner and a causal DiT renderer in a streaming pipeline:
the LM planner jointly models historical visual semantics, newly arriving instructions, and explicit event-change reasoning within a unified autoregressive context, continuously predicting spatiotemporal visual semantic tokens aligned with subsequent video chunks (\cref{subsec:planner});
in parallel, the causal DiT renderer synthesizes successive video chunks from streaming semantic tokens, leveraging semantically relevant historical visual context as memory to preserve long-range visual consistency (\cref{subsec:renderer}).
We begin by briefly reviewing autoregressive LMs, video DiTs, and causal video distillation in \cref{subsec:preliminary}.

\subsection{Preliminaries}
\label{subsec:preliminary}

\myparagraph{Autoregressive LMs.}
Transformer-based autoregressive LMs~\cite{llm,transformer} with causal attention model discrete token sequences via next-token prediction, factorizing the joint distribution as $p_{\theta}(y_{1:L}) = \prod_{l=1}^{L} p_{\theta}(y_l \mid y_{<l})$.
In this work, we build our semantic planner upon a pretrained LM~\cite{yang2024qwen25} by extending its vocabulary with discrete visual semantic tokens for unified text-visual autoregressive modeling.

\myparagraph{Video DiTs.}
Video DiTs typically follow the latent diffusion paradigm~\cite{rombach2022high}, employing a 3D variational autoencoder (VAE)~\cite{vae} for video compression and a DiT backbone~\cite{peebles2023dit} for denoising.
Each DiT block interleaves 3D bidirectional self-attention for spatiotemporal modeling, cross-attention for external conditioning, and feed-forward networks.

\myparagraph{Causal Adaptation and Distillation.}
To enable low-latency streaming synthesis, recent approaches~\cite{causvid,huang2026self} equip video DiTs with a chunk-wise causal attention mask, generating chunks sequentially with historical key-value (KV) caches.
A few-step causal generator is first initialized from ODE trajectories of the bidirectional teacher, and subsequently trained with distribution matching distillation (DMD)~\cite{dmd} on self-generated autoregressive rollouts~\cite{huang2026self} to mitigate exposure bias.
We follow this paradigm to build our real-time streaming renderer, further augmenting it with semantic-guided memory for long-horizon consistency.

\subsection{Autoregressive Semantic Planning}
\label{subsec:planner}

Leveraging the pretrained contextual reasoning and sequential planning capabilities of LMs, we develop an autoregressive semantic planner that jointly models streaming user instructions and evolving visual semantics within a unified context to guide subsequent video chunk synthesis.

In practice, following~\cite{planx,tatok}, we adopt a text-aligned visual tokenizer (TA-Tok) to encode the representative keyframe of each video chunk $V_i$ into $K=81$ spatially structured discrete semantic tokens, denoted as $S_i = (s_{i,1}, \dots, s_{i,K})$.
The TA-Tok codebook is aligned with the LM text embedding space, thus allowing visual semantic tokens to be seamlessly interleaved with textual tokens in a unified autoregressive sequence.
During interactive generation, the planner continuously predicts semantic tokens $(S_1,S_2,\ldots)$ along the unfolding visual semantic trajectory, while dynamically incorporating newly arriving instructions into the current context.
Formally, given the current instruction $T^{(m)}$ and the historical context $H_{<i}$, the planner predicts the semantic tokens of the $i$-th video chunk as:
\begin{equation}
    p_\theta(S_i \mid H_{<i}, T^{(m)})
    =\!\prod_{\ell=1}^{K}
    p_\theta(s_{i,\ell} \mid H_{<i}, T^{(m)}, s_{i,<\ell}),
    \label{eq:planner_ar}
\end{equation}
where $H_{<i}$ comprises preceding instructions and all previously generated visual semantic tokens.

By grounding future predictions in the sequentially unfolding semantic trajectory, the planner can guide each event through its natural progression toward completion, rather than repeatedly responding to the same instruction.

\myparagraph{Context-Aware CoT Reasoning.}
While direct sequential interleaving supports basic semantic continuation, concise user instructions often specify only target actions or new subjects, without detailing how the visual semantics evolve from the existing state.
To explicitly bridge newly arriving instructions with the existing visual semantic context, we introduce a \textit{context-aware chain-of-thought (CoT)} mechanism.
Upon receiving a new instruction $T^{(m)}$, the planner first predicts structured event-change descriptions $R^{(m)}$, specifying how subjects, actions, and scene states should evolve from the preceding visual context.
Concretely, at the event boundary prior to chunk $i$, we expand the autoregressive sequence from $[\ldots, S_{i-1}, T^{(m)}, S_i, \ldots]$ to $[\ldots, S_{i-1}, T^{(m)}, R^{(m)}, S_i, \ldots]$, having the planner itself generate explicit reasoning tokens before continuing semantic prediction.

Unlike common prompt enhancement, our context-aware CoT offers twofold benefits: 
predicting $R^{(m)}$ encourages the planner to explicitly comprehend and reason about the relationship between new instructions and the existing visual state, 
while retaining $R^{(m)}$ in the autoregressive context provides an explicit semantic bridge to guide subsequent token prediction toward coherent event transitions and progression.

\myparagraph{Training.}
We construct interleaved training sequences using user instructions, structured event-change descriptions, and visual semantic tokens encoded from ground-truth videos.
A unified next-token cross-entropy loss~\cite{llm} is applied to both the reasoning tokens $\{R^{(m)}\}$ and semantic tokens $\{S_i\}$, jointly supervising context-aware event reasoning and autoregressive semantic planning.

\begin{figure*}[t]
\centering
\includegraphics[width=.97\textwidth]{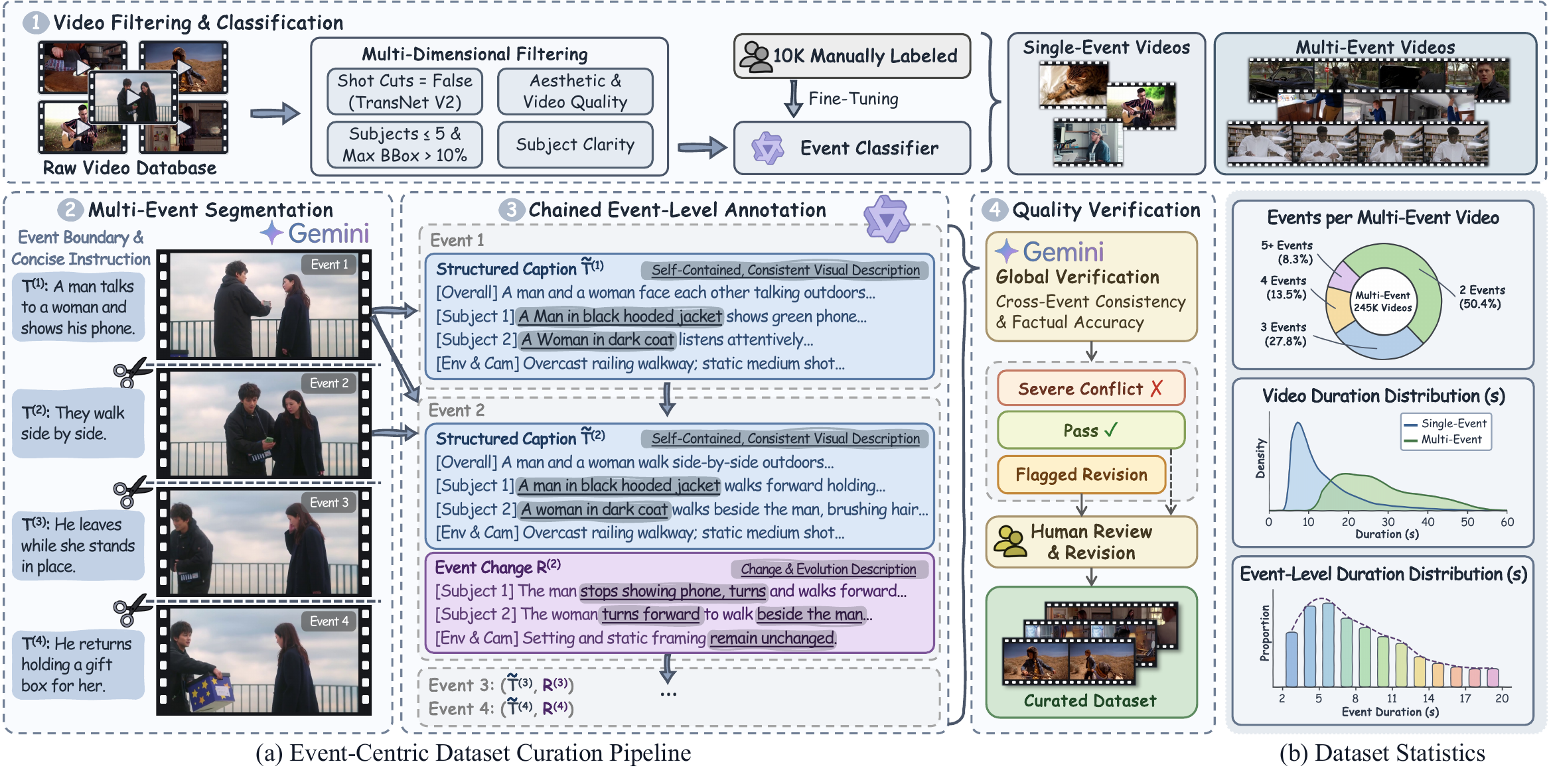}
\vspace{-0.7em}
\caption{\textbf{Overview of the \datasetname{} curation pipeline and dataset statistics.}
\textbf{(a)} We first collect and filter high-quality, cut-free videos into single- and multi-event clips. Multi-event videos are then partitioned into contiguous event segments with concise instructions, followed by chained annotation to produce self-contained structured captions and context-dependent event-change descriptions, with global automated verification and human revision ensuring final quality.
\textbf{(b)} Dataset statistics on event counts, video durations, and event durations.}
\label{fig:data_pipeline}
\vspace{-1em}
\end{figure*}

\subsection{Semantic-Guided Streaming Video Rendering}
\label{subsec:renderer}

While the LM planner captures high-level spatiotemporal dynamics and semantic structure, synthesizing videos with fine-grained visual details still relies on diffusion-based rendering.
Therefore, we employ a semantic-guided causal DiT renderer to progressively translate streaming visual semantics into continuous and consistent video chunks.
Given the semantic tokens $S_i$ of the upcoming video chunk $V_i$, the renderer generates $V_i$ conditioned on the corresponding textual description $\widetilde{T}^{(m)}$ and historical visual context $C_i$:
\begin{equation}
    V_i \sim p_\phi\big(V_i \mid S_i, \widetilde{T}^{(m)}, C_i\big).
    \label{eq:renderer}
\end{equation}
Within each DiT block, the spatially structured semantic tokens $S_i$ are projected and injected through cross-attention, providing temporally aligned guidance for spatiotemporal structure and dynamics.
Meanwhile, textual conditioning complements these high-level semantic representations with essential visual details, including subject appearance, environmental details, and fine-grained motion.
Further analyses of the spatiotemporal information encoded in visual semantics and the respective roles of semantic and textual conditioning are provided in \supp.

\myparagraph{Prompt Rewriter.}
During interactive generation, raw user instructions $T^{(m)}$ are often concise, incremental commands (\eg, ``he turns around'') that omit previously established visual appearances, potentially degrading rendering quality when directly fed into the DiT renderer.
To tackle this, we employ a lightweight prompt rewriter $\mathcal{P}$~\cite{qwen2026qwen35-2b} that aggregates historical instructions to rewrite the current prompt into a self-contained description of subjects and scenes:
\begin{equation}
    \widetilde{T}^{(m)} = \mathcal{P}\big(T^{(1)}, \dots, T^{(m)}\big),
    \label{eq:rewriter}
\end{equation}
which provides complete and consistent visual appearance descriptions tailored to the video renderer.

\myparagraph{Semantic-Guided Memory Retrieval.}
Beyond high-level semantic guidance, streaming video synthesis requires historical visual context $C_i$ to maintain continuity and long-range appearance consistency.
We follow~\cite{ji2025memflow,sun2025worldplay} to organize the visual context into three components:
\begin{equation}
    C_i = \big[C^{\mathrm{sink}},\, C_i^{\mathrm{mem}},\, C_i^{\mathrm{local}}\big],
    \label{eq:context}
\end{equation}
where $C^{\mathrm{sink}}$ serves as an attention sink from the initial chunk, $C_i^{\mathrm{local}}$ maintains recent local context within a sliding window, and $C_i^{\mathrm{mem}}$ preserves relevant historical chunks evicted from the local window.
Benefiting from our streaming pipeline, the planned semantics $S_i$ are already available before rendering $V_i$, allowing us to directly leverage $S_i$ as a fine-grained visual semantic prior to retrieve $C_i^{\mathrm{mem}}$.
Specifically, we spatially average the semantic embeddings of each chunk to obtain a compact chunk-level representation $\bar{s}_i = \mathrm{AvgPool}(S_i)$, and retrieve the most relevant historical chunk outside the local window based on cosine similarity:
\begin{equation}
    j^* = \operatorname*{arg\,max}_{j \in \{2,\dots,i-w\}}
    \operatorname{cos}(\bar{s}_i, \bar{s}_j),
    \label{eq:memory}
\end{equation}
where $w$ denotes the local window size.
The KV cache of video chunk $V_{j^*}$ then serves as the memory context $C_i^{\mathrm{mem}}$.

\myparagraph{Training.}
To equip the renderer with semantic controllability and long-horizon streaming generation capability, we train it in three stages:
1) \textit{Bidirectional semantic adaptation}: We first train the bidirectional DiT backbone with semantic conditioning using the standard flow matching objective~\cite{lipman2023flow}. Each video chunk attends to its corresponding semantic tokens and event prompt.
2) \textit{Causal distillation}: Following the causal distillation paradigm described in \cref{subsec:preliminary}, we replace bidirectional attention with chunk-wise causal attention and perform ODE initialization followed by DMD, obtaining a few-step causal video generator.
3) \textit{Memory-augmented streaming long tuning}: 
We adopt streaming long tuning~\cite{longlive, ji2025memflow} equipped with our semantic-guided memory retrieval. During this phase, the causal generator sequentially samples video chunks conditioned on visual context $C_i$ to perform rolling extension into long multi-event sequences, while the bidirectional teacher applies segment-wise DMD supervision throughout the entire rollout.

\section{\datasetname{}: Event-Centric Video Dataset}
\label{sec:data}

Existing video datasets are primarily organized around single-event clips~\cite{chen2024panda,nan2025openvid} or multi-shot videos~\cite{wang2026multishotmaster,zou2026mise,huang2026unityshots,luo2026shotstream} with holistic captions, lacking event-level temporal boundaries and annotations for continuous interactive video generation.
To bridge this gap, we develop an automated data pipeline (Fig.~\ref{fig:data_pipeline}) to construct \textbf{\datasetname}, a large-scale event-centric dataset covering both single-event and continuous multi-event videos with fine-grained event-wise annotations.

\myparagraph{Video Filtering and Classification.}
We first collect videos ranging from $5\text{s}$ to $60\text{s}$ from public datasets~\cite{wang2025koala} and internet sources.
Candidates are filtered using aesthetic and video quality scores~\cite{schuhmann2022laion,wang2025koala}, followed by TransNet V2~\cite{soucek2024transnet} to remove videos containing shot cuts.
We further combine a vision-language model (VLM)~\cite{qwen2026qwen35} with object detection~\cite{yolo} to exclude samples with unclear, overly small, or more than five primary subjects.
To distinguish single-event from multi-event videos, we manually label 10K videos based on semantic event distinctiveness and boundary clarity, and fine-tune Qwen3.6-27B~\cite{qwen2026qwen36} for event classification.

\myparagraph{Multi-Event Segmentation.}
For videos classified as multi-event, we employ Gemini 3.1 Pro~\cite{google2026gemini31} to locate event boundaries based on clear action transitions or completions at $0.5\text{s}$ temporal resolution, partitioning videos into contiguous, non-overlapping event segments while generating a concise natural-language instruction $T^{(m)}$ for each segment.
Samples containing event segments shorter than $2\text{s}$ are discarded to reduce boundary ambiguity and over-segmentation.

\myparagraph{Chained Event-Level Annotation.}
Directly annotating fine-grained events from an entire long video often leads to temporal grounding drift and cross-segment hallucinations.
We therefore adopt a segment-wise chained annotation pipeline using Qwen3.6-27B: single-event videos and the first segment of multi-event videos are independently annotated with a self-contained structured caption $\widetilde{T}^{(1)}$; for each subsequent segment $m \ge 2$, the model takes both the preceding and current video segments along with the previous caption $\widetilde{T}^{(m-1)}$ to jointly output a self-contained caption $\widetilde{T}^{(m)}$ and a context-dependent event-change description $R^{(m)}$.
Organized under structured fields (\eg, subject, environment, and camera), $\widetilde{T}^{(m)}$ independently details current appearances, actions, backgrounds, and camera motion, whereas $R^{(m)}$ captures how these elements evolve from the preceding segment.
Both annotations maintain consistent subject identities and persistent visual attributes across consecutive events.

\myparagraph{Quality Verification.}
To further ensure annotation quality, we perform global verification using each complete video and all its event annotations.
Gemini 3.1 Pro examines subject identity consistency and factual accuracy, accepting consistent annotations, discarding samples with severe conflicts, and flagging those requiring revision.
Flagged annotations undergo human review and correction, while randomly sampled accepted annotations are manually audited to assess the reliability of the automated pipeline.

The resulting dataset contains approximately 2.5M single-event and 245K continuous multi-event videos, equipping each event with concise user instructions, self-contained structured descriptions, and context-dependent event-change annotations to support the training of semantic reasoning, planning, and video rendering.

\section{Experiments}
\label{sec:exp}

\begin{figure*}[t]
    \centering
    \includegraphics[width=1.0\textwidth]{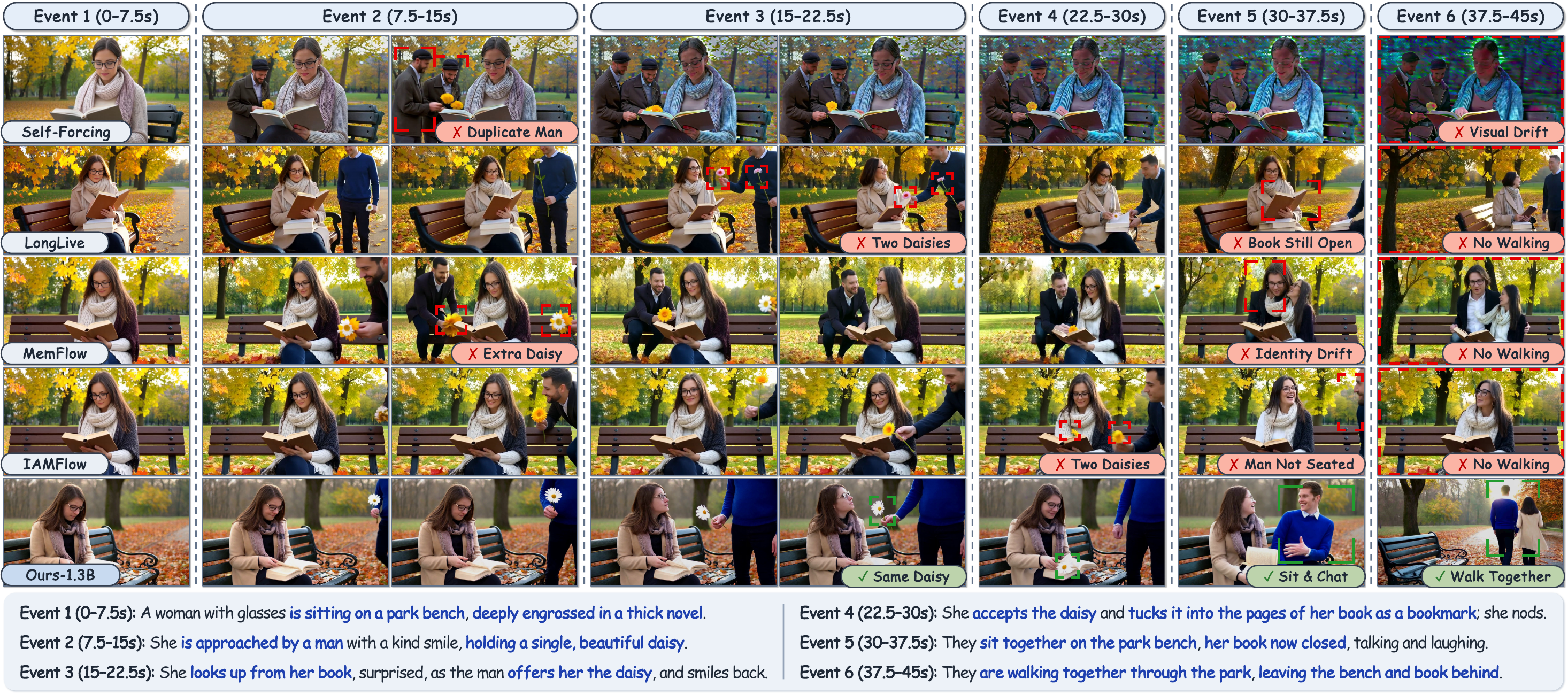}
    \vspace{-1.5em}
    \caption{\textbf{Qualitative comparison on multi-event interactive video generation.}
    Baselines suffer from incoherent within-event progression (\eg, spawning a duplicate man), abrupt cross-event transitions (\eg, hallucinating a second daisy instead of handing over the existing one), and poor instruction following or visual degradation in later events (\eg, failing to sit or walk).
    In contrast, our method maintains consistent entities and appearance while faithfully executing evolving instructions across all events.}
    \label{fig:main_comp}
    \vspace{-0.3em}
\end{figure*}

\input{tables/main_compare_interact}

\myparagraph{Implementation Details.}
We initialize our LM planner from Qwen2.5-7B-Instruct~\cite{yang2024qwen25} and build two renderer variants upon Wan2.1-1.3B and Wan2.2-5B~\cite{wan2025wan}, with semantic tokens and video chunks temporally aligned at $0.5\text{s}$ intervals.
The planner is trained with a $10\text{K}$-token context window.
All models are trained on $32$ A800 GPUs with a batch size of $64$.
Further implementation details are provided in \supp.

\subsection{Interactive Video Generation}
\label{sec:exp_interactive}

We compare \methodname{} with interactive streaming baselines LongLive~\cite{longlive}, MemFlow~\cite{ji2025memflow}, Anchor Forcing~\cite{anchorforcing2026}, and IAMFlow~\cite{iamflow}, and extend representative single-prompt streaming models to the multi-prompt setting, including Self-Forcing~\cite{huang2026self}, Rolling Forcing~\cite{rollingforcing2025}, and Infinity-RoPE~\cite{infinityrope2025}.
We evaluate on the $324$ prompt sequences from NarraStream-Bench~\cite{iamflow} using its official metrics.
To accommodate the planner's $10\text{K}$-token context budget, we adapt the six-segment protocol to $7.5\text{s}$ per segment ($45\text{s}$ in total), with all methods evaluated under the same setting.
Additional evaluations on the interactive benchmark introduced by MemFlow~\cite{ji2025memflow} are provided in \supp.

\myparagraph{Quantitative Comparison.}
As shown in Tab.~\ref{tab:main_comp_interact}, \methodname{} outperforms baselines across most metrics, with superior visual quality, temporal consistency, and instruction compliance.
In particular, our improvements in Temporal Flickering, Motion Smoothness, Temporal Stability, Boundary Smoothness, and Conditional Adjacent Consistency indicate smoother behavioral evolution and more coherent event transitions.
We attribute our slightly lower Subject Consistency to richer behavioral dynamics, as reflected in higher Dynamic Trajectory scores, while strong Background and Long-range Consistency, together with higher human-rated Global Consistency in Fig.~\ref{fig:teaser}(c), confirm faithful identity and scene preservation.
Notably, \methodname{} achieves substantial gains in VLM Alignment ($80.00$/$80.83$ \textit{vs.}\ $66.33$), demonstrating stronger adherence to evolving instructions and coherent narrative progression across segments.

To more directly evaluate semantic coherence across and within events, we incorporate two targeted metrics following the VLM-based evaluation paradigm of NarraStream-Bench:
\textit{Cross-Event Transition} assesses whether new events follow incoming instructions while transitioning coherently and plausibly
from the established visual context, 
whereas \textit{Within-Event Progression} evaluates whether instructed behaviors evolve meaningfully toward their intended outcomes without repetitive execution or abrupt discontinuities.
As shown in the rightmost two columns of Tab.~\ref{tab:main_comp_interact}, \methodname{} substantially surpasses all baselines on both metrics, confirming its effectiveness in generating semantically coherent narratives.
Detailed evaluation protocols and validation against human judgments are provided in \supp.

\myparagraph{Qualitative Comparison.}
As shown in Fig.~\ref{fig:teaser}(a) and Fig.~\ref{fig:main_comp}, baselines struggle with semantic evolution, often exhibiting abrupt, context-agnostic transitions at prompt switches, repetitive or discontinuous actions within events, and even instruction-following failures.
Moreover, without memory mechanisms, Self-Forcing suffers from severe visual degradation over long horizons.
In contrast, with online semantic reasoning, planning, and dedicated memory retrieval, \methodname{} achieves faithful instruction following, coherent event transitions, and meaningful behavioral progression while preserving consistent visual appearance.

\myparagraph{User Study.}
We conduct a user study with $30$ participants on $20$ multi-event prompt sequences, comparing our 1.3B variant against representative baselines on a 5-point Likert scale across six dimensions.
As shown in Fig.~\ref{fig:teaser}(c), \methodname{} consistently achieves the highest Mean Opinion Scores (MOS) across all dimensions, with pronounced gains in instruction following and semantic coherence, confirming superior perceptual and narrative quality.

\subsection{Single-Prompt Video Generation}
\label{sec:exp_single}

\input{tables/main_compare_short}

We also evaluate the general generation quality of \methodname on $5\text{s}$ single-prompt video generation using VBench~\cite{huang2024vbench}.
As shown in Tab.~\ref{tab:main_comp_short}, our method outperforms
all streaming baselines and corresponding Wan backbones
in overall performance, highlighting the benefits of explicit semantic planning even for short clips.
It also achieves better semantic alignment and comparable
visual quality to the offline planner-renderer framework
Bernini~\cite{bernini2026} at over $50{\times}$ higher throughput.
By overlapping planning and rendering in an asynchronous pipeline, \methodname{} reaches idealized steady-state throughput competitive with real-time AR baselines, with detailed breakdowns and end-to-end deployment results provided in \supp.
We also evaluate 30s generation
on VBench-Long~\cite{huang2025vbench++},
confirming strong semantic alignment and visual consistency (see \supp).

\subsection{Ablation Study}
\label{sec:exp_ablation}

\input{tables/main_ablation}

\begin{figure}[t]
    \centering
    \includegraphics[width=\columnwidth]{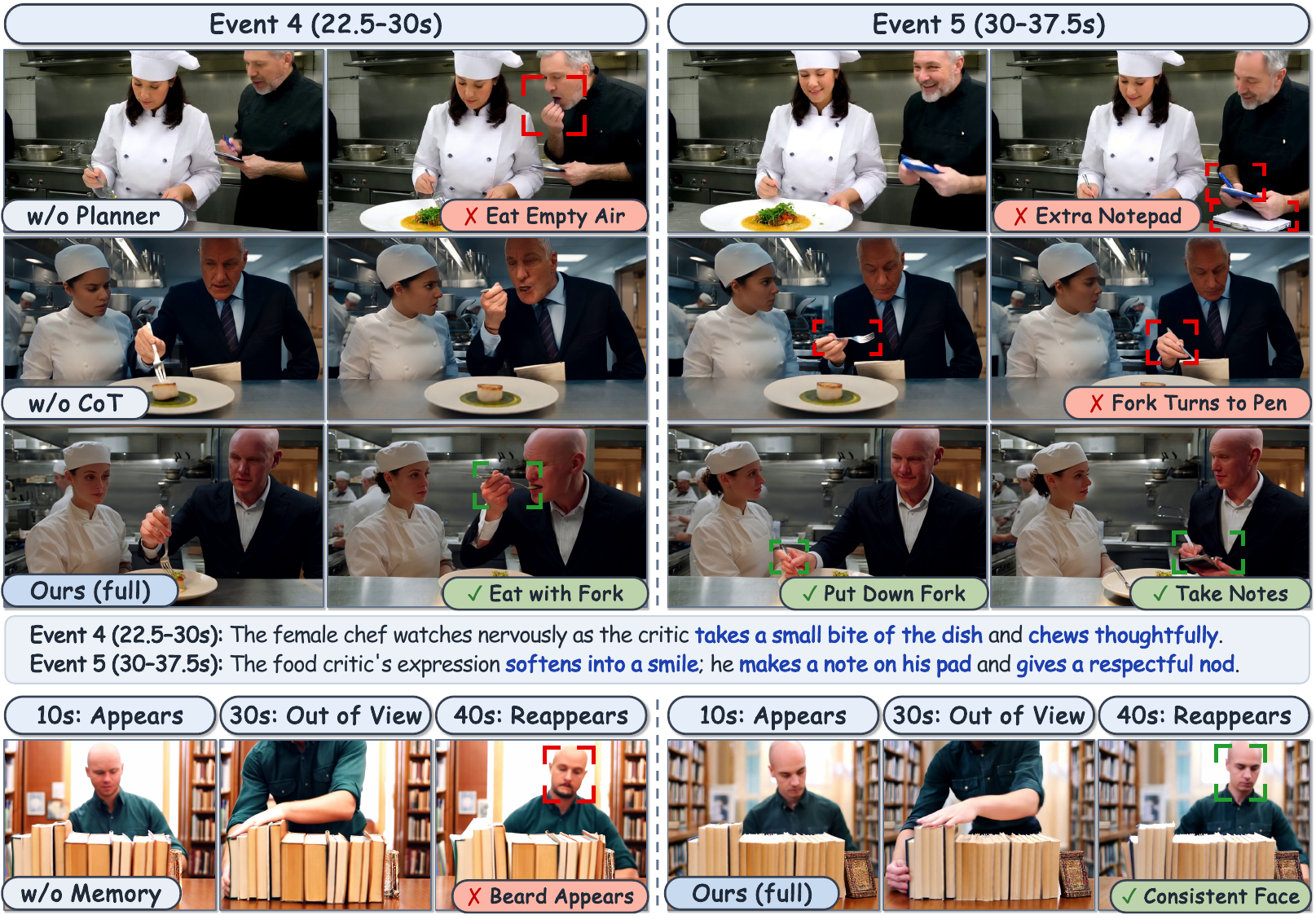}
    \caption{\textbf{Qualitative ablation study} on semantic planning and reasoning (top) and semantic-guided memory retrieval (bottom).}
    \label{fig:ablation}
    \vspace{-.8em}
\end{figure}

We ablate the key components on NarraStream-Bench using the $1.3\text{B}$ variant, with results shown in Tab.~\ref{tab:ablation} and Fig.~\ref{fig:ablation}.

\myparagraph{Semantic Planning and Reasoning.}
Removing the planner (\textit{w/o planner}), even when retraining the renderer on \datasetname{} with text-based memory retrieval, severely degrades instruction compliance and semantic coherence, indicating that multi-event data alone cannot compensate for the lack of explicit semantic reasoning and planning in causal DiTs.
Further introducing an external 7B VLM~\cite{qwen25vl} to infer event changes and evolution as textual conditioning (\textit{+ external reasoning}) yields only marginal gains, with incoherent transitions and progression
persisting.
This confirms that textual guidance alone cannot be reliably grounded by causal DiTs over extended horizons, highlighting the importance of autoregressively planning spatiotemporally aligned visual semantics.
Finally, removing context-aware CoT (\textit{w/o CoT})
primarily degrades cross-event transitions, validating its role as a semantic bridge for subsequent planning.

\myparagraph{Semantic-Guided Memory Retrieval.}
Removing retrieved visual memory (\textit{w/o memory} $C_i^{\mathrm{mem}}$)
primarily degrades visual quality and temporal consistency, while leaving instruction compliance and semantic coherence largely intact.
This manifests as appearance drift when subjects reappear after leaving the view.
Replacing our retrieval with random (\textit{random memory}) or text-based selection~\cite{ji2025memflow} (\textit{text-based retrieval}) partially restores performance but still lags behind our full model.
Unlike coarse event-level text prompts, our chunk-level retrieval operates in a visually grounded semantic space, enabling more precise recall of historical frames with matching subject and scene configurations.

\myparagraph{Prompt Rewriting.}
Compared to raw instructions, our self-contained rewritten prompts better preserve appearance and motion details while improving temporal consistency.
Further ablation settings are provided in \supp.

\section{Discussion and Conclusion}

\myparagraph{Limitations and Future Work.}
Our current planner is bounded by its finite context window, as visual semantic tokens accumulate throughout generation.
Future work will explore context compression and memory
mechanisms to support longer interactive streams.

\myparagraph{Conclusion.}
We presented \methodname{}, a streaming interactive video
generation framework combining autoregressive semantic planning
with causal DiT rendering.
Through explicit semantic evolution modeling and semantic-guided memory, our approach achieves coherent event transitions and progression
while maintaining visual consistency.
Together with our proposed \datasetname{} dataset, our work advances streaming video generation toward semantically coherent interactive narratives.

%% file: tables/main_compare_interact.tex
\begin{table*}[t]
\centering
\caption{
\textbf{Quantitative comparison of interactive video generation on NarraStream-Bench} under the 45s (6${\times}$7.5s) multi-prompt setting. 
Baselines take original instructions while \methodname{} retains its built-in prompt rewriter; all generated videos are evaluated against the original instructions for fairness. 
Additional comparisons using shared structured captions
are provided in \supp, yielding consistent conclusions.
Best and second-best scores are \textbf{bolded} and \uline{underlined} across all tables.}
\vspace{-.3em}
\label{tab:main_comp_interact}
\setlength{\tabcolsep}{2.8pt}
\resizebox{\textwidth}{!}{%
\begin{tabular}{@{}l*{13}{c}@{}}

\toprule
\multirow{2}{*}{Model}
& \multicolumn{5}{c}{Visual Quality $\uparrow$}
& \multicolumn{3}{c}{Temporal Consistency $\uparrow$}
& \multicolumn{3}{c}{Instruction Compliance $\uparrow$}
& \multicolumn{2}{c}{Semantic Coherence $\uparrow$} \\
\cmidrule(lr){2-6}
\cmidrule(lr){7-9}
\cmidrule(lr){10-12}
\cmidrule(lr){13-14}
& \shortstack{\strut Subject\\\strut Consist.}
& \shortstack{\strut Background\\\strut Consist.}
& \shortstack{\strut Temporal\\\strut Flicker.}
& \shortstack{\strut Motion\\\strut Smooth.}
& \shortstack{\strut Temporal\\\strut Stability}
& \shortstack{\strut Boundary\\\strut Smooth.}
& \shortstack{\strut Adjacent\\\strut Consist.}
& \shortstack{\strut Long-range\\\strut Consist.}
& \shortstack{\strut Entity\\\strut Ground.}
& \shortstack{\strut Dynamic\\\strut Traj.}
& \shortstack{\strut VLM\\\strut Align.}
& \shortstack[c]{\strut Cross-Event\\\strut Transition}
& \shortstack[c]{\strut Within-Event\\\strut Progression} \\
\midrule

Self-Forcing
& 84.02 & 83.46 & 88.29 & 58.58
& 29.69 & 48.75 & 57.99 & 45.88
& 61.18 & 55.17 & 53.62
& 45.42 & 45.99 \\

Rolling Forcing
& 92.52 & 89.81 & 85.61 & 49.46
& 68.06 & 40.14 & 76.54 & 61.27
& 68.17 & 58.67 & 64.45
& 52.28 & 51.27 \\

Infinity-RoPE
& 92.76 & 91.66 & 89.45 & 54.82
& 73.20 & 50.13 & 76.16 & 80.78
& 69.59 & 59.97 & 63.99
& 50.81 & 52.64 \\

LongLive
& \uline{95.35} & \uline{92.85}
& 91.27 & 58.95
& 74.20 & 49.21
& 83.81 & 89.63
& 72.58 & 63.41 & 65.07
& 56.15 & 58.70 \\

MemFlow
& 95.06 & 92.23 & 90.98 & 57.22
& 73.14 & 44.22 & 83.58 & 88.85
& 70.00 & 63.91 & 63.15
& 59.72 & 63.19 \\

Anchor Forcing
& 90.32 & 89.72 & 85.59 & 40.75
& 71.03 & \uline{55.86}
& 68.41 & 75.08
& 71.36 & 58.13 & 66.33
& 60.33 & 61.31 \\

IAMFlow
& \textbf{95.37} & 92.45
& 91.11 & 57.07
& 74.18 & 47.69
& 85.40 & \uline{90.44}
& 69.14 & 64.47 & 60.63
& 58.10 & 65.70 \\

\midrule

\rowcolor{OursBlue}
\textbf{Ours-1.3B}
& 93.47 & \textbf{93.16}
& \textbf{92.98} & \textbf{63.08}
& \uline{74.25} & 55.74
& \uline{85.42} & 89.68
& \uline{76.37} & \textbf{64.97}
& \uline{80.00}
& \uline{78.59} & \uline{81.68} \\

\rowcolor{OursBlue}
\textbf{Ours-5B}
& 94.41 & 92.58
& \uline{92.67} & \uline{62.26}
& \textbf{75.31} & \textbf{56.06}
& \textbf{85.57} & \textbf{90.47}
& \textbf{77.84} & \uline{64.93}
& \textbf{80.83}
& \textbf{79.63} & \textbf{83.05} \\

\bottomrule
\end{tabular}%
}
\vspace{-1em}
\end{table*}

%% file: tables/main_compare_short.tex
\begin{table}[t]
\centering
\setlength{\tabcolsep}{3pt}
\caption{
\textbf{Quantitative comparison of 5s single-prompt video generation on VBench}
using extended prompts~\cite{huang2026self}.
\#Params lists planner and renderer sizes where applicable.
Throughput is measured on H800 GPUs under model-specific inference configurations, with idealized pipeline overlap for pipelined methods.
}
\label{tab:main_comp_short}
\resizebox{\columnwidth}{!}{%
\begin{tabular}{@{}l@{\hspace{1pt}}c@{\hspace{2.5pt}}c@{\hspace{2.5pt}}c@{\hspace{2pt}}ccc@{}}
\toprule
\multirow{2}{*}{Model}
& \multirow{2}{*}{\#Params}
& \multirow{2}{*}{Resolution}
& \multirow{2}{*}{\shortstack{Throughput\\(FPS) $\uparrow$}}
& \multicolumn{3}{c}{Evaluation Scores $\uparrow$} \\
\cmidrule(lr){5-7}
& & & & Total & Quality & Semantic \\
\midrule

\rowcolor{GroupGray}
\multicolumn{7}{@{}l}{\textit{Non-streaming Video Models}} \\

Wan2.1-1.3B
& 1.3B & $832\!\times\!480$
& 0.91 & 84.27 & 84.99 & 81.39 \\

Wan2.2-5B
& 5B & $1280\!\times\!704$
& 0.77 & 83.75 & 83.89 & 83.16 \\

Bernini
& 7B+14B & $848\!\times\!480$
& 0.34 & \uline{84.83} & \textbf{85.51} & 82.12 \\

\midrule
\rowcolor{GroupGray}
\multicolumn{7}{@{}l}{\textit{Streaming Autoregressive Models}} \\

Self-Forcing
& 1.3B & $832\!\times\!480$
& 18.0 & 84.13 & 84.81 & 81.41 \\

Rolling Forcing
& 1.3B & $832\!\times\!480$
& 16.8 & 83.29 & 83.93 & 80.76 \\

Infinity-RoPE
& 1.3B & $832\!\times\!480$
& 26.3 & 83.49 & 84.41 & 79.82 \\

LongLive
& 1.3B & $832\!\times\!480$
& 20.0 & 83.30 & 83.87 & 81.02 \\

MemFlow
& 1.3B & $832\!\times\!480$
& 18.4 & 83.08 & 83.51 & 81.34 \\

Anchor Forcing
& 1.3B & $832\!\times\!480$
& 17.3 & 83.90 & 84.50 & 81.47 \\

IAMFlow
& 1.3B & $832\!\times\!480$
& 16.9 & 83.24 & 83.75 & 81.20 \\

\midrule
\rowcolor{OursBlue}
\textbf{Ours-1.3B}
& 7B+1.3B & $832\!\times\!480$
& 19.2 & 84.44 & 84.69 & \uline{83.46} \\

\rowcolor{OursBlue}
\textbf{Ours-5B}
& 7B+5B & $1280\!\times\!704$
& 19.1 & \textbf{84.90} & \uline{85.23} & \textbf{83.60} \\

\bottomrule
\end{tabular}
}
\end{table}

%% file: tables/main_ablation.tex
\begin{table}[t]
\centering
\caption{
\textbf{Ablations on NarraStream-Bench (45s).}
}
\vspace{-.3em}
\label{tab:ablation}
\setlength{\tabcolsep}{2pt}
\resizebox{\columnwidth}{!}{%
\begin{tabular}{@{}l@{\hspace{1pt}}ccccc@{}}
\toprule
Variant
& \shortstack{\strut Visual\\\strut Quality $\uparrow$}
& \shortstack{\strut Temp.\\\strut Consist. $\uparrow$}
& \shortstack{\strut Instr.\\\strut Compl. $\uparrow$}
& \shortstack{\strut Cross-Evt.\\\strut Transition $\uparrow$}
& \shortstack{\strut Within-Evt.\\\strut Progression $\uparrow$} \\
\midrule

\rowcolor{GroupGray}
\multicolumn{6}{@{}l}{\textit{Semantic Planning and Reasoning}} \\
w/o planner
& 82.66 & 73.54 & 66.50 & 57.11 & 63.85 \\
\quad + external reasoning
& 81.97 & 74.01 & 67.49 & 58.62 & 65.55 \\
w/o CoT
& \uline{82.90} & \uline{76.39} & 72.93 & 74.06 & 81.34 \\

\midrule
\rowcolor{GroupGray}
\multicolumn{6}{@{}l}{\textit{Visual Memory \& Prompt Rewriting}} \\
w/o memory $C_i^{\mathrm{mem}}$
& 79.43 & 70.04 & 73.30 & 77.98 & 80.66 \\
random memory
& 80.69 & 70.13 & 73.17 & 78.17 & 81.64 \\
text-based retrieval
& 80.71 & 72.59 & \uline{73.33} & \textbf{79.00} & 81.47 \\
w/o prompt rewriter
& 82.59 & 73.43 & 72.35 & 78.55 & \textbf{82.19} \\

\midrule
\rowcolor{OursBlue}
\textbf{Ours (full)}
& \textbf{83.39} & \textbf{76.95} & \textbf{73.78} & \uline{78.59} & \uline{81.68} \\

\bottomrule
\end{tabular}%
}
\vspace{-.5em}
\end{table}

%% file: sec/X_suppl.tex
\clearpage
\appendix

\definecolor{suppTocBlue}{RGB}{49,92,129}

\renewcommand{\maketitlesupplementary}{%
    \onecolumn
    \raggedbottom
    {\centering
     \Large
     \textbf{\thetitle}\\
     \vspace{0.5em}
     Supplementary Material\\
     \vspace{1.0em}\par}%
}
\maketitlesupplementary

\setcounter{figure}{0}
\setcounter{table}{0}
\setcounter{equation}{0}

\renewcommand{\thefigure}{S\arabic{figure}}
\renewcommand{\thetable}{S\arabic{table}}
\renewcommand{\theequation}{S\arabic{equation}}

\renewcommand{\theHfigure}{S\arabic{figure}}
\renewcommand{\theHtable}{S\arabic{table}}
\renewcommand{\theHequation}{S\arabic{equation}}

\startcontents[supp]

\section*{Contents}

\hypersetup{linkcolor=suppTocBlue}

\begingroup

\titlecontents{section}
    [2.0em]
    {\addvspace{8pt}\large\bfseries}
    {\contentslabel{2.0em}}
    {}
    {\hfill\contentspage\hspace*{1.5em}}

\printcontents[supp]{}{1}{%
    \setcounter{tocdepth}{1}%
}

\endgroup

\clearpage

\hypersetup{linkcolor=cvprblue}

\linespread{1.02}\selectfont

\section{Autoregressive Semantic Planner: Implementation Details}
\label{sec:supp_impl_planner}

\myparagraph{Model Initialization and Visual Tokenization.}
Our autoregressive semantic planner is built upon Qwen2.5-7B-Instruct~\cite{yang2024qwen25}; specifically, we initialize it from the image-pretrained Tar-7B checkpoint~\cite{tatok}, whose visual token embeddings are initialized from the text-aligned TA-Tok codebook.
Following~\cite{tatok}, we adopt the SigLIP2-based text-aligned visual tokenizer (TA-Tok), which encodes individual visual frames at three spatial scales (\texttt{<S0>}, \texttt{<S1>}, and \texttt{<S2>}) into $729$ ($27\times27$), $169$ ($13\times13$), and $81$ ($9\times9$) discrete tokens, respectively.
The visual vocabulary contains $65{,}536$ tokens (\texttt{<I0>}-\texttt{<I65535>}), with \texttt{<im\_start>} and \texttt{<im\_end>} delimiting each visual token sequence.

For video tokenization, we sample one representative keyframe per $0.5\text{s}$ video chunk $V_i$ (\ie, $2\text{ fps}$).
The sampled keyframe is resized to $384\times384$ and encoded at the coarsest scale (\texttt{<S2>}) into $9\times9$ spatially structured tokens.
Each grid is flattened in raster order into $K=81$ discrete token indices,
$S_i=(s_{i,1},\ldots,s_{i,K})$, with $s_{i,k}\in\{0,\ldots,65535\}$.
We denote the corresponding visual token block by $\mathcal{B}(S_i)$, which consists of \texttt{<im\_start>}, \texttt{<S2>}, the $81$ visual tokens, and \texttt{<im\_end>}, occupying $84$ token positions in the LM sequence.

\myparagraph{Autoregressive Sequence Formatting.}
Similar to~\cite{tatok,planx}, we interleave textual and visual tokens in the Qwen chat format, organizing user instructions, structured event-change reasoning, and visual semantic tokens into a multi-turn conversational sequence.
The special tokens \texttt{<|im\_start|>} and \texttt{<|im\_end|>} delimit individual messages, with the former followed by a role identifier (system, user, or assistant).
We introduce reasoning delimiters \texttt{<think>} and \texttt{</think>} to enclose context-aware CoT predictions prior to visual semantic tokens.
The standard autoregressive cross-entropy objective supervises assistant-side reasoning and visual semantic tokens, while system and user messages are excluded from the loss.
Let $n_m$ denote the cumulative number of video chunks through event $m$.
The multi-turn training sequence format is detailed below, where gray bracketed entries denote optional conditioning inputs or turn-ending markers controlled during training.

\input{figures/supp_planner_format}

Within this unified format, we support both text-to-video (T2V) and text-image-to-video (TI2V) generation.
T2V starts directly from $T^{(1)}$, whereas TI2V additionally supplies the initial-frame tokens $\mathcal{B}_0(I_{\mathrm{init}})$ at the finest scale (\texttt{<S0>}) to provide fine-grained spatial details.
Both settings autoregressively predict sequences of visual semantic blocks $\mathcal{B}(S_i)$, and at each subsequent prompt switch ($m\ge2$), the planner generates structured event-change reasoning $R^{(m)}$ prior to predicting the new event's visual semantics.

\myparagraph{Optional Duration Conditioning.}
To accommodate both specified-duration planning (\eg, user-prescribed event horizons or standardized benchmark protocols) and open-ended planning (where events unfold naturally without a preset duration until user interruption), we apply optional duration conditioning during training.
With a probability of $0.5$, an explicit event duration $D^{(m)}$ is included in the user prompt and the assistant turn-ending token \texttt{<|im\_end|>} is supervised at the event boundary, encouraging the planner to pace semantic progression toward the target horizon.
Otherwise, the duration hint is omitted and \texttt{<|im\_end|>} supervision is masked, allowing semantic prediction to continue seamlessly until a new user instruction arrives (at which point the ongoing turn is closed and the new instruction is appended).
Further analyses of these two planning modes are provided in Sec.~\ref{sec:supp_flexible}.

\myparagraph{Three-Stage Training Curriculum.}
Starting from the Tar-7B checkpoint, we train our semantic planner in three progressive stages:
1) \textit{Large-scale semantic pretraining:}
We train on $10$M single-event video-text pairs from Koala-36M~\cite{wang2025koala} for $6$ epochs, establishing foundational text-to-semantic sequence modeling across temporally continuous video frames.
2) \textit{Single-event refinement:}
We further train on $2.5$M curated single-event videos from \datasetname{} for $5$ epochs to refine spatiotemporal semantic dynamics and strengthen instruction adherence.
3) \textit{Multi-event interactive fine-tuning:}
Finally, we fine-tune on $245$K multi-event videos from \datasetname{} for $3$ epochs, incorporating context-aware CoT supervision at event boundaries to enable coherent interactive semantic planning.

All stages use the AdamW optimizer with a learning rate of $1\times10^{-5}$, a global batch size of $64$, and a $10\text{K}$-token ($10{,}240$) context window on $32$ NVIDIA A800 GPUs.
Across all stages, T2V and TI2V samples are mixed at a $0.7:0.3$ ratio.
When training on \datasetname{}, initial prompts randomly alternate between concise instructions $T^{(1)}$ and self-contained structured captions $\widetilde{T}^{(1)}$ to accommodate diverse user prompt styles.
Furthermore, to reduce exposure bias and autoregressive rollout drift, we apply stochastic token perturbation during teacher-forced training: each ground-truth visual token in the input sequence is replaced with probability $0.1$ by a uniform sample from its top-$16$ cosine-nearest neighbors in the TA-Tok codebook, while target tokens for loss computation remain unperturbed.

\myparagraph{Planner Inference Settings.}
We employ separate sampling temperatures for textual reasoning and visual semantic prediction.
Structured CoT tokens are sampled at a temperature of $0.6$ to favor stable event-transition reasoning, while visual semantic tokens use a temperature of $0.9$ with top-$p$ sampling ($p=0.9$) to allow diverse and natural semantic evolution.
For quantitative benchmark evaluations, we operate strictly in pure T2V mode (from text prompts alone without initial images).
By default, we provide the prescribed event duration (\eg, $7.5\text{s}$) and trigger context-aware CoT at each prompt switch ($m\geq 2$).
Open-ended planning without advance duration conditioning is further evaluated in Sec.~\ref{sec:supp_flexible}.

\section{Semantic-Guided Streaming DiT Renderer: Implementation Details}
\label{sec:supp_impl_renderer}

\myparagraph{Backbones and Semantic Feature Conditioning.}
We develop two streaming renderer variants based on Wan2.1-T2V-1.3B ($832\times480$, $16$ fps) and Wan2.2-TI2V-5B ($1280\times704$, $24$ fps)~\cite{wan2025wan}, both adapted for T2V and TI2V generation.
Both renderers generate video chunks aligned with the planner's $0.5\text{s}$ semantic prediction interval.
Under the $4\times$ temporal compression of their causal VAEs, each chunk contains $2$ latent frames ($8$ video frames) for the $1.3$B variant and $3$ latent frames ($12$ video frames) for the $5$B variant.

While the planner predicts discrete semantic tokens, the renderer operates on continuous semantic features.
Specifically, the frozen TA-Tok reconstruction module decodes each predicted $9\times9$ token grid into a continuous semantic feature map in $\mathbb{R}^{9\times9\times1152}$ (denoted by the same symbol $S_i$ in Sec.~3.3 of the main paper for notational simplicity).
We flatten the spatial dimensions and project the resulting $81$ semantic features through a two-layer MLP to the DiT hidden dimension ($1536$ for $1.3$B and $3072$ for $5$B).

Following the joint conditioning formulation in~\cite{planx}, we augment each DiT block's text cross-attention with semantic-specific key and value projections.
Text and semantic tokens share the video queries, while rotary positional encoding is applied to the semantic query-key interactions to preserve spatiotemporal alignment, leaving the original text cross-attention unchanged.
Denoting the attention logits for the two branches by $L_{\mathrm{text}}$ and $L_{\mathrm{sem}}$, we compute
\begin{equation}
\begin{aligned}
L_{\mathrm{text}} &= \frac{QK_{\mathrm{text}}^\top}{\sqrt{d}},\quad
L_{\mathrm{sem}} &= \frac{\widetilde Q\widetilde K_{\mathrm{sem}}^\top}{\sqrt{d}},
\end{aligned}
\end{equation}
where the tilde denotes the features with semantic-branch positional encoding.
The logits are jointly normalized to aggregate both textual and semantic conditions:
\begin{equation}
\operatorname{Attn} = \operatorname{softmax}\left([L_{\mathrm{text}}\Vert L_{\mathrm{sem}}]\right)[V_{\mathrm{text}}\Vert V_{\mathrm{sem}}],
\end{equation}
where $\Vert$ denotes concatenation along the conditioning sequence dimension.

\begin{algorithm}[!t]
\caption{Streaming Interactive Inference of Semantic Forcing}
\label{alg:supp_streaming_inference}
\begin{algorithmic}[1]
\Require Autoregressive planner $\mathcal{M}_{\mathrm{LM}}$, prompt rewriter $\mathcal{P}$, $4$-step causal DiT renderer $G_\theta$ (with clean KV extractor $G_\theta^{\mathrm{KV}}$), denoising timesteps $\{t_4, t_3, t_2, t_1\}$, local window width $w=3$
\State Initialize planner context $\mathcal{H} \leftarrow \emptyset$, chunk KV archive $\mathcal{K} \leftarrow \emptyset$, descriptor bank $\mathcal{D} \leftarrow \emptyset$, output video $\mathcal{V} \leftarrow []$
\For{chunk index $i = 1, 2, \dots$}
    \If{new user instruction $T^{(m)}$ arrives at chunk $i$}
        \State Obtain rewritten prompt $\widetilde{T}^{(m)} \leftarrow \mathcal{P}(T^{(1:m)})$ 
        \State Append $T^{(m)}$ to $\mathcal{H}$; if $m \ge 2$, sample CoT reasoning $R^{(m)} \sim \mathcal{M}_{\mathrm{LM}}(\cdot \mid \mathcal{H})$ and append to $\mathcal{H}$
    \EndIf
    \State Predict discrete semantic block $S_i \sim \mathcal{M}_{\mathrm{LM}}(\cdot \mid \mathcal{H})$, append to $\mathcal{H}$, and decode continuous features $S_i \in \mathbb{R}^{9\times 9\times 1152}$
    \State Compute chunk descriptor $\bar{s}_i \leftarrow \operatorname{AvgPool}(S_i)$
    \If{$i \le w + 1$}
        \State Assemble active KV context $C_i \leftarrow \mathcal{K}_{1:i-1}$ \Comment{Sink and local window cover all preceding chunks}
    \Else
        \State Retrieve $j^* \leftarrow \arg\max_{j \in \{2, \dots, i-w\}} \cos(\bar{s}_i, \bar{s}_j)$ from $\mathcal{D}$
        \State Assemble active KV context $C_i \leftarrow [\mathcal{K}_1,\, \mathcal{K}_{j^*},\, \mathcal{K}_{i-w+1:i-1}]$ with compact RoPE reassignment
    \EndIf
    \State Sample initial chunk latent $z_{t_4}^i \sim \mathcal{N}(0, I)$
    \For{$k = 4, 3, 2, 1$}
        \State Predict clean latent $\hat{z}_0^i \leftarrow G_\theta(z_{t_k}^i;\, t_k, S_i, \widetilde{T}^{(m)}, C_i)$
        \If{$k > 1$}
            \State Sample $\epsilon \sim \mathcal{N}(0, I)$ and step forward $z_{t_{k-1}}^i \leftarrow \Psi(\hat{z}_0^i, \epsilon, t_{k-1})$
        \EndIf
    \EndFor
    \State Extract clean chunk KV cache $\mathcal{K}_i \leftarrow G_\theta^{\mathrm{KV}}(\hat{z}_0^i;\, t=0, S_i, \widetilde{T}^{(m)}, C_i)$
    \State Store $\mathcal{K}_i$ in $\mathcal{K}$, store $\bar{s}_i$ in $\mathcal{D}$, and decode $\hat{z}_0^i$ via causal VAE to append $V_i$ to $\mathcal{V}$
\EndFor
\State \Return $\mathcal{V}$
\end{algorithmic}
\end{algorithm}

\myparagraph{Causal Visual Context and Memory Retrieval.}
During streaming generation, each target chunk $V_i$ is synthesized conditioned on the historical visual context $C_i=[C^{\mathrm{sink}},C_i^{\mathrm{mem}},C_i^{\mathrm{local}}]$ using the KV caches of previously denoised chunks.
Specifically, $C^{\mathrm{sink}}$ retains the initial chunk $V_1$, while the local attention window has width $w=3$, comprising the current target chunk $V_i$ and the two most recent chunks $\{V_{i-2},V_{i-1}\}$.
The retrieved memory $C_i^{\mathrm{mem}}$ contains one historical chunk $V_{j^*}$ selected from $\{V_2,\ldots,V_{i-3}\}$, excluding the sink and local context.

For retrieval, we spatially average the continuous semantic features to obtain a chunk-level descriptor $\bar{s}_i=\operatorname{AvgPool}(S_i)\in\mathbb{R}^{1152}$.
As described in Eq.~(5) of the main paper, the most relevant historical chunk is selected by cosine similarity between the planned semantics of the upcoming chunk and preceding semantic features.
Its cached keys and values are then incorporated into the current visual context before rendering $V_i$.
Similar to~\cite{rollingforcing2025}, we reassign rotary positions to the selected sink, retrieved memory, and local chunks within a compact contiguous temporal range, avoiding excessively distant relative positions in causal self-attention.
Alg.~\ref{alg:supp_streaming_inference} summarizes the complete streaming inference procedure.

\myparagraph{Three-Stage Training and Distillation.}
We train the renderer on $32$ NVIDIA A800 GPUs with a global batch size of $64$ in three stages:
1) \textit{Bidirectional semantic adaptation:}
We first train the bidirectional DiT with semantic conditioning for $20$K iterations ($\mathrm{lr}=1\times10^{-5}$) using the standard flow-matching objective~\cite{lipman2023flow}.
Training samples are randomly cropped into $5\text{s}$ clips from \datasetname{} multi-event videos, containing $81$ or $121$ frames for the $1.3$B and $5$B renderers, respectively.
The cropped clips may span event boundaries, with each temporally aligned chunk conditioned on its corresponding semantic features and event-specific text description.
We randomly alternate between original instructions $T^{(m)}$ and self-contained structured captions $\widetilde T^{(m)}$ to improve robustness to different levels of textual detail.
2) \textit{Causal adaptation and distillation:}
We replace bidirectional self-attention with chunk-wise causal attention and initialize a $4$-step causal generator through regression on $16$K ODE solution trajectories obtained from the Stage-1 bidirectional teacher ($3$K iterations, $\mathrm{lr}=5\times10^{-6}$).
We subsequently apply distribution matching distillation (DMD)~\cite{dmd} on $5\text{s}$ self-generated autoregressive rollouts~\cite{huang2026self} for $2$K generator updates, using learning rates of $2\times10^{-6}$ and $4\times10^{-7}$ for the generator and fake-score critic, respectively.
3) \textit{Memory-augmented streaming long tuning:}
Equipping the causal generator with $C_i = [C^{\mathrm{sink}}, C_i^{\mathrm{mem}}, C_i^{\mathrm{local}}]$, we apply LoRA ($\text{rank}=256$) to perform rolling autoregressive rollouts spanning the full natural duration of multi-event sequences in \datasetname{} (matching the length of each video's ground-truth semantic sequence) for $3\text{K}$ steps (learning rates $1\times10^{-5}$ for the generator and $2\times10^{-6}$ for the critic), supervised segment-wise by the bidirectional teacher via DMD.

Across all three stages, T2V and TI2V training samples are mixed at a $0.7:0.3$ ratio.
For TI2V samples, the initial latent frame is kept noise-free ($t=0$) as clean conditioning context and excluded from the training loss.
Moreover, throughout all stages, before decoding ground-truth semantic tokens into continuous conditioning features, we apply the same codebook-neighbor token perturbation ($p=0.1$, top-$16$ neighbors) as in planner training so that the renderer remains robust to minor token prediction discrepancies at inference.
In Stages 2 and 3, we use a $1:5$ generator-to-critic update ratio.
The real-score bidirectional teacher applies joint text-and-semantic classifier-free guidance (CFG scale $3.5$), while the distilled causal student supports CFG-free inference.
The $4$-step generator uses nominal timestep indices $[1000,750,500,250]$, corresponding to shifted timesteps $[1000,937.5,833.3,625.0]$ under a flow shift of $5.0$.

During interactive inference, we employ Qwen3.5-2B~\cite{qwen2026qwen35-2b} in non-thinking mode as the lightweight prompt rewriter $\mathcal{P}$ in Eq.~(3) of the main paper to aggregate historical instructions into self-contained descriptions $\widetilde{T}^{(m)}$.
Detailed system prompts and analyses of $\mathcal{P}$ are provided in Sec.~\ref{sec:supp_rewriter}.

\section{Complementary Roles of Visual Semantics and Textual Conditioning}
\label{sec:supp_roles}

\begin{figure}[!t]
\centering
\begin{subfigure}{\textwidth}
    \centering
    \includegraphics[width=\linewidth]{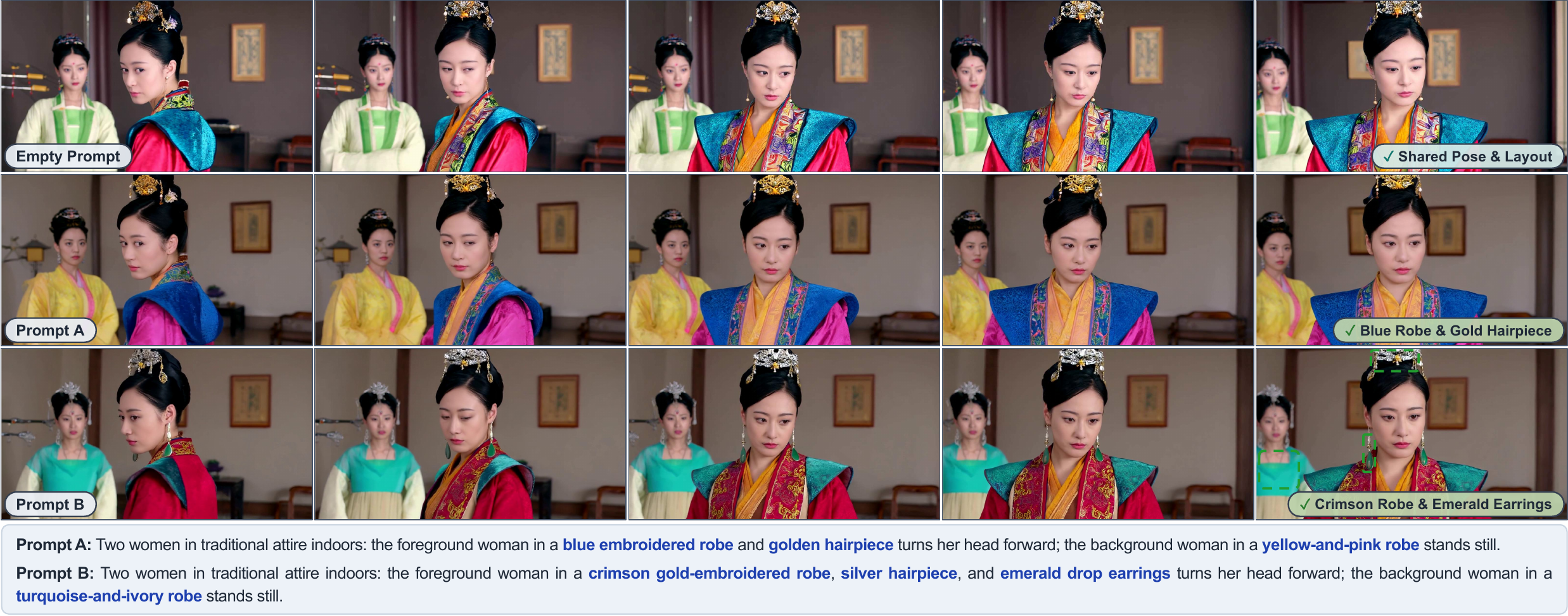}
    \caption{Appearance and Scene Control under Shared Visual Semantics}
    \label{fig:supp_roles_appearance}
\end{subfigure}

\vspace{6pt}

\begin{subfigure}{\textwidth}
    \centering
    \includegraphics[width=\linewidth]{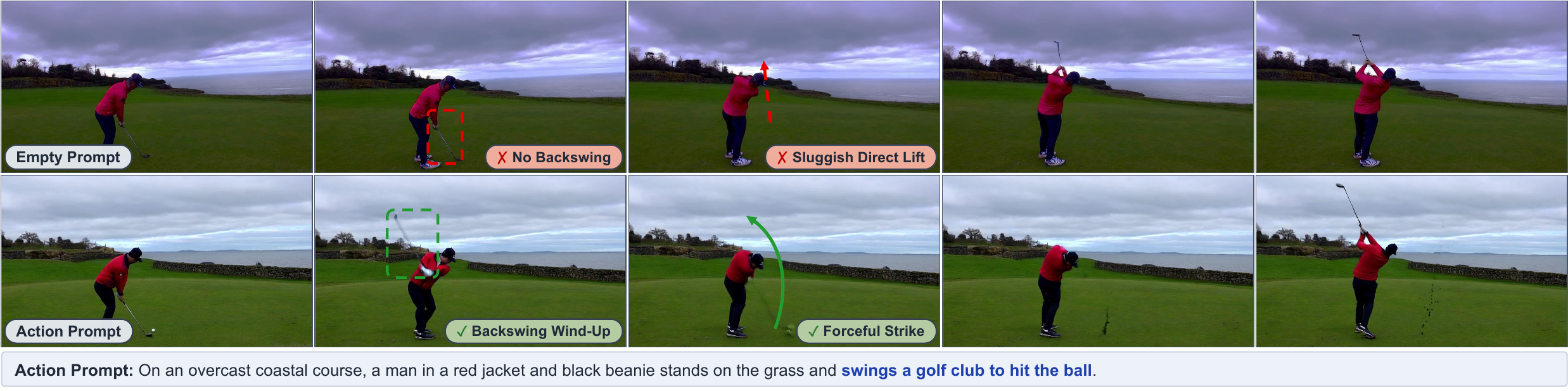}
    \caption{Fine-Grained Motion Refinement through Text Conditioning}
    \label{fig:supp_roles_motion}
\end{subfigure}
\caption{
\textbf{Complementary roles of visual semantics and textual conditioning} under fixed ground-truth semantic sequences.
(a) Varying textual descriptions change subject appearance and scene aesthetics while preserving identical spatial layout and pose evolution.
(b) Explicit action prompts complement $0.5\text{s}$ semantic anchors with fine-grained sub-chunk kinematics and velocity dynamics, enriching anchor-guided pose progression with forceful, realistic motion dynamics.
}
\label{fig:supp_roles_probe}
\end{figure}

\myparagraph{Complementary Conditioning Roles.}
As discussed in Sec.~3.3 of the main paper, our renderer synthesizes each video chunk conditioned jointly on spatially structured visual semantics and textual descriptions.
Fig.~\ref{fig:supp_roles_probe} analyzes their respective roles by fixing the visual semantic sequence and varying the text prompts supplied to the DiT renderer.
Encoded on a $9\times9$ spatial grid at $0.5\text{s}$ intervals, visual semantics govern global spatiotemporal layout, subject pose trajectories, and camera framing.
Meanwhile, textual conditioning complements these coarse semantic representations in two essential aspects:
1) \textit{Appearance and scene details:} As shown in Fig.~\ref{fig:supp_roles_probe}(a), modifying subject or environmental descriptions alters visual attributes such as identity, attire, lighting, and background style, while preserving the underlying spatial composition and motion structure dictated by the visual semantics.
2) \textit{Fine-grained sub-chunk dynamics:} While $0.5\text{s}$ visual semantics reliably establish coherent overall motion trajectories, transient motion details during rapid actions, such as intermediate poses and changes in motion speed, may not be fully specified by adjacent semantic anchors.
Explicit action prompts further supplement these anchors by activating the renderer's motion priors to refine fine-grained kinematics within $0.5\text{s}$ chunks, as illustrated in Fig.~\ref{fig:supp_roles_probe}(b).

\begin{figure}[!ht]
\centering
\includegraphics[width=\textwidth]{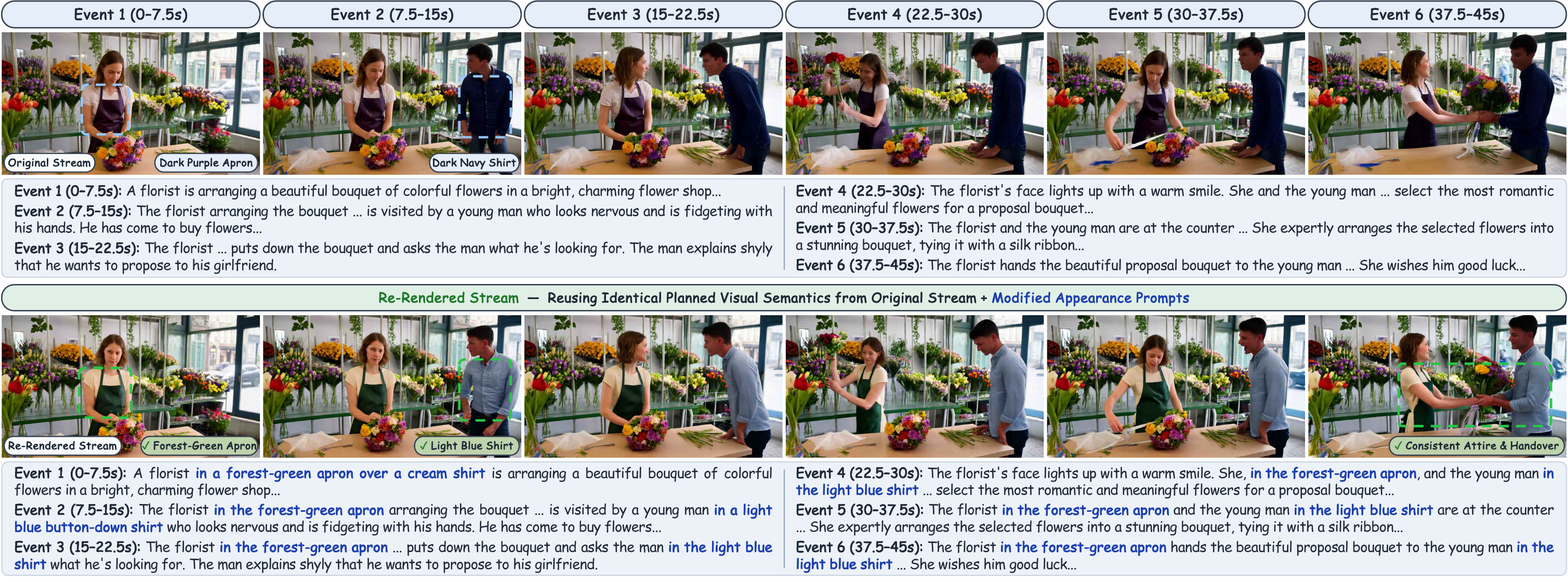}
\caption{
\textbf{Narrative-preserving video re-rendering by reusing planned semantics.}
\textit{Top:} Original $45\text{s}$ ($6\times7.5\text{s}$) interactive video stream generated by \methodname{}.
\textit{Bottom:} Re-rendered multi-event stream conditioned on the same predicted visual semantic sequence with modified text descriptions, customizing subject appearance while maintaining coherent narrative evolution across all six events.
}
\label{fig:supp_rerender}
\end{figure}

\myparagraph{Broader Application: Narrative-Preserving Video Re-Rendering.}
This disentanglement between high-level spatiotemporal semantics and fine-grained visual appearance naturally supports post-hoc visual customization of generated interactive streams.
Once a multi-event narrative is generated, users can retain the predicted semantic trajectory from the autoregressive planner and re-render the video offline under modified text prompts without re-running semantic planning.
As shown in Fig.~\ref{fig:supp_rerender}, updating subject or environmental descriptions flexibly customizes character identity, clothing, and scene style across all events, while preserving the original narrative progression, cross-event transitions, and action choreography.

\section{Semantic Progression: Goal-Directed Actions vs. Ongoing Activities}
\label{sec:supp_progression_types}

In interactive video generation, coherent within-event progression does not necessarily imply reaching a discrete endpoint.
Specifically, user instructions naturally span two primary behavior types:
1) \textit{goal-directed actions} (\eg, sitting down or handing over an object), which naturally evolve toward an identifiable outcome and settle into a plausible post-action state without repetitive execution; and
2) \textit{ongoing activities} (\eg, reading, walking, or organizing items), which sustain continuous, meaningful behavioral evolution without requiring a predefined endpoint.
By learning from diverse visual semantics across real-world videos, our semantic planner supports coherent progression across both behavior types, as demonstrated throughout the multi-event qualitative cases and the variable-duration interactive stream in Fig.~\ref{fig:supp_flexible_horizon}(b).

\section{Flexible-Horizon Planning: Specified-Duration vs. Open-Ended Modes}
\label{sec:supp_flexible}

\begin{figure}[!ht]
\centering
\begin{subfigure}{\textwidth}
    \centering
    \includegraphics[width=\linewidth]{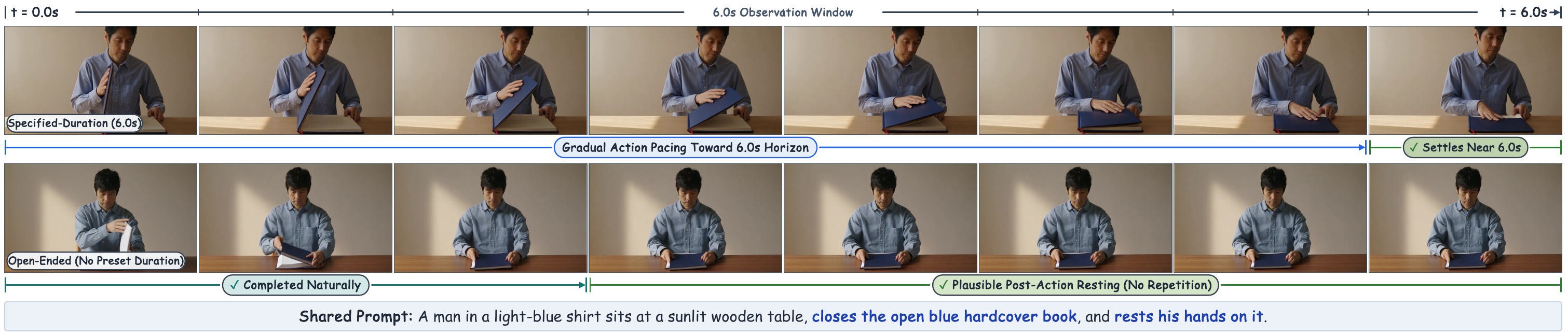}
    \caption{Specified-Duration vs.\ Open-Ended Planning}
    \label{fig:supp_horizon_comparison}
\end{subfigure}

\vspace{6pt}

\begin{subfigure}{\textwidth}
    \centering
    \includegraphics[width=\linewidth]{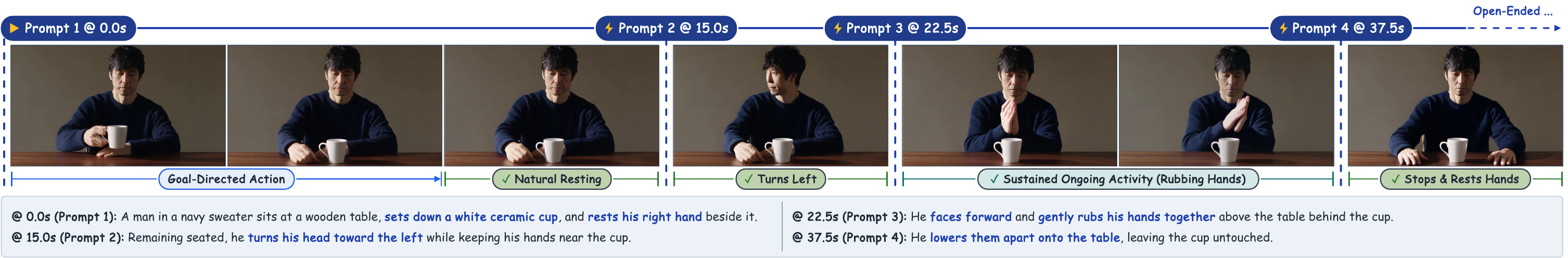}
    \caption{Open-Ended Variable-Duration Interactive Streaming}
    \label{fig:supp_variable_interaction}
\end{subfigure}
\caption{
\textbf{Flexible-horizon semantic planning and open-ended variable-duration interaction.}
(a) Given the same goal-directed prompt over a $6\text{s}$ window, specified-duration planning paces action progression toward the target horizon, whereas open-ended generation completes the action at its intrinsic pace and settles into a plausible post-action state without repetition.
(b) Open-ended multi-event interaction across variable event durations with instructions issued at arbitrary timestamps, supporting both sustained ongoing activities and goal-directed actions.
}
\label{fig:supp_flexible_horizon}
\end{figure}

\myparagraph{Two Flexible-Horizon Planning Modes.}
As described in Sec.~\ref{sec:supp_impl_planner}, our semantic planner supports two planning modes depending on whether an explicit target event duration $D^{(m)}$ is included in the input sequence:
1) \textit{Specified-duration mode:} In this mode, the planner receives an explicit target duration in advance and can adjust the pacing of semantic progression toward the specified horizon.
This mode facilitates duration-controlled multi-event video creation and standardized evaluation with prescribed event durations, enabling more cohesive narrative pacing across successive events.
2) \textit{Open-ended mode:} In this mode, no target duration is specified in advance, allowing semantic progression to continue naturally at its intrinsic pace until a new user instruction arrives.
This mode facilitates free-form interactive streaming where users issue new instructions dynamically at arbitrary moments.
As shown in Fig.~\ref{fig:supp_flexible_horizon}(a), specified-duration planning can distribute the stages of a goal-directed action toward the target horizon, whereas open-ended planning may complete the action earlier and maintain a plausible post-action state thereafter.

\myparagraph{Open-Ended Planning under Fixed-Interval Evaluation.}
Standardized benchmark evaluations of interactive video generation adopt fixed-interval protocols for prompt switching~\cite{iamflow,longlive,ji2025memflow}.
We therefore use specified-duration planning by default to align semantic progression with the prescribed segment lengths.
To examine performance without advance duration information, we further evaluate \methodname{} in open-ended mode by omitting duration conditioning $D^{(m)}$ from the planner's input sequence while retaining the identical $45\text{s}$ ($6{\times}7.5\text{s}$) evaluation protocol of Tab.~1 in the main paper.

\input{tables/supp_comp_open_ended}

As shown in Tab.~\ref{tab:supp_open_ended}, both variants of our method maintain performance comparable to their specified-duration counterparts, with consistent advantages over existing streaming baselines.
We partly attribute the modest decreases in VLM Alignment, Cross-Event Transition, and Within-Event Progression to a mismatch between natural event pacing and fixed-interval prompt switching, which may interrupt complex actions before completion and complicate subsequent event transitions.
Nevertheless, both variants still outperform the strongest baseline by over 10 points on each of these three metrics, demonstrating that the benefits of autoregressive semantic planning are largely retained without explicit duration conditioning.

\myparagraph{Open-Ended Variable-Duration Interaction.}
Unlike fixed-interval benchmark evaluations, practical open-ended interaction allows users to introduce new instructions at arbitrary times based on the evolving visual state, either after observing the desired progress or by intentionally interrupting ongoing actions.
Fig.~\ref{fig:supp_flexible_horizon}(b) demonstrates such open-ended, variable-duration interactive generation.
Throughout the stream, the planner maintains coherent progression within each event, including natural continuation after goal completion and sustained evolution during ongoing activities.

\FloatBarrier
\section{Evaluation Protocols and Coherence Metric Validation}
\label{sec:supp_eval_protocol}

\myparagraph{Evaluation Standardization.}
To ensure fair metric computation across models with different native output formats, all generated videos are spatially resized to 480p and temporally resampled to 16 fps prior to automatic evaluation on NarraStream-Bench, VBench, and VBench-Long.

\myparagraph{Semantic Coherence Evaluation Protocol.}
In Tab.~1 of the main paper, official NarraStream-Bench metrics such as Boundary Smoothness, Conditional Adjacent Consistency, Dynamic Trajectory, and VLM Alignment capture important aspects of transition smoothness, motion dynamics, and text-video alignment.
To complement these metrics with a direct assessment of high-level narrative logic, we follow the VLM-based evaluation paradigm of NarraStream-Bench~\cite{iamflow} and employ Gemini 3.1 Pro~\cite{google2026gemini31} to evaluate semantic coherence in \textit{Cross-Event Transition} and \textit{Within-Event Progression}.
Given a generated video spanning $M$ contiguous event segments, we uniformly sample frames at 4 fps from each segment to represent its temporal evolution.

Specifically, semantic coherence in Cross-Event Transition is evaluated over each adjacent segment pair from the second event onward.
Conditioned on the sampled frames of two consecutive segments alongside historical instructions and the incoming instruction, the VLM judge assesses whether the new event fulfills the updated instruction while maintaining semantic coherence with the established visual state of the preceding segment.
Semantic coherence in Within-Event Progression is evaluated on each individual segment against its corresponding instruction, assessing whether instructed behaviors unfold coherently without partial execution, abrupt discontinuities, or unnatural action looping.
Both metrics assign scores on a 0--100 scale using the prompt templates and five-tier rubrics detailed below.
Segment-level scores are averaged across all $M-1$ transitions or $M$ segments within each video and then averaged across all benchmark sequences.

\vspace{4pt}
\noindent
\begin{tcolorbox}[
    enhanced,
    breakable,
    skin first=enhanced,
    skin middle=enhanced,
    skin last=enhanced,
    pad at break*=4pt,
    colback=black!2,
    colframe=black!55,
    colbacktitle=black!12,
    coltitle=black,
    fonttitle=\bfseries\small,
    boxrule=0.5pt,
    arc=2.5pt,
    left=6pt, right=6pt, top=5pt, bottom=5pt,
    title={(a) Prompt Template and Rubric for Semantic Coherence in Cross-Event Transition}
]
\footnotesize
\textbf{System Prompt:} You are an expert evaluator for interactive multi-event video generation. Given uniformly sampled frames from two consecutive video segments alongside historical instructions and the new instruction for the second segment, evaluate the cross-event semantic coherence by assessing whether the new event fulfills the incoming instruction while transitioning plausibly and naturally from the preceding visual state.

\vspace{2.5pt}
\textbf{Guidelines:}
\begin{itemize}[leftmargin=1.2em, itemsep=0.5pt, topsep=1pt, parsep=0pt]
    \item Verify whether the transition respects established subject positions, postures, and held or surrounding objects at the end of the preceding segment rather than hallucinating duplicate entities.
    \item Assess whether subjects and scene states bridge into the new event through physically and causally plausible motions without abrupt posture resets or unrequested scene jumps.
    \item Check whether the newly instructed event is genuinely initiated and integrated with the prior visual context rather than ignored.
\end{itemize}

\vspace{2.5pt}
\textbf{Scoring Rubric (0--100):}
\begin{itemize}[leftmargin=1.2em, itemsep=0.5pt, topsep=1pt, parsep=0pt]
    \item \textbf{81--100:} Seamless, context-grounded transition; subjects, objects, and poses evolve naturally from the prior ending state to fulfill the new instruction.
    \item \textbf{61--80:} Coherent transition overall; respects prior visual context and fulfills the instruction with only minor motion stiffness or transition artifacts.
    \item \textbf{41--60:} Partially coherent transition; responds to the instruction but exhibits noticeable posture resets or weak grounding in prior object states.
    \item \textbf{21--40:} Incoherent transition; disregards prior context by spawning duplicate entities, causing abrupt state jumps, or executing only partial fragments.
    \item \textbf{0--20:} Severe coherence failure; ignores the incoming instruction entirely or exhibits such severe discontinuities that the event transition becomes unrecognizable.
\end{itemize}

\vspace{2.5pt}
\textbf{User Input:}\\
Previous segment: \texttt{[Segment m-1 frames (4 fps)]} \textbar\ Current segment: \texttt{[Segment m frames (4 fps)]}\\
Historical instructions: $T^{(1:m-1)}$ \textbar\ New instruction: $T^{(m)}$

\vspace{2pt}
\textbf{Output Format:}\\
\texttt{\{"reasoning": "<concise analysis of context grounding and transition>", "score": <int 0-100>\}}
\end{tcolorbox}

\vspace{4pt}
\noindent
\begin{tcolorbox}[
    enhanced,
    breakable,
    skin first=enhanced,
    skin middle=enhanced,
    skin last=enhanced,
    pad at break*=4pt,
    colback=black!2,
    colframe=black!55,
    colbacktitle=black!12,
    coltitle=black,
    fonttitle=\bfseries\small,
    boxrule=0.5pt,
    arc=2.5pt,
    left=6pt, right=6pt, top=5pt, bottom=5pt,
    title={(b) Prompt Template and Rubric for Semantic Coherence in Within-Event Progression}
]
\footnotesize
\textbf{System Prompt:} You are an expert evaluator for video semantic dynamics. Given uniformly sampled frames from an individual video segment and its text instruction, evaluate the within-event semantic coherence by assessing whether the instructed behaviors and interactions unfold progressively, naturally, and logically across the segment duration.

\vspace{2.5pt}
\textbf{Guidelines:}
\begin{itemize}[leftmargin=1.2em, itemsep=0.5pt, topsep=1pt, parsep=0pt]
    \item Assess whether goal-directed actions advance through ordered stages toward intended outcomes and settle appropriately when completed, and whether ongoing activities sustain continuous, meaningful behavioral evolution.
    \item Penalize unnatural action repetition, mechanical back-and-forth looping of one-off actions, or premature freezing after partial keyword execution.
    \item Verify spatiotemporal integrity within the segment, penalizing abrupt motion discontinuities or duplicate subject spawning.
\end{itemize}

\vspace{2.5pt}
\noindent\begin{minipage}{\linewidth}
\textbf{Scoring Rubric (0--100):}
\begin{itemize}[leftmargin=1.2em, itemsep=0.5pt, topsep=1pt, parsep=0pt]
    \item \textbf{81--100:} Highly coherent progression; goal-directed actions advance naturally toward intended outcomes and settle appropriately when completed, while ongoing activities evolve meaningfully without repetition.
    \item \textbf{61--80:} Mostly coherent progression; unfolds logically overall with only minor pacing unevenness or brief motion hesitations.
    \item \textbf{41--60:} Stalled or repetitive progression; executes only partial actions, freezes prematurely, or exhibits unnatural action looping.
    \item \textbf{21--40:} Incoherent within-event dynamics; suffers from abrupt motion jumps, duplicate subject spawning, or implausible action trajectories.
    \item \textbf{0--20:} Severe coherence failure; the instructed event fails to occur or exhibits such severe temporal discontinuities that progression becomes unrecognizable.
\end{itemize}
\end{minipage}

\vspace{2.5pt}
\textbf{User Input:}\\
Current segment: \texttt{[Segment m frames (4 fps)]} \textbar\ Segment instruction: $T^{(m)}$

\vspace{2pt}
\textbf{Output Format:}\\
\texttt{\{"reasoning": "<concise analysis of behavioral progression>", "score": <int 0-100>\}}
\end{tcolorbox}
\vspace{4pt}

\myparagraph{Validation Against Human Judgments.}
To verify that both automatic coherence metrics align with human perception, we conduct a dedicated human alignment study on 30 randomly sampled prompt sequences from NarraStream-Bench under the 45s setting.
In our main user study, several streaming baselines exhibit noticeable visual degradation, potentially confounding coherence judgments with large differences in overall visual quality.
To better assess whether these metrics capture semantic coherence across varying levels of planning capability under comparable visual quality, we construct a four-method validation set sharing the 1.3B renderer backbone: 1) the streaming baseline \textit{LongLive}, 2) our planner-free variant \textit{w/o planner}, 3) our reasoning-free variant \textit{w/o CoT}, and 4) \textit{Ours-1.3B}.

In a separate evaluation session following the same participant qualifications, informed consent procedures, ethical compliance, and compensation standards described in Sec.~\ref{sec:supp_user_study}, 30 independent participants blindly inspect the four anonymized videos for each sequence and provide separate strict rankings without ties for semantic coherence in cross-event transition and within-event progression.
Following the pairwise preference analysis of IAMFlow~\cite{iamflow}, we aggregate the 30 participant rankings into human pairwise preference rates for all six model pairs per sequence and compare them with the corresponding automatic VLM score differences, yielding 180 pairwise comparisons per metric.
We report Spearman's rank correlation $\rho$ between human preference rates and automatic score differences, alongside pairwise agreement with majority human preferences.
As shown in Tab.~\ref{tab:supp_metric_val}, both metrics exhibit strong agreement with human judgments, supporting their effectiveness for evaluating multi-event semantic coherence across baselines and ablations.

\begin{table}[!ht]
\centering
\caption{
\textbf{Alignment between automatic semantic coherence metrics and human judgments.}
Evaluated across 30 randomly sampled 45s sequences (180 model-pair comparisons per metric) over \textit{LongLive}, \textit{w/o planner}, \textit{w/o CoT}, and \textit{Ours-1.3B}.
}
\label{tab:supp_metric_val}
\small
\setlength{\tabcolsep}{12pt}
\begin{tabular}{lcc}
\toprule
Semantic Coherence Metric & Spearman's $\rho$ $\uparrow$ & Pairwise Agreement (\%) $\uparrow$ \\
\midrule
Cross-Event Transition & 0.81 & 85.56 \\
Within-Event Progression & 0.78 & 82.78 \\
\bottomrule
\end{tabular}
\end{table}

\section{Complete Ablation Study}
\label{sec:supp_ablation}

All ablation variants are built upon the Wan2.1-T2V-1.3B renderer and evaluated on NarraStream-Bench~\cite{iamflow} under the identical $45\text{s}$ ($6\times7.5\text{s}$) multi-prompt streaming protocol.
Tab.~\ref{tab:supp_ablation_full} reports the complete quantitative ablation results.
Beyond the variants summarized in Tab.~3 of the main paper, we provide two finer-grained breakdowns:
1) we evaluate the planner-free variant under both raw concise instructions (\textit{w/o planner (raw prompts)}) and rewritten structured captions (\textit{w/o planner (rewritten prompts)}, corresponding to \textit{w/o planner} in Tab.~3 of the main paper) to disentangle the effect of prompt rewriting from autoregressive semantic planning; and
2) we decompose historical visual context into \textit{w/o memory $C_i^{\mathrm{mem}}$} and \textit{w/o memory \& sink} to examine the respective contributions of retrieved long-range memory and the initial sink anchor.

\input{tables/supp_ablation_full}

\myparagraph{Autoregressive Semantic Planning.}
To evaluate streaming generation without semantic planning, we remove semantic cross-attention and train a planner-free renderer on \datasetname{} following the identical three-stage pipeline: Stage-1 bidirectional training on randomly cropped $5\text{s}$ multi-event clips conditioned on text prompts without visual semantic features (exposing the model to cross-chunk prompt transitions while alternating between concise instructions $T^{(m)}$ and structured captions $\widetilde{T}^{(m)}$), Stage-2 causal ODE initialization and DMD distillation, and Stage-3 streaming long tuning.
Because planned visual semantics $S_i$ are unavailable in this setting, to ensure a fair comparison without disadvantaging the planner-free variant in long-range visual memory, we equip it with a text-guided memory retrieval mechanism following MemFlow~\cite{ji2025memflow}: textual tokens of the active prompt query historical chunk KV caches via cross-attention relevance to select $C_i^{\mathrm{mem}}$ over the exact same candidate pool $\{V_2, \ldots, V_{i-3}\}$ as our full model.
At inference, \textit{w/o planner (rewritten prompts)} receives structured captions $\widetilde{T}^{(m)}$ from $\mathcal{P}$, whereas \textit{w/o planner (raw prompts)} feeds raw user instructions $T^{(m)}$ directly into the same renderer.

As shown in Tab.~\ref{tab:supp_ablation_full}, removing the semantic planner causes severe degradation in instruction compliance and both semantic coherence metrics regardless of prompt formatting.
In particular, \textit{w/o planner (raw prompts)} and \textit{w/o planner (rewritten prompts)} exhibit similarly weak cross-event transition ($57.13$ and $57.11$) and within-event progression ($62.58$ and $63.85$).
This confirms that the absence of explicit semantic evolution planning, rather than prompt brevity, is the fundamental bottleneck of reactive causal DiTs.

\myparagraph{External Textual Reasoning.}
To examine whether an external VLM providing explicit reasoning and planning guidance in textual form can substitute for autoregressive visual semantic planning under a matched model parameter budget ($7\text{B}+1.3\text{B}$), we evaluate \textit{w/o planner + external reasoning}.
Specifically, we adapt the planner-free renderer by incorporating textual descriptions combining self-contained entity appearance with explicit event transition and evolution cues ($\widetilde{T}^{(m)}$ augmented with $R^{(m)}$).
At each prompt switch ($m\ge2$) during inference, Qwen2.5-VL-7B-Instruct~\cite{qwen25vl} serves as an external reasoning module that inspects the preceding $7.5\text{s}$ generated video segment alongside historical instructions and the new instruction $T^{(m)}$, producing an enriched prompt with explicit transition and evolution directives.
The prompt template and a representative reasoning example for this external VLM baseline are detailed below.

\input{figures/supp_external_vlm_prompt}

Despite operating under this matched parameter budget, \textit{+ external reasoning} yields only marginal gains over \textit{w/o planner (rewritten prompts)} ($58.62$ vs.\ $57.11$ in cross-event transition; $65.55$ vs.\ $63.85$ in within-event progression) and still lags far behind Ours (full) ($78.59$ and $81.68$).
This demonstrates that our performance gains stem from the decoupled visual semantic planning paradigm rather than extra model capacity: text-level reasoning alone remains too abstract for a causal DiT renderer to reliably translate into coherent frame-level dynamics, underscoring the necessity of planning directly in a spatially structured visual semantic space.

\myparagraph{Context-Aware CoT.}
For \textit{w/o CoT}, we train the semantic planner without intermediate event-change reasoning (\texttt{<think>}$R^{(m)}$\texttt{</think>}) or reasoning supervision, directly predicting visual semantic blocks $\mathcal{B}(S_i)$ after each incoming instruction $T^{(m)}$ while keeping the semantic-guided renderer unchanged.
Even without CoT, autoregressive visual semantic planning substantially improves instruction compliance ($72.93$ vs.\ $66.50$), cross-event transition ($74.06$ vs.\ $57.11$), and within-event progression ($81.34$ vs.\ $63.85$) over \textit{w/o planner (rewritten prompts)} as well as existing streaming baselines.
Incorporating context-aware CoT in Ours (full) further boosts cross-event transition from $74.06$ to $78.59$ and instruction compliance from $72.93$ to $73.78$ by explicitly resolving cross-event state changes before predicting subsequent visual semantics.

\myparagraph{Visual Context and Memory Retrieval.}
To dissect the contribution of each visual context component and evaluate alternative memory retrieval mechanisms, we perform streaming long tuning of the semantic-guided renderer under four context configurations while keeping the semantic planner fixed:
1) \textit{w/o memory \& sink} retains only the local sliding window $C_i=[C_i^{\mathrm{local}}]$ ($w=3$);
2) \textit{w/o memory $C_i^{\mathrm{mem}}$} retains the initial sink and local window $C_i=[C^{\mathrm{sink}}, C_i^{\mathrm{local}}]$;
3) \textit{random memory} samples $C_i^{\mathrm{mem}}$ uniformly from $\{V_2, \ldots, V_{i-3}\}$; and
4) \textit{text-based retrieval} replaces semantic cosine similarity in Eq.~(5) of the main paper with the text-query cross-attention relevance scoring of~\cite{ji2025memflow} over $\{V_2, \ldots, V_{i-3}\}$.

Comparing the two memory-free variants reveals clear complementary roles.
Removing both $C_i^{\mathrm{mem}}$ and $C^{\mathrm{sink}}$ (\textit{w/o memory \& sink}) leaves only a short local window, causing severe late-horizon degradation in visual quality ($78.92$) and temporal consistency ($67.38$). The resulting subject and background distortions in later segments also indirectly impair semantic coherence ($74.63$ in cross-event transition and $73.11$ in within-event progression).
Retaining $C^{\mathrm{sink}}$ (\textit{w/o memory $C_i^{\mathrm{mem}}$}) stabilizes early global attributes and prevents late-horizon collapse, substantially recovering overall performance (reaching $77.98$ and $80.66$ in semantic coherence), yet temporal consistency ($70.04$ vs.\ $76.95$) and visual quality ($79.43$ vs.\ $83.39$) still drop markedly compared to Ours (full) whenever subjects, poses, or scene views emerge after the initial chunk or deviate substantially from it.

Among memory-enabled variants, \textit{random memory} ($80.69$ in visual quality; $70.13$ in temporal consistency) often recalls irrelevant historical chunks.
Meanwhile, comparing \textit{w/o planner (rewritten prompts)} and \textit{text-based retrieval}, which share identical rewritten prompts and text-based retrieval, further isolates the large gains brought by our semantic planner ($+21.89$ in cross-event transition and $+17.62$ in within-event progression).
However, because semantic planning drives substantial pose, action, and camera evolution within each $7.5\text{s}$ segment (unlike near-static planner-free generation), relying on a single static segment-level prompt in \textit{text-based retrieval} ($80.71$ and $72.59$) is too coarse to match evolving visual states.
Replacing segment-level text retrieval with our fine-grained chunk-level semantic-guided retrieval in Ours (full) resolves this mismatch, achieving the highest visual quality ($83.39$) and temporal consistency ($76.95$) alongside strong narrative coherence ($78.59$ and $81.68$).

\myparagraph{Built-in Prompt Rewriter.}
To examine the role of the built-in prompt rewriter $\mathcal{P}$, we evaluate \textit{w/o prompt rewriter} by bypassing $\mathcal{P}$ at inference and feeding raw concise instructions $T^{(m)}$ directly into the renderer while preserving semantic planning and semantic-guided retrieval.
Because our renderer is trained by randomly alternating between concise instructions $T^{(m)}$ and structured captions $\widetilde{T}^{(m)}$, it natively accommodates concise prompts without train-test format mismatch.
In our framework, $\mathcal{P}$ is designed around a clear functional division, as analyzed in detail in Sec.~\ref{sec:supp_roles}: planned visual semantics $S_i$ govern spatiotemporal layout, action evolution, and narrative transitions, whereas text conditioning primarily supplies entity appearance and fine-grained visual details to the renderer.
Consistent with this design, removing $\mathcal{P}$ leaves cross-event transition ($78.55$ vs.\ $78.59$), within-event progression ($82.19$ vs.\ $81.68$), and instruction compliance ($72.35$ vs.\ $73.78$) virtually unchanged, while still outperforming existing streaming baselines in Tab.~1 of the main paper in visual quality ($82.59$) and most baselines in temporal consistency ($73.43$).
Together with the planner-free comparisons above, this confirms that our primary gains stem from autoregressive semantic planning rather than prompt rewriting. $\mathcal{P}$ specifically serves our decoupled design by consolidating cross-turn appearance cues for the renderer, further lifting temporal consistency to $76.95$ and visual quality to $83.39$.
To further verify this, we provide a controlled study supplying shared structured captions $\widetilde{T}^{(m)}$ to all baseline methods in Sec.~\ref{sec:supp_rewriter}.

\section{Prompt Rewriting and Evaluation with Shared Structured Captions}
\label{sec:supp_rewriter}

\myparagraph{Evaluation Protocol Rationale.}
In practical multi-turn interactive video generation, users naturally issue concise follow-up instructions $T^{(m)}$ ($m\ge2$) that specify new actions or events without repeatedly restating established subject appearances and scene settings.
Accordingly, for a fair end-to-end comparison, Tab.~1 of the main paper evaluates complete generation pipelines under identical raw user instructions $T^{(1:M)}$, with all generated videos assessed against the original instructions.
Within \methodname{}, the lightweight prompt rewriter $\mathcal{P}$ serves as a built-in component tailored to our framework, consolidating user instructions $T^{(1:m)}$ across turns into self-contained descriptions $\widetilde{T}^{(m)}$ without introducing extra semantic content or accessing privileged information.
Moreover, for all single-prompt benchmarks, including VBench (Tab.~2 of the main paper) and VBench-Long (Tab.~\ref{tab:supp_vbench_long} in Sec.~\ref{sec:supp_single_prompt}), where prompts are already self-contained and involve no multi-turn omissions, we do not invoke $\mathcal{P}$ and evaluate our models and all baselines under the same extended prompts provided by~\cite{huang2026self,huang2024vbench}.

\input{tables/supp_comp_shared_captions}

\myparagraph{Controlled Comparison with Shared Structured Captions.}
To further rule out the possibility that our interactive generation gains stem from richer prompt phrasing rather than autoregressive semantic planning, we conduct a controlled experiment on NarraStream-Bench ($45\text{s}$) by supplying the exact same rewritten structured captions $\widetilde{T}^{(1:6)}$ generated by $\mathcal{P}$ directly to all baseline methods.
Consistent with Tab.~1 of the main paper, all generated videos are evaluated against the original instructions $T^{(1:6)}$.
Tab.~\ref{tab:supp_comp_shared_captions} reports the complete sub-metric results, and Tab.~\ref{tab:supp_comp_shared_summary} summarizes the aggregated dimension scores alongside their changes ($\Delta$) relative to raw instructions.

Comparing Tab.~\ref{tab:supp_comp_shared_captions} and Tab.~\ref{tab:supp_comp_shared_summary} with Tab.~1 of the main paper reveals a clear pattern.
Explicitly restating cross-turn subject appearances and scene settings in $\widetilde{T}^{(m)}$ predictably helps existing streaming baselines recover static entity appearance in later segments, moderately improving instruction compliance on static entity alignment rather than dynamic action execution, yet still leaving them clearly behind our method.
Meanwhile, for competitive streaming baselines, overall visual quality and temporal consistency exhibit little improvement or minor fluctuations, and crucially, semantic coherence in both cross-event transition and within-event progression remains largely unchanged and substantially inferior to our method.

Notably, the effect of prompt rewriting differs somewhat within \methodname{} (Tab.~\ref{tab:supp_ablation_full} and Tab.~\ref{tab:supp_comp_shared_summary}): because our semantic planner already anchors instruction compliance and semantic coherence, $\mathcal{P}$ primarily improves temporal consistency and visual quality under the complementary division of roles analyzed in Sec.~\ref{sec:supp_roles}.
Overall, these results confirm that rewriting prompts into self-contained descriptions with carried-forward static appearance and scene cues alone brings little benefit to semantic coherence in cross-event transitions or within-event progression; rather, these improvements stem from our autoregressive visual semantic planning.

\myparagraph{Prompt Rewriter Specification.}
As introduced in Sec.~\ref{sec:supp_impl_renderer}, we implement $\mathcal{P}$ using Qwen3.5-2B~\cite{qwen2026qwen35-2b} in non-thinking mode.
Operating solely on textual instructions $T^{(1:m)}$ without visual inputs, $\mathcal{P}$ resolves coreferences across turns and carries forward established subject appearances and scene settings into a self-contained structured caption $\widetilde{T}^{(m)}$, while strictly discarding completed past actions and avoiding hallucinated attributes.
The system prompt template and a representative rewriting example are presented below, illustrating how subject identity, appearance, and environment are carried forward into $\widetilde{T}^{(m)}$ while restricting the action description to the current instruction.

\input{figures/supp_rewriter_prompt}

\section{Additional Evaluation of Interactive Video Generation}
\label{sec:supp_interactive_eval}

\begin{table}[!ht]
\centering
\caption{
\textbf{Quantitative comparison on the 100 multi-prompt scripts from MemFlow (45s, 6\texttimes7.5s).}
Quality Score and Consistency Score (averaging subject and background consistency) from VBench-Long are measured over the full video, while CLIP Score is computed over each $7.5\text{s}$ prompt interval against the corresponding original prompt.
Best and second-best scores are \textbf{bolded} and \uline{underlined}.
}
\label{tab:supp_memflow_scripts}
\setlength{\tabcolsep}{4.5pt}
\resizebox{0.7\textwidth}{!}{%
\begin{tabular}{@{}lcccccccc@{}}
\toprule
\multirow{2}{*}{Model}
& \multirow{2}{*}{\shortstack{\strut Quality\\\strut Score $\uparrow$}}
& \multirow{2}{*}{\shortstack{\strut Consistency\\\strut Score $\uparrow$}}
& \multicolumn{6}{c}{CLIP Score $\uparrow$} \\
\cmidrule(lr){4-9}
& & & $0$--$7.5\text{s}$ & $7.5$--$15\text{s}$ & $15$--$22.5\text{s}$ & $22.5$--$30\text{s}$ & $30$--$37.5\text{s}$ & $37.5$--$45\text{s}$ \\
\midrule
Self-Forcing
& 82.65 & 94.07 & 29.16 & 25.90 & 25.33 & 23.89 & 22.50 & 21.04 \\
Rolling Forcing
& 83.26 & 95.31 & 28.06 & 25.40 & 24.96 & 24.66 & 24.53 & 23.76 \\
Infinity-RoPE
& 84.02 & 96.42 & 27.87 & 25.54 & 25.65 & 24.92 & 25.02 & 24.73 \\
LongLive
& 83.52 & 96.39 & 28.84 & 26.73 & 26.77 & 26.23 & 26.34 & 26.33 \\
MemFlow
& 82.96 & 96.38 & 28.95 & 26.58 & 25.89 & 25.39 & 25.50 & 25.41 \\
Anchor Forcing
& 83.60 & 95.43 & 28.52 & 26.23 & 25.97 & 25.49 & 25.53 & 25.18 \\
IAMFlow
& 83.13 & 96.54 & 29.02 & 26.10 & 26.13 & 25.74 & 25.91 & 25.63 \\
\midrule
\rowcolor{OursBlue}
\textbf{Ours-1.3B}
& \uline{84.93} & \textbf{96.84}
& \uline{31.62} & \uline{30.46} & \uline{31.24}
& \uline{30.97} & \uline{30.87} & \uline{30.21} \\
\rowcolor{OursBlue}
\textbf{Ours-5B}
& \textbf{85.07} & \uline{96.78}
& \textbf{32.20} & \textbf{31.01} & \textbf{31.55}
& \textbf{31.18} & \textbf{31.08} & \textbf{30.38} \\
\bottomrule
\end{tabular}%
}
\end{table}

\begin{table}[!ht]
\centering
\caption{
\textbf{Quantitative comparison on NarraStream-Bench under the native 10s per-event duration (4\texttimes10s, 40s total).}
Best and second-best scores are \textbf{bolded} and \uline{underlined}.
}
\label{tab:supp_narrastream_4x10}
\setlength{\tabcolsep}{2.8pt}
\resizebox{\textwidth}{!}{%
\begin{tabular}{@{}l*{13}{c}@{}}

\toprule
\multirow{2}{*}{Model}
& \multicolumn{5}{c}{Visual Quality $\uparrow$}
& \multicolumn{3}{c}{Temporal Consistency $\uparrow$}
& \multicolumn{3}{c}{Instruction Compliance $\uparrow$}
& \multicolumn{2}{c}{Semantic Coherence $\uparrow$} \\
\cmidrule(lr){2-6}
\cmidrule(lr){7-9}
\cmidrule(lr){10-12}
\cmidrule(lr){13-14}
& \shortstack{\strut Subject\\\strut Consist.}
& \shortstack{\strut Background\\\strut Consist.}
& \shortstack{\strut Temporal\\\strut Flicker.}
& \shortstack{\strut Motion\\\strut Smooth.}
& \shortstack{\strut Temporal\\\strut Stability}
& \shortstack{\strut Boundary\\\strut Smooth.}
& \shortstack{\strut Adjacent\\\strut Consist.}
& \shortstack{\strut Long-range\\\strut Consist.}
& \shortstack{\strut Entity\\\strut Ground.}
& \shortstack{\strut Dynamic\\\strut Traj.}
& \shortstack{\strut VLM\\\strut Align.}
& \shortstack[c]{\strut Cross-Event\\\strut Transition}
& \shortstack[c]{\strut Within-Event\\\strut Progression} \\
\midrule
Self-Forcing
& 84.57 & 83.40 & 87.92 & 59.81 & 29.64 & 48.77 & 58.06 & 45.91 & 61.86 & 55.71 & 52.58 & 46.52 & 47.05 \\
Rolling Forcing
& 92.10 & 90.07 & 85.35 & 49.23 & 68.49 & 40.06 & 76.39 & 61.30 & 67.48 & 59.10 & 62.86 & 53.54 & 51.45 \\
Infinity-RoPE
& 92.38 & 91.56 & 88.80 & 53.38 & 72.46 & 50.27 & 75.95 & 80.80 & 70.44 & 59.65 & 62.98 & 50.65 & 52.17 \\
LongLive
& 94.68 & 92.53 & 91.38 & 57.98 & 74.95 & 49.36 & 83.65 & 89.63 & 70.67 & 63.25 & 63.71 & 50.54 & 60.00 \\
MemFlow
& \uline{95.49} & 92.22 & 91.99 & 56.46 & 73.01 & 44.48 & 83.36 & 88.89 & 69.91 & 64.02 & 66.00 & 57.97 & 65.14 \\
Anchor Forcing
& 91.54 & 89.41 & 85.33 & 39.08 & 70.83 & 55.64 & 68.37 & 75.09 & 70.93 & 58.14 & 65.81 & 59.26 & 64.01 \\
IAMFlow
& 95.37 & 92.51 & 91.72 & 57.45 & 74.92 & 47.94 & 85.45 & \uline{90.40} & 69.72 & 63.54 & 59.97 & 57.73 & 65.39 \\
\midrule
\rowcolor{OursBlue}
\textbf{Ours-1.3B}
& 95.17 & \textbf{93.06} & \uline{92.30} & \uline{61.50} & \uline{75.15} & \uline{55.65} & \uline{85.65} & 89.61 & \uline{76.45} & \textbf{65.65} & \uline{80.64} & \uline{76.64} & \uline{79.65} \\
\rowcolor{OursBlue}
\textbf{Ours-5B}
& \textbf{95.53} & \uline{92.69} & \textbf{92.97} & \textbf{62.46} & \textbf{75.58} & \textbf{55.93} & \textbf{85.67} & \textbf{90.44} & \textbf{77.23} & \uline{64.85} & \textbf{81.77} & \textbf{78.74} & \textbf{84.75} \\
\bottomrule
\end{tabular}%
}
\end{table}

\myparagraph{Evaluation on MemFlow Multi-Prompt Scripts.}
To align with the evaluation setups adopted in representative baselines~\cite{ji2025memflow, longlive, anchorforcing2026} beyond NarraStream-Bench, we evaluate \methodname{} and all baselines on the 100 multi-prompt narrative scripts from MemFlow~\cite{ji2025memflow} under the $45\text{s}$ ($6\times7.5\text{s}$) setting.
Following~\cite{ji2025memflow, longlive}, we evaluate VBench-Long~\cite{huang2025vbench++} dimensions that support customized prompt videos over each full generated sequence, together with segment-wise CLIP scores computed over each prompt interval to measure text-video semantic alignment.

As shown in Tab.~\ref{tab:supp_memflow_scripts}, our method achieves the strongest overall perceptual quality and consistency while maintaining consistently higher segment-wise CLIP alignment across all six prompt intervals, demonstrating minimal semantic degradation over extended multi-turn interactions.

\myparagraph{Evaluation under Native 10s Event Duration (4\texttimes10s).}
We also evaluate all models on NarraStream-Bench~\cite{iamflow} under its native $10\text{s}$ per-event duration across the first four events ($4\times10\text{s}$, $40\text{s}$ in total), which remains within our planner's $10\text{K}$-token context budget while ruling out potential protocol adaptation bias from shortening segment durations.
As shown in Tab.~\ref{tab:supp_narrastream_4x10}, our method outperforms existing baselines across nearly all metrics, especially in instruction compliance and both semantic coherence dimensions, corroborating the findings and conclusions in Tab.~1 of the main paper.
We further explore extending our framework beyond the current $10\text{K}$-token context budget in Sec.~\ref{sec:supp_long_horizon}.

\section{Additional Evaluation of Single-Prompt Video Generation}
\label{sec:supp_single_prompt}

\input{tables/supp_comp_single_prompt}

\myparagraph{Short Video Generation.}
Tab.~\ref{tab:supp_comp_short} expands Tab.~2 of the main paper on $5\text{s}$ VBench~\cite{huang2024vbench} by including additional streaming autoregressive baselines, Deep Forcing~\cite{deepforcing2025} and Helios-Distilled~\cite{helios2026}.
Consistent with the findings in the main paper, our method achieves the highest Total Score and Semantic Score across both non-streaming and streaming models while maintaining real-time streaming throughput, outperforming all streaming baselines by a clear margin in semantic alignment without sacrificing visual quality.

\myparagraph{Long Video Generation.}
We further evaluate $30\text{s}$ single-prompt video generation on VBench-Long~\cite{huang2025vbench++} using official extended prompts~\cite{huang2026self}.
As shown in Tab.~\ref{tab:supp_vbench_long}, both variants of \methodname{} outperform all streaming baselines in Total Score and across nearly all sub-metrics on $30\text{s}$ videos, especially in Semantic Score and Overall Consistency measured by video-text similarity, demonstrating superior semantic alignment and visual coherence over extended durations.

\section{Inference Efficiency and End-to-End System Deployment}
\label{sec:supp_efficiency}

\myparagraph{Steady-State Throughput Breakdown.}
Throughput in Tab.~2 of the main paper and Tab.~\ref{tab:supp_comp_short} is measured on NVIDIA H800 GPUs under model-specific inference configurations in the steady-state generation regime, excluding startup and VAE decoding overhead following~\cite{huang2026self,longlive}.
The Wan backbones and DiT-only streaming baselines (Self-Forcing, Rolling Forcing, Deep Forcing, Infinity-RoPE, Helios-Distilled, LongLive, MemFlow, and Anchor Forcing) are benchmarked on one GPU, covering core DiT denoising and KV-cache updates.
Bernini~\cite{bernini2026} uses its 8-GPU configuration, with throughput calculated from sequential semantic planning and DiT rendering.
IAMFlow~\cite{iamflow} and \methodname{} use two GPUs with asynchronous execution.
For our method, the 7B planner achieves equivalent generation rates of $19.2$ and $28.7\text{ FPS}$ for the 1.3B and 5B variants, respectively, converting semantic token throughput to video FPS at their native frame rates.
The corresponding 4-step DiT renderers achieve $19.9$ and $19.1\text{ FPS}$.
Overlapping both stages yields idealized pipelined throughputs of $19.2$ and $19.1\text{ FPS}$, bounded by the slower stage.
Notably, long-horizon semantic planning enables coherent generation with a compact five-chunk visual context without prompt-switch KV recaching, maintaining high rendering efficiency.

\begin{table}[!ht]
\centering
\caption{
\textbf{End-to-end deployment efficiency on 45s (6\texttimes7.5s) multi-event interactive generation} on two H800 GPUs with streaming TinyVAE~\cite{taehv2025} decoding.
Native Frame Rate specifies the target playback frame rate of generated videos.
E2E Latency measures total wall-clock time from initial prompt input to final CPU frame assembly across all six events.
E2E FPS divides total output frames by E2E Latency, while Chunk FPS follows~\cite{iamflow} to measure steady chunk-loop throughput including synchronization waits.
}
\label{tab:supp_e2e_deploy}
\setlength{\tabcolsep}{5.5pt}
\resizebox{0.74\textwidth}{!}{%
\begin{tabular}{@{}lccccc@{}}
\toprule
Model & Resolution & Native Frame Rate & E2E Latency (s) $\downarrow$ & E2E FPS $\uparrow$ & Chunk FPS $\uparrow$ \\
\midrule
Self-Forcing    & $832\!\times\!480$  & $16\text{ fps}$ & 43.63 & 16.43 & 17.22 \\
Rolling Forcing & $832\!\times\!480$  & $16\text{ fps}$ & 46.36 & 15.47 & 16.10 \\
Deep Forcing    & $832\!\times\!480$  & $16\text{ fps}$ & 42.55 & 16.85 & 17.71 \\
LongLive        & $832\!\times\!480$  & $16\text{ fps}$ & 40.80 & 17.57 & 18.34 \\
MemFlow         & $832\!\times\!480$  & $16\text{ fps}$ & 46.56 & 15.40 & 16.03 \\
Anchor Forcing  & $832\!\times\!480$  & $16\text{ fps}$ & 51.52 & 13.92 & 14.41 \\
IAMFlow         & $832\!\times\!480$  & $16\text{ fps}$ & 80.11 & 8.95  & 10.75 \\
\midrule
\rowcolor{OursBlue}
\textbf{Ours-1.3B} & $832\!\times\!480$  & $16\text{ fps}$ & 43.49 & 16.58 & 17.44 \\
\rowcolor{OursBlue}
\textbf{Ours-5B}   & $1280\!\times\!704$ & $24\text{ fps}$ & 65.41 & 16.53 & 18.00 \\
\bottomrule
\end{tabular}%
}
\end{table}

\myparagraph{End-to-End Interactive Deployment.}
We further evaluate end-to-end system efficiency on complete $45\text{s}$ ($6\times7.5\text{s}$) interactive generation using two NVIDIA H800 GPUs and streaming TinyVAE~\cite{taehv2025} decoding for all methods.
DiT-only streaming baselines run DiT generation on GPU~0 and offload streaming TinyVAE decoding to GPU~1.
IAMFlow uses its asynchronous deployment with FP8 DiT inference and TinyVAE decoding on GPU~0, and separate LLM and VLM agents on GPU~1.
For our method, all modules run in BF16, with GPU~0 hosting the semantic planner and lightweight prompt rewriter while GPU~1 performs text encoding, DiT rendering, and TinyVAE decoding.
Semantic tokens are streamed to the renderer as they become available through a bounded two-chunk queue to overlap planning and rendering.

As shown in Tab.~\ref{tab:supp_e2e_deploy}, Ours-1.3B achieves real-time end-to-end generation at $16.58\text{ E2E FPS}$ in practical deployment, remaining competitive with efficient DiT-only streaming methods.
Ours-5B maintains $16.53\text{ E2E FPS}$ at higher spatial resolution and a native frame rate of $24\text{ fps}$.
IAMFlow incurs additional visual-feedback dependencies, as its VLM verification relies on decoded video frames to update entity memory.
In contrast, our semantic planner and renderer operate in a feed-forward pipeline, achieving lower end-to-end latency.

\myparagraph{Latency and Bottleneck Analysis.}
At each prompt switch, the text-only rewriter runs concurrently with planner prefill and context-aware CoT decoding on GPU~0.
Across a $45\text{s}$ interaction, total text preparation, including planner prefill, CoT decoding, and prompt rewriting, takes $4.98\text{s}$, compared with $36.8\text{s}$ and $57.5\text{s}$ of core DiT rendering for the 1.3B and 5B variants.
Most text-processing overhead is hidden through asynchronous execution.
Overall, Ours-1.3B operates in a well-balanced pipeline regime, whereas Ours-5B is primarily renderer-bound.

\section{Analysis of Semantic-Guided Memory Retrieval}
\label{sec:supp_memory_analysis}

\begin{figure}[!ht]
\centering
\includegraphics[width=0.85\textwidth]{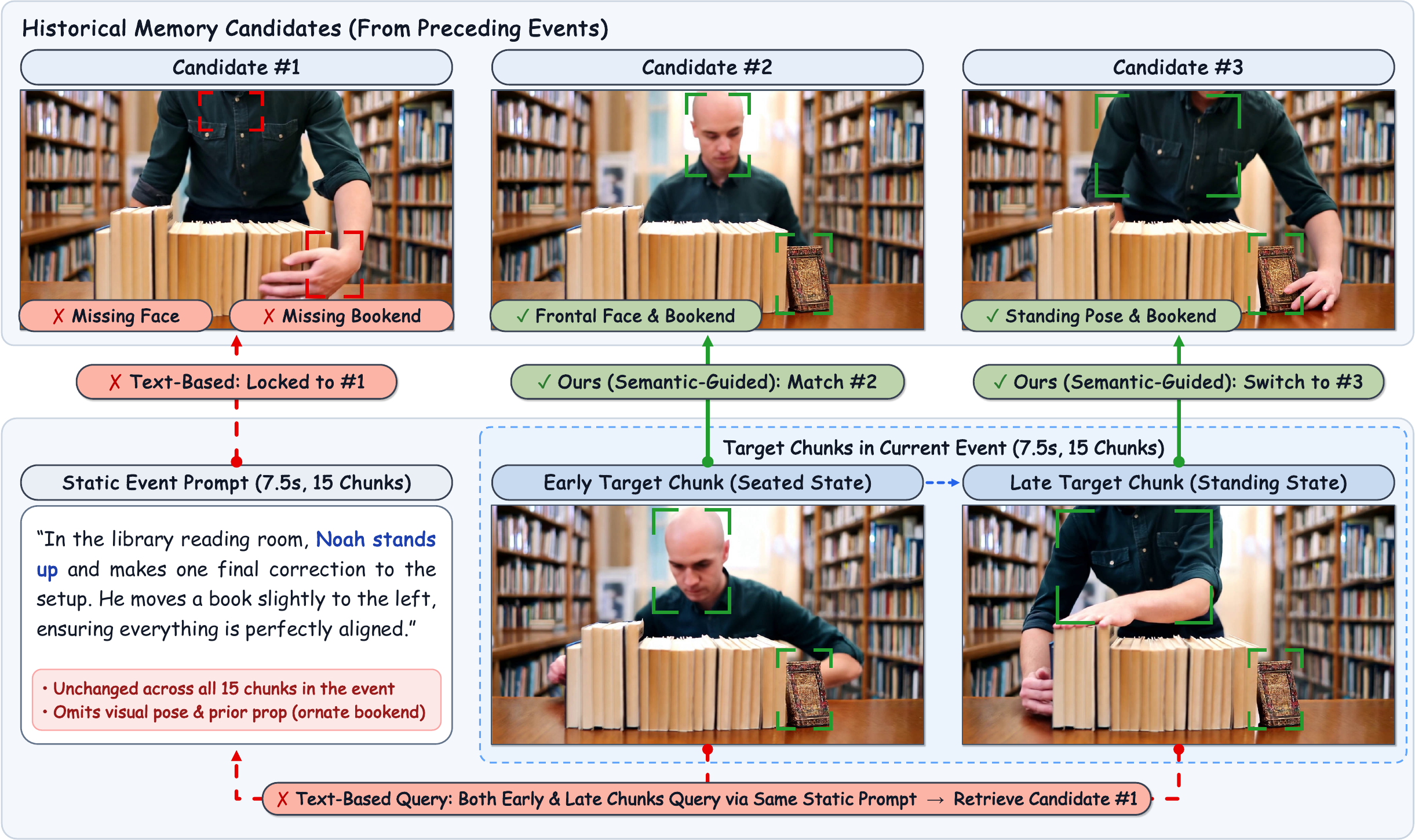}
\caption{
\textbf{Qualitative comparison of semantic-guided and text-based memory retrieval.}
Given historical candidates generated from preceding events by our full model (top), we compare the memories selected by our semantic-guided retrieval (green) and text-based retrieval (red) for two target chunks within a subsequent $7.5\text{s}$ event (bottom).
Note that retrieval is performed prior to rendering using predicted visual semantics (or the static event prompt); the rendered target frames are displayed to visualize the corresponding target visual states.
}
\label{fig:supp_memory_retrieval}
\end{figure}

Fig.~\ref{fig:supp_memory_retrieval} compares our semantic-guided retrieval with text-based retrieval following MemFlow~\cite{ji2025memflow}, which selects historical chunks using cross-attention relevance between the current text query and cached visual features.
The comparison highlights two advantages: 
1) \textit{Visual-state matching:} Text-based retrieval primarily captures coarse event-level similarity, such as the same subjects or broadly related actions, but may overlook the specific visual state required by the current target chunk.
In contrast, our semantic-guided retrieval identifies historical chunks with more closely matched subject states, poses, and overall scene configurations, providing more relevant visual context for preserving facial appearance and object arrangements.
2) \textit{Chunk-level adaptivity:} Within the same event, textual instructions remain unchanged even as subjects and objects evolve over time.
Consequently, text-based retrieval tends to repeatedly select the same historical chunk, whereas our method dynamically retrieves different memories according to the evolving visual semantics of each target chunk at $0.5\text{s}$ intervals.

Although our retrieval uses spatially pooled descriptors for efficiency, the underlying text-aligned visual features capture holistic visual information, including subject configurations, camera perspective, and scene composition.
Such information allows the pooled descriptors to distinguish different visual states beyond coarse textual similarity, enabling more relevant historical context retrieval for streaming generation.

\section{Longer-Horizon Context Exploration and Future Directions}
\label{sec:supp_long_horizon}

\myparagraph{Evaluation Setup and Context Management Strategies.}
In our framework, the semantic planner generates $84$ visual tokens every $0.5\text{s}$, accumulating $1{,}680$ visual tokens per $10\text{s}$ segment alongside user instructions and CoT reasoning.
Consequently, multi-event streams remain within the $10\text{K}$-token training window through $45\text{s}$, approach the context limit near $50\text{s}$, and extend beyond it during $50$--$60\text{s}$.
To examine inference-time behavior beyond this horizon without retraining, we extend the $100$ multi-prompt scripts from MemFlow~\cite{ji2025memflow} to $60\text{s}$ across six $10\text{s}$ events using \textit{Ours-1.3B}.

All runs share the identical generated prefix before reaching the context limit, after which we compare three training-free planner context strategies:
1) \textit{Direct Extrapolation} retains all historical tokens and continues decoding beyond the trained window;
2) \textit{Sliding-Window Truncation} enforces the $10\text{K}$ budget by discarding the earliest context tokens as new segments arrive; and
3) \textit{Sparse Semantic Retention} preserves all instructions, CoT tokens, the initial semantic block, and dense $0.5\text{s}$ semantics for the two most recent events, while subsampling earlier completed events at $2.0\text{s}$ intervals for $4\times$ compression.
Upon truncation or subsampling, the planner refreshes its state over the updated sequence, whereas the renderer retains its historical visual KV archive for semantic-guided retrieval.
Following Sec.~\ref{sec:supp_interactive_eval}, we evaluate each $10\text{s}$ segment using custom-input VBench Quality Score and CLIP Score against the original instructions.

\begin{figure}[!ht]
\centering
\includegraphics[width=\linewidth]{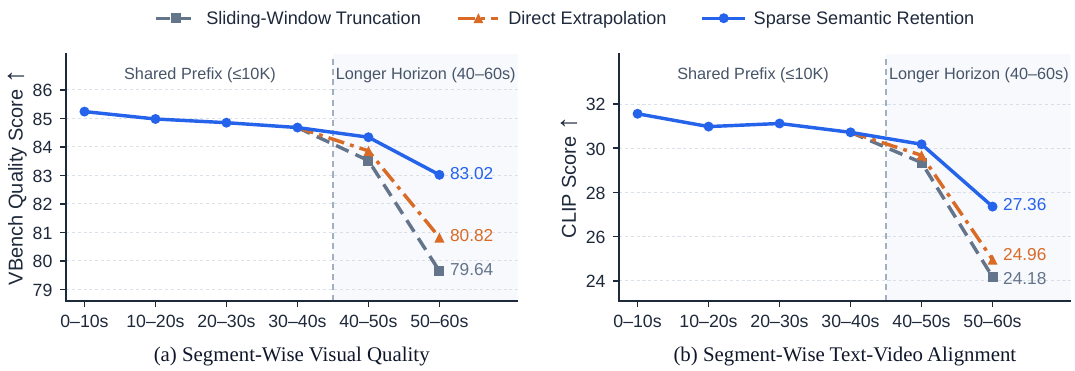}
\caption{
\textbf{Exploration of $60\text{s}$ interactive generation across six $10\text{s}$ events on the $100$ multi-prompt scripts from MemFlow.}
We report segment-wise custom-input VBench Quality Score in (a) and CLIP Score in (b) for \textit{Ours-1.3B}.
All strategies share the same generated prefix over $0$--$40\text{s}$, while the shaded region marks the $40$--$60\text{s}$ horizon where sequences approach and exceed the $10\text{K}$-token training window.
}
\label{fig:supp_long_horizon}
\end{figure}

\myparagraph{Results and Bottleneck Analysis.}
As shown in Fig.~\ref{fig:supp_long_horizon}, both metrics remain stable through $40\text{s}$ and dip only slightly over $40$--$50\text{s}$ since most sequences reach the context limit near $50\text{s}$, before diverging clearly in $50$--$60\text{s}$.
\textit{Sliding-Window Truncation} degrades most severely because dropping early context erases initial entity and scene anchors.
\textit{Direct Extrapolation} also drops markedly due to positional extrapolation beyond the trained context range.
In contrast, \textit{Sparse Semantic Retention} better preserves visual quality and text-video alignment by maintaining long-range semantic anchors within budget.

Even so, noticeable degradation still arises past $50\text{s}$ under sparse retention, occasionally causing structural distortions and stiffer motion.
This remaining gap likely reflects three factors:
1) heuristic subsampling alters the uniform $0.5\text{s}$ token spacing seen during training, introducing a distribution shift;
2) multi-event training clips beyond $50\text{s}$ are scarce in the long tail of \datasetname{}, as shown in Fig.~3(b) of the main paper; and
3) autoregressive errors gradually accumulate across both the planner and causal renderer over extended rollouts.

\myparagraph{Future Directions.}
These findings motivate two extensions of Sec.~6 in the main paper.
First, learnable semantic compression and hierarchical semantic memory could bound planner context growth without disrupting temporal structure.
Second, longer multi-event training sequences and long-horizon error-recovery training could mitigate cumulative drift in minute-scale interactive generation.

\section{\datasetname{} Dataset Construction Details}
\label{sec:supp_dataset}

\myparagraph{Candidate Collection and Multi-Dimensional Filtering.}
We assemble an initial pool of approximately $7\text{M}$ raw videos lasting $5$ to $60\text{s}$, comprising around $6\text{M}$ clips from Koala-36M~\cite{wang2025koala} and $1\text{M}$ publicly accessible online videos collected for academic research.
Candidates first undergo aesthetic and video quality filtering following~\cite{schuhmann2022laion, wang2025koala} to discard blurry, poorly exposed, or heavily compressed footage, followed by TransNet V2~\cite{soucek2024transnet} to remove clips with hard shot cuts.

We then combine YOLO26~\cite{yolo} and Qwen3.5-27B~\cite{qwen2026qwen35} to screen for clear, trackable foreground dynamics.
Salient subjects include people, animals, animated characters, and visually prominent objects that drive the primary activity.
YOLO26 localizes foreground entities on sampled keyframes and removes clips whose largest subject bounding box covers less than $10\%$ of the frame area.

In parallel, Qwen3.5-27B evaluates full-clip suitability across camera continuity and scene complexity.
It removes egocentric first-person recordings, split-screen layouts, intrusive text overlays, and residual edited transitions such as dissolves, fades, wipes, or title cards, while preserving continuous handheld or tracking camera motion.
Clips with unreadable subject appearances or actions, severe multi-subject occlusion, or more than five salient foreground subjects are also excluded.

\myparagraph{Single- vs.\ Multi-Event Classification.}
We recruit $10$ trained annotators under the informed consent protocol in Sec.~\ref{sec:supp_user_study}, compensated at approximately USD $20$ per hour, to label $10\text{K}$ filtered clips while discarding ambiguous borderline cases.

Specifically, annotators answer two auxiliary diagnostic questions before making an overall judgment:
1) \textit{Semantic Event Distinctiveness} evaluates whether the video contains two or more semantically distinct activities or goals rather than steps of a single routine; 2) \textit{Event Boundary Clarity} evaluates whether an identifiable shift in action, goal, or interaction separates consecutive phases.
We fine-tune Qwen3.6-27B~\cite{qwen2026qwen36} with LoRA of rank $64$ on these $10\text{K}$ examples for $3$ epochs, using AdamW at a learning rate of $1\times10^{-4}$ and batch size $32$, training the model to predict both diagnostic judgments before outputting the final event label.
For readability, the prompt templates throughout this section summarize the core annotation instructions, semantic constraints, and output schemas; implementation-specific JSON validation rules and few-shot examples are omitted for brevity.

\input{figures/supp_data_classifier_prompt}

\myparagraph{Multi-Event Segmentation.}
For clips classified as multi-event, Gemini 3.1 Pro~\cite{google2026gemini31} partitions each video into contiguous, non-overlapping semantic events with one-decimal timestamps, which are quantized to $0.5\text{s}$ intervals to align with our planner step size, while generating a concise English instruction $T^{(m)}$ for each segment.
To maintain moderate event granularity, boundaries are placed only at substantive shifts in activity focus, interaction pattern, or behavioral goal, whereas brief gestures, gaze shifts, and transitional micro-motions are merged into adjacent events.
Videos containing any segment shorter than $2.0\text{s}$ after quantization are discarded to avoid over-segmentation.

\myparagraph{Chained Event-Level Annotation.}
To prevent temporal grounding drift across long videos, we employ Qwen3.6-27B for segment-wise chained annotation.
While global screening allows up to five foreground entities in a scene, structured annotation tracks the most visually salient active subjects per segment (up to three in initial segments and four in subsequent segments, summarizing any minor auxiliary entities in \texttt{[Overall]}).
For the initial segment ($m=1$), the model generates a self-contained single-paragraph caption $\widetilde{T}^{(1)}$ ordered as \texttt{[Overall]}, \texttt{[Subject $k$]}, \texttt{[Environment]}, and \texttt{[Camera]}, pairing each subject's visual identity anchor first with its dominant action second.
For the $2.5\text{M}$ single-event videos, Qwen3.6-27B produces both a concise instruction $T^{(1)}$ and a structured caption $\widetilde{T}^{(1)}$ in one pass.

For each subsequent segment $m\ge2$, Qwen3.6-27B receives the preceding and current video segments, the previous caption $\widetilde{T}^{(m-1)}$, and the next unused subject index \texttt{next\_subject\_id} to jointly predict the event-change description $R^{(m)}$, the updated caption $\widetilde{T}^{(m)}$, and the updated \texttt{next\_subject\_id}.
Every \texttt{[Subject $k$]} serves as a permanent physical identity index across the video: subjects that exited in earlier segments and remain absent are omitted without renumbering remaining indices, reappearing subjects retain their original index, and newly entering salient subjects receive consecutive fresh indices.
Moreover, $R^{(m)}$ captures before-to-after transitions and ongoing continuities across \texttt{[Subject $k$]}, \texttt{[Environment]}, and \texttt{[Camera]}, whereas $\widetilde{T}^{(m)}$ describes segment $m$ standalone without cross-segment comparison words such as ``still'', ``continues'', or ``as before''.
The segmentation and chained annotation templates are detailed below.

\input{figures/supp_data_annotation_prompts}

\myparagraph{Global Verification and Human Audit.}
In the final stage, Gemini 3.1 Pro reviews each complete multi-event video alongside its event boundaries and full annotation chain in two sequential steps.
It first checks persistent subject identity consistency across all segments to detect duplicate indices for reappearing entities, index swaps, or multi-entity mixing, and then applies a conservative materiality gate to flag clear factual conflicts in appearance, action transitions, camera behavior, or absent-subject hallucinations while ignoring minor stylistic variations.

Videos with no identity or material factual issues are accepted as \textit{Pass}, whereas samples with unresolvable multi-entity corruption or pervasive cross-segment contradictions are discarded as \textit{Severe Conflict}.
Samples flagged with localized identity remappings or factual corrections are routed as \textit{Flagged Revision}, accounting for roughly $8\%$ of candidates, and are manually reviewed and revised by our $10$ trained annotators.
This pipeline yields $245\text{K}$ curated multi-event videos alongside $2.5\text{M}$ single-event videos.
In addition, a blind human audit on $1{,}000$ randomly sampled \textit{Pass} multi-event sequences confirms a $97.4\%$ acceptance precision.
The verification prompt and a curated four-event annotation example from \datasetname{} are presented below.

\input{figures/supp_data_verify_and_example}

\section{User Study Details}
\label{sec:supp_user_study}

\begin{figure}[!ht]
\centering
\begin{subfigure}{0.82\textwidth}
    \centering
    \includegraphics[width=\linewidth]{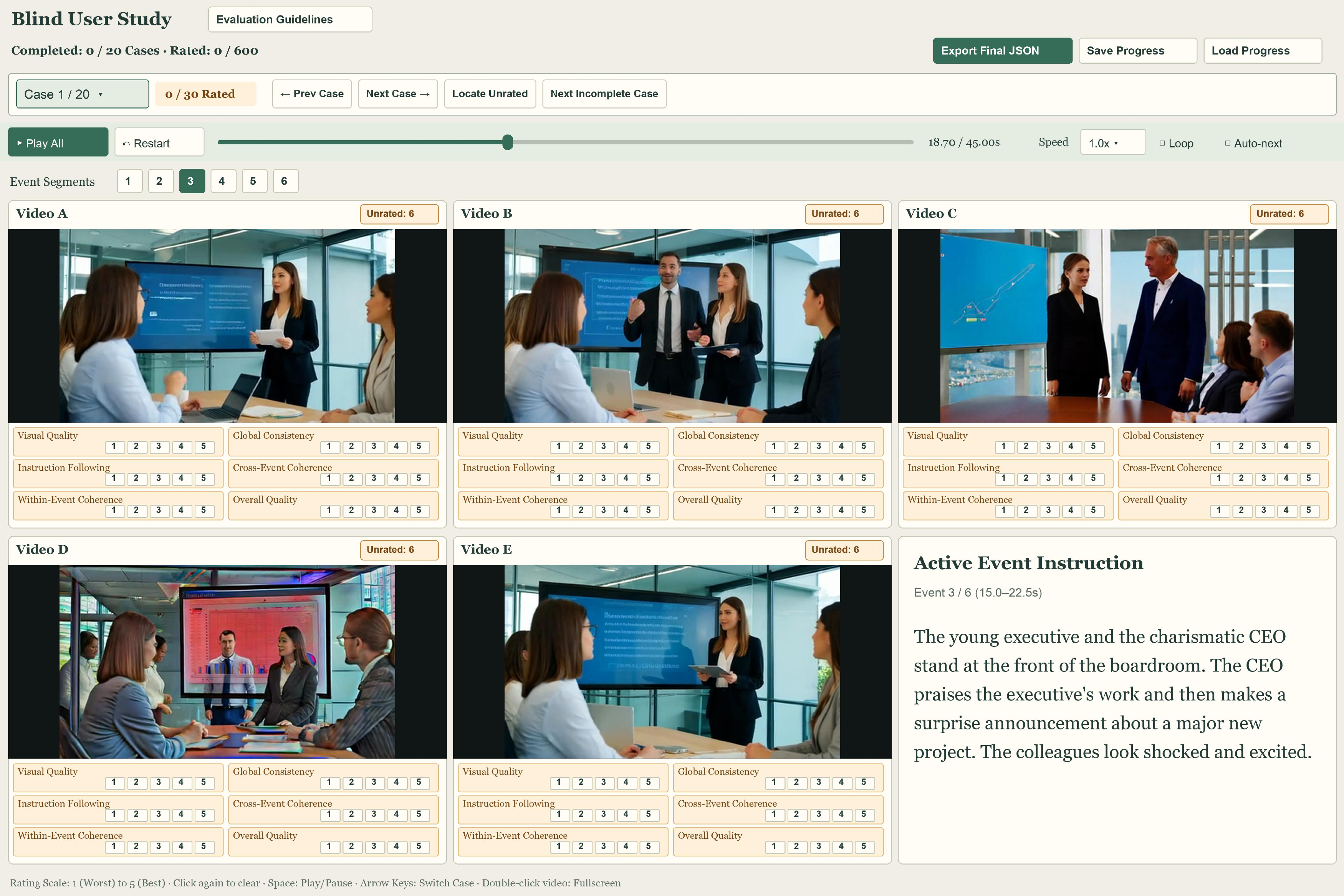}
    \caption{Main Evaluation Interface}
    \label{fig:supp_user_study_interface}
\end{subfigure}

\vspace{6pt}

\begin{subfigure}{0.82\textwidth}
    \centering
    \includegraphics[width=\linewidth]{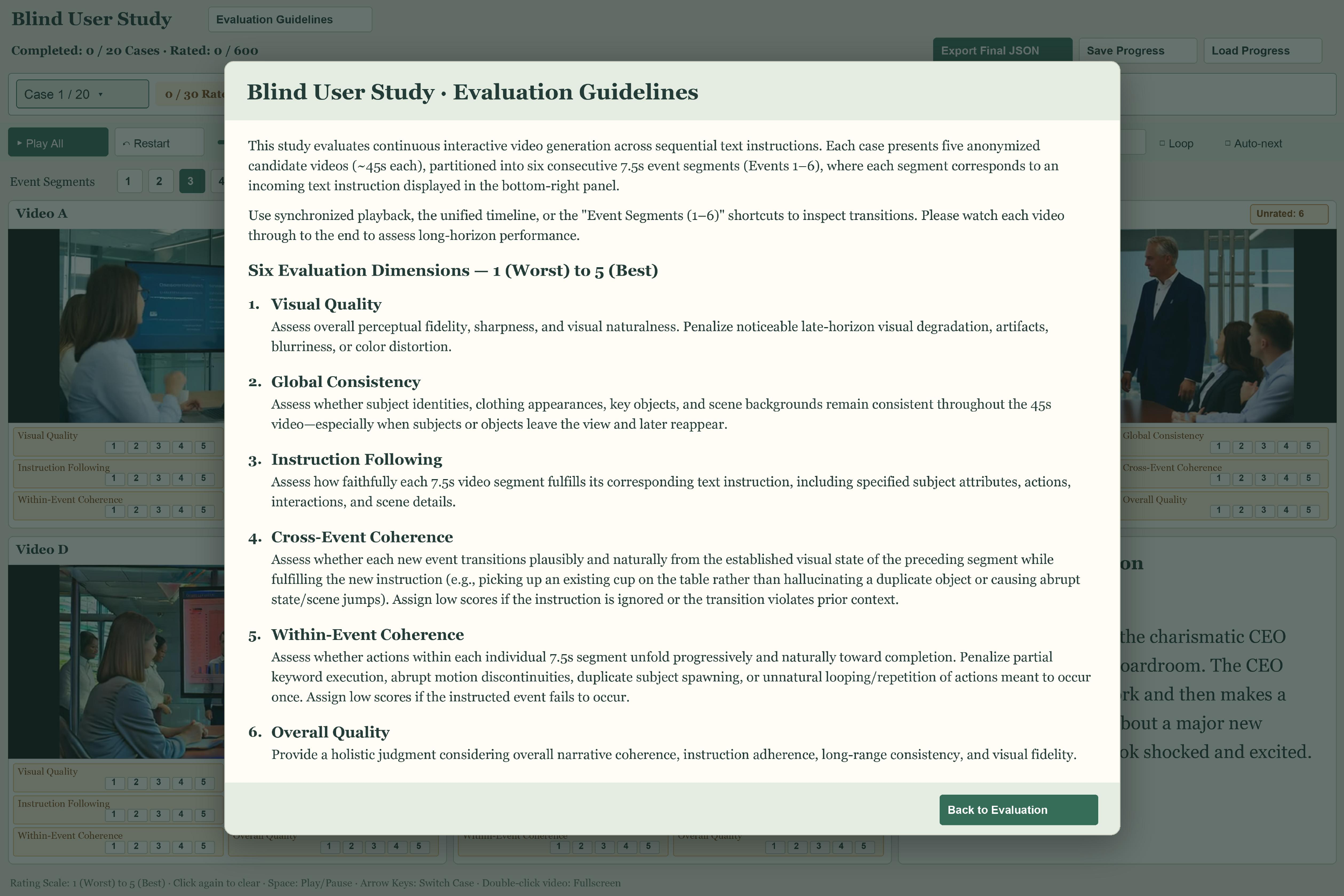}
    \caption{Evaluation Guidelines}
    \label{fig:supp_user_study_guidelines}
\end{subfigure}
\caption{
\textbf{Interface and rating guidelines for the blind user study.}
(a) Main evaluation workspace displaying five anonymized, randomly ordered candidate videos alongside synchronized playback controls, event-segment shortcuts, and the active event prompt.
(b) Evaluation guidelines defining the task setup and the six five-point rating criteria.
}
\label{fig:supp_user_study_ui}
\end{figure}

\myparagraph{Participants and Ethical Compliance.}
We conduct a blind user study with 30 independent participants to assess the perceptual quality and narrative coherence of generated interactive videos.
All participants are unaffiliated with this project, hold graduate-level backgrounds in computer vision or digital media, and are proficient in English.
Each participant receives approximately USD 25 for completing an evaluation session lasting around one hour.
Informed consent is obtained from all participants prior to the session, and participation is strictly voluntary.
Because the study involves only non-invasive perceptual ratings of benign synthetic videos, poses minimal risk to participants, and collects no personally identifiable information (PII), no formal Institutional Review Board (IRB) review is sought.

\myparagraph{Evaluation Protocol and Interface.}
Following Sec.~5.1 of the main paper, we evaluate 20 randomly sampled multi-event prompt sequences ($45\text{s}$, $6\times7.5\text{s}$), comparing Ours-1.3B with Self-Forcing, LongLive, MemFlow, and IAMFlow under the identical generation protocol.
For each sequence, every participant views all five candidate videos side by side with method identities hidden and display order independently randomized per case.
Participants inspect each full $45\text{s}$ sequence across all event boundaries and rate every video on a five-point Likert scale ($1$: worst, $5$: best) across six dimensions: Visual Quality, Global Consistency, Instruction Following, Cross-Event Coherence, Within-Event Coherence, and Overall Quality.
Ratings are aggregated into Mean Opinion Scores (MOS) with 95\% confidence intervals, as reported in Fig.~1(c) of the main paper.
Fig.~\ref{fig:supp_user_study_ui} shows the evaluation interface and rating guidelines provided to participants.

\FloatBarrier
\section{Ethical Considerations and Societal Impact}
\label{sec:supp_ethics}

\myparagraph{Dataset Sourcing and Safety Filtering.}
All candidate videos in \datasetname{} are sourced from public academic datasets~\cite{wang2025koala} and publicly accessible online platforms strictly for non-commercial academic research.
During VLM screening and global verification in Sec.~\ref{sec:supp_dataset}, clips containing NSFW content, graphic violence, hate symbols, prominent commercial watermarks, exposed private identity or contact information, or non-public personal recordings are filtered out, and all textual annotations use generic visual descriptions without real-world personal names.

To respect creator copyrights and platform terms of service, and in accordance with standard practices in large-scale video dataset curation~\cite{chen2024panda, wang2025koala}, raw copyrighted video files will not be redistributed.
Any released annotations and metadata will be licensed under CC BY-NC 4.0, with our data curation pipeline and scripts provided under research-only terms strictly for non-commercial academic use; access to underlying source videos remains governed by their original licenses and platform terms, supported by a dedicated opt-out and takedown channel.

\myparagraph{Human Subjects and Annotator Welfare.}
All perceptual evaluation sessions in Sec.~\ref{sec:supp_eval_protocol} and Sec.~\ref{sec:supp_user_study} as well as dataset annotation and audit tasks in Sec.~\ref{sec:supp_dataset} involve adult contributors participating voluntarily with prior informed consent.
All participants and annotators receive fair hourly compensation above local minimum wage standards, inspect only filtered benign video content posing minimal risk, and have no PII collected or stored from them during any study or annotation session.

\myparagraph{Societal Impact and Misuse Mitigation.}
While Semantic Forcing advances controllable narrative video generation for creative storyboarding, digital media, and interactive simulation, generative video models carry potential risks of misuse such as deceptive media synthesis or impersonation.
All models, code, and data developed in this work are intended strictly for academic research, and we strongly discourage their use for deceptive, harmful, or non-consensual human synthesis.
Furthermore, we emphasize the importance of transparency in synthetic media, including clear provenance labeling and watermarking of AI-generated content, and support ongoing efforts across the research community to develop robust synthetic video detection mechanisms and safety safeguards against potential misuse.

%% file: figures/supp_planner_format.tex
\vspace{4pt}
\noindent
\begin{tcolorbox}[
    enhanced,
    breakable,
    skin first=enhanced,
    skin middle=enhanced,
    skin last=enhanced,
    pad at break*=4pt,
    colback=black!2,
    colframe=black!55,
    colbacktitle=black!12,
    coltitle=black,
    fonttitle=\bfseries\small,
    fontupper=\footnotesize\ttfamily\linespread{0.98}\selectfont,
    boxrule=0.5pt,
    arc=2.5pt,
    left=6pt, right=6pt, top=5pt, bottom=5pt,
    title={(a) Initial Event ($m=1$)} %
]
\detokenize{<|im_start|>system}\\
You are a helpful assistant.\\
\detokenize{<|im_end|>}\\[1mm]
\detokenize{<|im_start|>user}\\
Generate a continuous sequence of video frames based on the following instruction:\\
$T^{(1)}$\\
\textcolor{gray}{[Optional duration: $D^{(1)}$ seconds]}\\
\textcolor{gray}{[Optional TI2V: $\mathcal{B}_0(I_{\mathrm{init}})$]}\\
\detokenize{<|im_end|>}\\[1mm]
\detokenize{<|im_start|>assistant}\\
$\mathcal{B}(S_1)\;\mathcal{B}(S_2)\;\cdots\;\mathcal{B}(S_{n_1})$\\
\textcolor{gray}{[Optional \detokenize{<|im_end|>}]}
\end{tcolorbox}

\vspace{1mm}
\noindent
\begin{tcolorbox}[
    enhanced,
    breakable,
    skin first=enhanced,
    skin middle=enhanced,
    skin last=enhanced,
    pad at break*=4pt,
    colback=black!2,
    colframe=black!55,
    colbacktitle=black!12,
    coltitle=black,
    fonttitle=\bfseries\small,
    fontupper=\footnotesize\ttfamily\linespread{0.98}\selectfont,
    boxrule=0.5pt,
    arc=2.5pt,
    left=6pt, right=6pt, top=5pt, bottom=5pt,
    title={\textbf{(b) Subsequent Events ($m\ge2$)}}
]
\textcolor{gray}{[Preceding conversation and visual semantics retained]}\\[1mm]
\detokenize{<|im_start|>user}\\
Continue the video frames based on the updated instruction:\\
$T^{(m)}$\\
\textcolor{gray}{[Optional duration: $D^{(m)}$ seconds]}\\
\detokenize{<|im_end|>}\\[1mm]
\detokenize{<|im_start|>assistant}\\
\detokenize{<think>}$R^{(m)}$\detokenize{</think>}\\
$\mathcal{B}(S_{n_{m-1}+1})\;\cdots\;\mathcal{B}(S_{n_m})$\\
\textcolor{gray}{[Optional \detokenize{<|im_end|>}]}
\end{tcolorbox}

\vspace{1mm}
{\centering\footnotesize
$\mathcal{B}(S_i) = \texttt{<im\_start><S2>}\,s_{i,1}\cdots s_{i,81}\,\texttt{<im\_end>}$ ($84$ tokens),\quad
$\mathcal{B}_0(I_{\mathrm{init}}) = \texttt{<im\_start><S0>}\,s_{\mathrm{init},1}\cdots s_{\mathrm{init},729}\,\texttt{<im\_end>}$ ($732$ tokens)\par
}
\vspace{4pt}

%% file: tables/supp_comp_open_ended.tex
\begin{table*}[t]
\centering
\caption{
\textbf{Quantitative evaluation of open-ended planning on NarraStream-Bench}
under the 45s (6${\times}$7.5s) multi-prompt setting.
Baseline results are taken from Tab.~1 of the main paper for direct comparison.
$\Delta$ denotes the score change relative to the corresponding specified-duration result in Tab.~1.
Best and second-best absolute scores are \textbf{bolded} and \uline{underlined}.
}
\vspace{-.3em}
\label{tab:supp_open_ended}
\setlength{\tabcolsep}{2.8pt}
\newcommand{\opendelta}[1]{{\footnotesize\color{black!70}\ensuremath{#1}}}
\resizebox{\textwidth}{!}{%
\begin{tabular}{@{}l*{13}{c}@{}}

\toprule
\multirow{2}{*}{Model}
& \multicolumn{5}{c}{Visual Quality $\uparrow$}
& \multicolumn{3}{c}{Temporal Consistency $\uparrow$}
& \multicolumn{3}{c}{Instruction Compliance $\uparrow$}
& \multicolumn{2}{c}{Semantic Coherence $\uparrow$} \\
\cmidrule(lr){2-6}
\cmidrule(lr){7-9}
\cmidrule(lr){10-12}
\cmidrule(lr){13-14}
& \shortstack{\strut Subject\\\strut Consist.}
& \shortstack{\strut Background\\\strut Consist.}
& \shortstack{\strut Temporal\\\strut Flicker.}
& \shortstack{\strut Motion\\\strut Smooth.}
& \shortstack{\strut Temporal\\\strut Stability}
& \shortstack{\strut Boundary\\\strut Smooth.}
& \shortstack{\strut Adjacent\\\strut Consist.}
& \shortstack{\strut Long-range\\\strut Consist.}
& \shortstack{\strut Entity\\\strut Ground.}
& \shortstack{\strut Dynamic\\\strut Traj.}
& \shortstack{\strut VLM\\\strut Align.}
& \shortstack[c]{\strut Cross-Event\\\strut Transition}
& \shortstack[c]{\strut Within-Event\\\strut Progression} \\
\midrule

Self-Forcing
& 84.02 & 83.46 & 88.29 & 58.58
& 29.69 & 48.75
& 57.99 & 45.88
& 61.18 & 55.17 & 53.62
& 45.42 & 45.99 \\

Rolling Forcing
& 92.52 & 89.81 & 85.61 & 49.46
& 68.06 & 40.14
& 76.54 & 61.27
& 68.17 & 58.67 & 64.45
& 52.28 & 51.27 \\

Infinity-RoPE
& 92.76 & 91.66 & 89.45 & 54.82
& 73.20 & 50.13
& 76.16 & 80.78
& 69.59 & 59.97 & 63.99
& 50.81 & 52.64 \\

LongLive
& \uline{95.35} & 92.85
& 91.27 & 58.95
& 74.20 & 49.21
& 83.81 & 89.63
& 72.58 & 63.41 & 65.07
& 56.15 & 58.70 \\

MemFlow
& 95.06 & 92.23 & 90.98 & 57.22
& 73.14 & 44.22
& 83.58 & 88.85
& 70.00 & 63.91 & 63.15
& 59.72 & 63.19 \\

Anchor Forcing
& 90.32 & 89.72 & 85.59 & 40.75
& 71.03 & \uline{55.86}
& 68.41 & 75.08
& 71.36 & 58.13 & 66.33
& 60.33 & 61.31 \\

IAMFlow
& \textbf{95.37} & 92.45
& 91.11 & 57.07
& 74.18 & 47.69
& 85.40 & \textbf{90.44}
& 69.14 & 64.47 & 60.63
& 58.10 & 65.70 \\

\midrule

\rowcolor{OursBlue}
\textbf{Ours-1.3B (Open-Ended)}
& 94.02 & \textbf{93.07}
& \textbf{92.85} & \textbf{62.99}
& \uline{74.29} & 55.60
& \uline{85.64} & 90.21
& \uline{76.45} & \uline{64.58}
& \uline{77.25}
& \uline{76.42} & \uline{78.82} \\

\rowcolor{OursBlue}
{\footnotesize\color{black!70}\hspace{0.5em}$\Delta$ vs.\ Specified-Duration}
& \opendelta{+0.55} & \opendelta{-0.09}
& \opendelta{-0.13} & \opendelta{-0.09}
& \opendelta{+0.04} & \opendelta{-0.14}
& \opendelta{+0.22} & \opendelta{+0.53}
& \opendelta{+0.08} & \opendelta{-0.39}
& \opendelta{-2.75}
& \opendelta{-2.17} & \opendelta{-2.86} \\

\rowcolor{OursBlue}
\textbf{Ours-5B (Open-Ended)}
& 94.88 & \uline{92.95}
& \uline{92.70} & \uline{62.58}
& \textbf{75.00} & \textbf{55.97}
& \textbf{86.03} & \uline{90.35}
& \textbf{77.80} & \textbf{65.01}
& \textbf{77.39}
& \textbf{76.44} & \textbf{80.05} \\

\rowcolor{OursBlue}
{\footnotesize\color{black!70}\hspace{0.5em}$\Delta$ vs.\ Specified-Duration}
& \opendelta{+0.47} & \opendelta{+0.37}
& \opendelta{+0.03} & \opendelta{+0.32}
& \opendelta{-0.31} & \opendelta{-0.09}
& \opendelta{+0.46} & \opendelta{-0.12}
& \opendelta{-0.04} & \opendelta{+0.08}
& \opendelta{-3.44}
& \opendelta{-3.19} & \opendelta{-3.00} \\

\bottomrule
\end{tabular}%
}
\vspace{-1em}
\end{table*}

%% file: tables/supp_ablation_full.tex
\begin{table}[!ht]
\centering
\caption{\textbf{Complete quantitative ablations on NarraStream-Bench (45s).} Best and second-best scores are \textbf{bolded} and \uline{underlined}.}
\label{tab:supp_ablation_full}
\setlength{\tabcolsep}{3pt}
\resizebox{\textwidth}{!}{%
\begin{tabular}{@{}lccccc@{}}
\toprule
 & & & & \multicolumn{2}{c}{Semantic Coherence} \\
\cmidrule(lr){5-6}
Variant & Visual Quality $\uparrow$ & Temporal Consistency $\uparrow$ & Instruction Compliance $\uparrow$ & Cross-Event Transition $\uparrow$ & Within-Event Progression $\uparrow$ \\
\midrule
\rowcolor{GroupGray}
\multicolumn{6}{@{}l}{\textit{Semantic Planning and Reasoning}} \\
w/o planner (raw prompts) & 80.50 & 72.78 & 65.74 & 57.13 & 62.58 \\
w/o planner (rewritten prompts) & 82.66 & 73.54 & 66.50 & 57.11 & 63.85 \\
\quad + external reasoning & 81.97 & 74.01 & 67.49 & 58.62 & 65.55 \\
w/o CoT & \uline{82.90} & \uline{76.39} & 72.93 & 74.06 & 81.34 \\
\midrule
\rowcolor{GroupGray}
\multicolumn{6}{@{}l}{\textit{Visual Memory \& Prompt Rewriting}} \\
w/o memory \& sink & 78.92 & 67.38 & 72.52 & 74.63 & 73.11 \\
w/o memory $C_i^{\mathrm{mem}}$ & 79.43 & 70.04 & 73.30 & 77.98 & 80.66 \\
random memory & 80.69 & 70.13 & 73.17 & 78.17 & 81.64 \\
text-based retrieval & 80.71 & 72.59 & \uline{73.33} & \textbf{79.00} & 81.47 \\
w/o prompt rewriter & 82.59 & 73.43 & 72.35 & 78.55 & \textbf{82.19} \\
\midrule
\rowcolor{OursBlue}
\textbf{Ours (full)} & \textbf{83.39} & \textbf{76.95} & \textbf{73.78} & \uline{78.59} & \uline{81.68} \\
\bottomrule
\end{tabular}%
}
\end{table}

%% file: figures/supp_external_vlm_prompt.tex
\vspace{4pt}
\noindent
\begin{tcolorbox}[
    enhanced,
    breakable,
    skin first=enhanced,
    skin middle=enhanced,
    skin last=enhanced,
    pad at break*=4pt,
    colback=black!2,
    colframe=black!55,
    colbacktitle=black!12,
    coltitle=black,
    fonttitle=\bfseries\small,
    fontupper=\footnotesize,
    boxrule=0.5pt,
    arc=2.5pt,
    left=6pt, right=6pt, top=5pt, bottom=5pt,
    title={\textbf{(a) Prompt Template for External VLM Reasoning}}
]
\textbf{System Prompt:}
You are a video event analyst. Given the visual context from a previously generated video segment and historical instructions, describe how the current visual state should transition and evolve to fulfill the new instruction while retaining established appearance details.
\vspace{1mm}

\textbf{Guidelines:}
\begin{itemize}
    \setlength{\itemsep}{0.5pt}\setlength{\parskip}{0pt}\setlength{\topsep}{1pt}
    \item Identify existing subjects, held objects, and their ending states from the preceding video segment.
    \item Specify the minimal grounded transition (e.g., resolving occupied hands first) and how actions, interactions, and spatial relationships evolve under the new instruction, without replaying completed past actions.
    \item Preserve established subject identities and exact appearance descriptions from the instructions unless changes are explicitly requested.
    \item Describe environmental states and camera movements separately without inventing unsupported objects or story beats.
\end{itemize}
\vspace{1mm}

\textbf{User Input:}\\
Previous video: \texttt{[Preceding $7.5\text{s}$ video segment]} $\;\mid\;$
Historical instructions: $T^{(1:m-1)}$ $\;\mid\;$
New instruction: $T^{(m)}$
\vspace{1mm}

\textbf{Output Format:}\\
\texttt{[Overall]} Concise summary of the event transition and scene progression.\\
\texttt{[Subject $k$]} One entry per active subject ($k=1,2,\dots$), specifying retained appearance alongside grounded state transition and action evolution.\\
\texttt{[Environment]} Retained or updated scene setting.\quad
\texttt{[Camera]} Retained or updated camera state.
\end{tcolorbox}

\vspace{1mm}
\noindent
\begin{tcolorbox}[
    enhanced,
    breakable,
    skin first=enhanced,
    skin middle=enhanced,
    skin last=enhanced,
    pad at break*=4pt,
    colback=black!2,
    colframe=black!55,
    colbacktitle=black!12,
    coltitle=black,
    fonttitle=\bfseries\small,
    fontupper=\footnotesize,
    boxrule=0.5pt,
    arc=2.5pt,
    left=6pt, right=6pt, top=5pt, bottom=5pt,
    title={\textbf{(b) Representative Example of External VLM Output}}
]
\textbf{Preceding Visual State:} A young man in a denim jacket and a woman in a beige coat sit talking on a wooden bench along a park walkway.\\
\textbf{New Instruction $T^{(m)}$:} ``They walk side by side along the walkway.''\\[0.8mm]
\textbf{External VLM Output:} \texttt{[Overall]} The young man and woman transition from sitting on the bench to standing up and walking side by side along the walkway. \texttt{[Subject 1]} The young man in the denim jacket rises from the wooden bench, steps onto the walkway, and walks forward at a steady pace. \texttt{[Subject 2]} The woman in the beige coat stands up alongside him and walks forward in step. \texttt{[Environment]} The wooden bench and tree-lined park walkway remain consistent. \texttt{[Camera]} The camera remains static, maintaining a steady view along the walkway.
\end{tcolorbox}
\vspace{4pt}

%% file: tables/supp_comp_shared_captions.tex
\begin{table}[!ht]
\centering
\caption{
\textbf{Controlled sub-metric comparison on NarraStream-Bench (45s) using shared structured captions.}
All baseline methods and our renderer receive the exact same rewritten structured captions $\widetilde{T}^{(m)}$ produced by $\mathcal{P}$, and all generated videos are evaluated against the original instructions $T^{(m)}$.
Best and second-best scores are \textbf{bolded} and \uline{underlined}.
}
\label{tab:supp_comp_shared_captions}
\setlength{\tabcolsep}{2.8pt}
\resizebox{\textwidth}{!}{%
\begin{tabular}{@{}l*{13}{c}@{}}
\toprule
\multirow{2}{*}{Model}
& \multicolumn{5}{c}{Visual Quality $\uparrow$}
& \multicolumn{3}{c}{Temporal Consistency $\uparrow$}
& \multicolumn{3}{c}{Instruction Compliance $\uparrow$}
& \multicolumn{2}{c}{Semantic Coherence $\uparrow$} \\
\cmidrule(lr){2-6}
\cmidrule(lr){7-9}
\cmidrule(lr){10-12}
\cmidrule(lr){13-14}
& \shortstack{\strut Subject\\\strut Consist.}
& \shortstack{\strut Background\\\strut Consist.}
& \shortstack{\strut Temporal\\\strut Flicker.}
& \shortstack{\strut Motion\\\strut Smooth.}
& \shortstack{\strut Temporal\\\strut Stability}
& \shortstack{\strut Boundary\\\strut Smooth.}
& \shortstack{\strut Adjacent\\\strut Consist.}
& \shortstack{\strut Long-range\\\strut Consist.}
& \shortstack{\strut Entity\\\strut Ground.}
& \shortstack{\strut Dynamic\\\strut Traj.}
& \shortstack{\strut VLM\\\strut Align.}
& \shortstack[c]{\strut Cross-Event\\\strut Transition}
& \shortstack[c]{\strut Within-Event\\\strut Progression} \\
\midrule

Self-Forcing
& 85.76 & 83.55 & 89.51 & 60.23 & 39.93
& 51.89 & 62.35 & 49.88
& 68.54 & 56.42 & 59.92
& 46.82 & 47.06 \\

Rolling Forcing
& 92.13 & 89.62 & 85.77 & 50.25 & 70.19
& 37.29 & 74.65 & 70.12
& 73.43 & 57.98 & 72.98
& 50.17 & 53.65 \\

Infinity-RoPE
& 91.98 & 90.28 & 89.11 & 54.16 & 69.69
& 49.91 & 72.76 & 77.85
& 76.00 & 60.09 & 68.93
& 50.10 & 51.66 \\

LongLive
& \uline{94.97} & 91.77 & 91.17 & 59.79 & \uline{74.45}
& 46.50 & 82.75 & 88.74
& 76.13 & 62.84 & 72.46
& 55.47 & 57.35 \\

MemFlow
& 94.10 & 91.05 & 90.75 & 57.98 & 72.85
& 41.37 & 79.54 & 86.60
& 74.65 & 62.23 & 73.21
& 60.03 & 62.54 \\

Anchor Forcing
& 90.51 & 88.54 & 84.93 & 41.38 & 73.30
& 55.66 & 69.63 & 76.95
& 75.16 & 58.69 & 72.36
& 61.80 & 61.60 \\

IAMFlow
& \textbf{95.02} & 91.88 & 90.98 & 58.10 & 73.36
& 46.22 & 83.79 & 89.09
& 74.56 & 64.31 & 70.83
& 60.33 & 64.19 \\

\midrule

\rowcolor{OursBlue}
\textbf{Ours-1.3B}
& 93.47 & \textbf{93.16} & \textbf{92.98} & \textbf{63.08} & 74.25
& \uline{55.74} & \uline{85.42} & \uline{89.68}
& \uline{76.37} & \textbf{64.97} & \uline{80.00}
& \uline{78.59} & \uline{81.68} \\

\rowcolor{OursBlue}
\textbf{Ours-5B}
& 94.41 & \uline{92.58} & \uline{92.67} & \uline{62.26} & \textbf{75.31}
& \textbf{56.06} & \textbf{85.57} & \textbf{90.47}
& \textbf{77.84} & \uline{64.93} & \textbf{80.83}
& \textbf{79.63} & \textbf{83.05} \\

\bottomrule
\end{tabular}%
}
\end{table}

\begin{table}[!ht]
\centering
\caption{
\textbf{Aggregated dimension scores and changes ($\Delta$) under shared structured captions on NarraStream-Bench (45s).}
Parentheses denote score changes ($\Delta$) relative to using raw instructions without prompt rewriting.
Best and second-best aggregated scores are \textbf{bolded} and \uline{underlined}.
}
\label{tab:supp_comp_shared_summary}
\setlength{\tabcolsep}{5pt}
\resizebox{\textwidth}{!}{%
\begin{tabular}{@{}lccccc@{}}
\toprule
 & & & & \multicolumn{2}{c}{Semantic Coherence} \\
\cmidrule(lr){5-6}
Model & Visual Quality $\uparrow$ & Temporal Consistency $\uparrow$ & Instruction Compliance $\uparrow$ & Cross-Event Transition $\uparrow$ & Within-Event Progression $\uparrow$ \\
\midrule
Self-Forcing & 71.80 {\footnotesize(+2.99)} & 54.71 {\footnotesize(+3.84)} & 61.63 {\footnotesize(+4.97)} & 46.82 {\footnotesize(+1.40)} & 47.06 {\footnotesize(+1.07)} \\
Rolling Forcing & 77.59 {\footnotesize(+0.50)} & 60.69 {\footnotesize(+1.37)} & 68.13 {\footnotesize(+4.37)} & 50.17 {\footnotesize(-2.11)} & 53.65 {\footnotesize(+2.38)} \\
Infinity-RoPE & 79.04 {\footnotesize(-1.34)} & 66.84 {\footnotesize(-2.18)} & 68.34 {\footnotesize(+3.82)} & 50.10 {\footnotesize(-0.71)} & 51.66 {\footnotesize(-0.98)} \\
LongLive & 82.43 {\footnotesize(-0.09)} & 72.66 {\footnotesize(-1.56)} & 70.48 {\footnotesize(+3.46)} & 55.47 {\footnotesize(-0.68)} & 57.35 {\footnotesize(-1.35)} \\
MemFlow & 81.35 {\footnotesize(-0.38)} & 69.17 {\footnotesize(-3.05)} & 70.03 {\footnotesize(+4.34)} & 60.03 {\footnotesize(+0.31)} & 62.54 {\footnotesize(-0.65)} \\
Anchor Forcing & 75.73 {\footnotesize(+0.25)} & 67.41 {\footnotesize(+0.96)} & 68.74 {\footnotesize(+3.47)} & 61.80 {\footnotesize(+1.47)} & 61.60 {\footnotesize(+0.29)} \\
IAMFlow & 81.87 {\footnotesize(-0.17)} & 73.03 {\footnotesize(-1.48)} & 69.90 {\footnotesize(+5.15)} & 60.33 {\footnotesize(+2.23)} & 64.19 {\footnotesize(-1.51)} \\
\midrule
\rowcolor{OursBlue}
\textbf{Ours-1.3B} & \uline{83.39} {\footnotesize(+0.80)} & \uline{76.95} {\footnotesize(+3.52)} & \uline{73.78} {\footnotesize(+1.43)} & \uline{78.59} {\footnotesize(+0.04)} & \uline{81.68} {\footnotesize(-0.51)} \\
\rowcolor{OursBlue}
\textbf{Ours-5B} & \textbf{83.45} {\footnotesize(+0.75)} & \textbf{77.37} {\footnotesize(+3.04)} & \textbf{74.53} {\footnotesize(+1.28)} & \textbf{79.63} {\footnotesize(+0.06)} & \textbf{83.05} {\footnotesize(+0.03)} \\
\bottomrule
\end{tabular}%
}
\end{table}

%% file: figures/supp_rewriter_prompt.tex
\vspace{4pt}
\noindent
\begin{tcolorbox}[
    enhanced,
    breakable,
    skin first=enhanced,
    skin middle=enhanced,
    skin last=enhanced,
    pad at break*=4pt,
    colback=black!2,
    colframe=black!55,
    colbacktitle=black!12,
    coltitle=black,
    fonttitle=\bfseries\small,
    fontupper=\footnotesize,
    boxrule=0.5pt,
    arc=2.5pt,
    left=6pt, right=6pt, top=5pt, bottom=5pt,
    title={\textbf{(a) Prompt Template for the Built-in Prompt Rewriter $\mathcal{P}$}}
]
\textbf{System Prompt:}
Convert the current user instruction into a self-contained structured video caption.
Historical instructions are provided only for resolving entity references and carrying forward established appearance and scene attributes.
Do not invent new subjects, actions, attributes, or scene details beyond what is necessary for reference resolution and self-containment.
\vspace{1mm}

\textbf{Guidelines:}
\begin{itemize}
    \setlength{\itemsep}{0.5pt}\setlength{\parskip}{0pt}\setlength{\topsep}{1pt}
    \item Resolve pronouns and ambiguous references using historical instructions, and list one entry for each currently active subject in first-appearance order.
    \item Carry forward each subject's appearance description exactly when previously specified, without cross-subject mixing.
    If no appearance details were provided, do not invent clothing or visual attributes.
    \item Describe only the actions specified in the current instruction.
    Do not repeat completed past actions or add unrequested gestures, interactions, or scene events.
    \item Retain the established environment unless it is explicitly updated in the current instruction.
    \item Set camera motion to \texttt{Unspecified.} unless it is explicitly requested in the current instruction.
\end{itemize}
\vspace{1mm}

\textbf{User Input:}\\
Historical instructions: $T^{(1:m-1)}$ (empty at $m=1$) $\;\mid\;$
Current instruction: $T^{(m)}$
\vspace{1mm}

\textbf{Output Format:}\\
\texttt{[Overall]} Concise summary of the current event.\\
\texttt{[Subject $k$]} Active subject identity, carried-forward appearance (if specified), and current actions ($k=1,2,\dots$).\\
\texttt{[Environment]} Retained or updated scene setting.\\
\texttt{[Camera]} Explicitly requested camera motion (or \texttt{Unspecified.}).
\end{tcolorbox}

\vspace{1mm}
\noindent
\begin{tcolorbox}[
    enhanced,
    breakable,
    skin first=enhanced,
    skin middle=enhanced,
    skin last=enhanced,
    pad at break*=4pt,
    colback=black!2,
    colframe=black!55,
    colbacktitle=black!12,
    coltitle=black,
    fonttitle=\bfseries\small,
    fontupper=\footnotesize,
    boxrule=0.5pt,
    arc=2.5pt,
    left=6pt, right=6pt, top=5pt, bottom=5pt,
    title={\textbf{(b) Representative Example of Prompt Rewriting}}
]
\textbf{Historical Excerpt ($T^{(1)}$):}
``In the library reading room, Noah, a tall man with a shaved head and a dark green work shirt, quietly arranges items on the table.''\\
\textbf{Current Instruction ($T^{(5)}$):}
``He stands up and makes one final correction to the setup, moving a book slightly to the left.''\\[0.8mm]
\textbf{Rewritten Structured Caption $\widetilde{T}^{(5)}$:}\\
\texttt{[Overall]} Noah stands up and makes a final correction to the setup in the library reading room.\\
\texttt{[Subject 1]} Noah, a tall man with a shaved head and a dark green work shirt, stands up and makes one final correction to the setup, moving a book slightly to the left.\\
\texttt{[Environment]} The library reading room.\\
\texttt{[Camera]} Unspecified.
\end{tcolorbox}
\vspace{4pt}

%% file: tables/supp_comp_single_prompt.tex
\begin{table}[!ht]
\centering
\begin{minipage}[t]{0.498\textwidth}
\centering
\caption{
\textbf{Expanded 5s single-prompt comparison on VBench} using extended prompts~\cite{huang2026self}.
\#Params lists planner+renderer sizes; throughput is measured on H800 GPUs with idealized pipeline overlap.
Best and second-best evaluation scores are \textbf{bolded} and \uline{underlined}.
}
\label{tab:supp_comp_short}
\setlength{\tabcolsep}{3pt}
\resizebox{\textwidth}{!}{%
\begin{tabular}{@{}lcccccc@{}}
\toprule
\multirow{2}{*}{Model}
& \multirow{2}{*}{\#Params}
& \multirow{2}{*}{Resolution}
& \multirow{2}{*}{\shortstack{Throughput\\(FPS) $\uparrow$}}
& \multicolumn{3}{c}{Evaluation Scores $\uparrow$} \\
\cmidrule(lr){5-7}
& & & & Total & Quality & Semantic \\
\midrule

\rowcolor{GroupGray}
\multicolumn{7}{@{}l}{\textit{Non-streaming Video Models}} \\

Wan2.1-1.3B
& 1.3B & $832\!\times\!480$
& 0.91 & 84.27 & 84.99 & 81.39 \\

Wan2.2-5B
& 5B & $1280\!\times\!704$
& 0.77 & 83.75 & 83.89 & 83.16 \\

Bernini
& 7B+14B & $848\!\times\!480$
& 0.34 & \uline{84.83} & \textbf{85.51} & 82.12 \\

\midrule
\rowcolor{GroupGray}
\multicolumn{7}{@{}l}{\textit{Streaming Autoregressive Models}} \\

Self-Forcing
& 1.3B & $832\!\times\!480$
& 18.0 & 84.13 & 84.81 & 81.41 \\

Rolling Forcing
& 1.3B & $832\!\times\!480$
& 16.8 & 83.29 & 83.93 & 80.76 \\

Deep Forcing
& 1.3B & $832\!\times\!480$
& 17.7 & 84.36 & 85.17 & 81.13 \\

Infinity-RoPE
& 1.3B & $832\!\times\!480$
& 26.3 & 83.49 & 84.41 & 79.82 \\

Helios-Distilled
& 14B & $640\!\times\!384$
& 16.5 & 83.54 & 84.57 & 79.42 \\

LongLive
& 1.3B & $832\!\times\!480$
& 20.0 & 83.30 & 83.87 & 81.02 \\

MemFlow
& 1.3B & $832\!\times\!480$
& 18.4 & 83.08 & 83.51 & 81.34 \\

Anchor Forcing
& 1.3B & $832\!\times\!480$
& 17.3 & 83.90 & 84.50 & 81.47 \\

IAMFlow
& 1.3B & $832\!\times\!480$
& 16.9 & 83.24 & 83.75 & 81.20 \\

\midrule
\rowcolor{OursBlue}
\textbf{Ours-1.3B}
& 7B+1.3B & $832\!\times\!480$
& 19.2 & 84.44 & 84.69 & \uline{83.46} \\

\rowcolor{OursBlue}
\textbf{Ours-5B}
& 7B+5B & $1280\!\times\!704$
& 19.1 & \textbf{84.90} & \uline{85.23} & \textbf{83.60} \\

\bottomrule
\end{tabular}%
}
\end{minipage}%
\renewcommand{\arraystretch}{1.20}
\hfill
\begin{minipage}[t]{0.485\textwidth}
\centering
\caption{
\textbf{Quantitative comparison of 30s single-prompt video generation on VBench-Long}
using official extended prompts~\cite{huang2026self}.
Best and second-best scores across all streaming models are \textbf{bolded} and \uline{underlined}.
}
\label{tab:supp_vbench_long}
\setlength{\tabcolsep}{2.8pt}
\resizebox{\textwidth}{!}{%
\begin{tabular}{@{}lcccccc@{}}
\toprule
Model
& \shortstack{\strut Total\\\strut Score $\uparrow$}
& \shortstack{\strut Quality\\\strut Score $\uparrow$}
& \shortstack{\strut Semantic\\\strut Score $\uparrow$}
& \shortstack{\strut Subject\\\strut Consist. $\uparrow$}
& \shortstack{\strut Background\\\strut Consist. $\uparrow$}
& \shortstack{\strut Overall\\\strut Consist. $\uparrow$} \\
\midrule

Self-Forcing
& 81.87 & 83.46 & 75.49
& 97.79 & 97.01 & 27.85 \\

Rolling Forcing
& 81.72 & 83.31 & 75.38
& 98.03 & 97.36 & 28.29 \\

Deep Forcing
& 81.97 & 83.72 & 74.95
& 98.06 & 96.94 & 28.12 \\

Infinity-RoPE
& 82.40 & 84.00 & 76.02
& 98.02 & 97.08 & 27.57 \\

Helios-Distilled
& 82.64 & 84.53 & 75.07
& 95.57 & 97.30 & 26.22 \\

LongLive
& 82.38 & 83.90 & 76.31
& 98.23 & 97.37 & 28.22 \\

MemFlow
& 82.74 & 84.36 & 76.28
& 98.15 & 97.00 & 28.38 \\

Anchor Forcing
& 83.40 & \uline{85.42} & 75.32
& 96.97 & 96.36 & 27.73 \\

IAMFlow
& 82.15 & 83.53 & 76.61
& \textbf{98.27} & 97.21 & 28.65 \\

\midrule

\rowcolor{OursBlue}
\textbf{Ours-1.3B}
& \uline{83.54} & 84.69 & \textbf{78.94}
& \uline{98.26} & \textbf{97.75}
& \uline{30.09} \\

\rowcolor{OursBlue}
\textbf{Ours-5B}
& \textbf{84.38} & \textbf{85.81} & \uline{78.67}
& 97.99 & \uline{97.52}
& \textbf{30.25} \\

\bottomrule
\end{tabular}%
}
\end{minipage}
\end{table}

%% file: figures/supp_data_classifier_prompt.tex
\vspace{4pt}
\noindent
\begin{tcolorbox}[
    enhanced,
    breakable,
    skin first=enhanced,
    skin middle=enhanced,
    skin last=enhanced,
    pad at break*=4pt,
    colback=black!2,
    colframe=black!55,
    colbacktitle=black!12,
    coltitle=black,
    fonttitle=\bfseries\small,
    fontupper=\footnotesize,
    boxrule=0.5pt,
    arc=2.5pt,
    left=6pt, right=6pt, top=5pt, bottom=5pt,
    title={Prompt Template for Single- vs.\ Multi-Event Video Classification (Qwen3.6-27B)}
]
\textbf{System Prompt:}
Watch the continuous cut-free video carefully and evaluate its temporal event structure. First answer two auxiliary diagnostic questions to distinguish sequential sub-steps of one routine from transitions across distinct events, and then determine the overall event structure label.
\vspace{1mm}

\textbf{Diagnostic Questions \& Decision Rules:}
\begin{itemize}
    \setlength{\itemsep}{0.5pt}\setlength{\parskip}{0pt}\setlength{\topsep}{1pt}
    \item \textbf{Semantic Event Distinctiveness:} Does the video contain two or more semantically distinct activities or goals, rather than multiple motion stages of a single coherent activity (\eg, picking up a cup, drinking, and setting it down constitutes one event)?
    \item \textbf{Event Boundary Clarity:} Is there an identifiable transition between activities, such as a clear change in behavioral goal, subject interaction, or scene engagement?
    \item \textbf{Overall Event Structure:} Synthesize the visual evidence and auxiliary judgments to classify the clip as \texttt{single\_event} or \texttt{multi\_event}, without reducing the final label to a mechanical boolean combination of the two questions.
\end{itemize}
\vspace{1mm}

\textbf{Output Format:}\\
\texttt{\{"semantic\_event\_distinctiveness": "Yes | No", "event\_boundary\_clarity": "Yes | No", "event\_structure": "single\_event | multi\_event"\}}
\end{tcolorbox}
\vspace{4pt}

%% file: figures/supp_data_annotation_prompts.tex
\vspace{4pt}
\noindent
\begin{tcolorbox}[
    enhanced,
    breakable,
    skin first=enhanced,
    skin middle=enhanced,
    skin last=enhanced,
    pad at break*=4pt,
    colback=black!2,
    colframe=black!55,
    colbacktitle=black!12,
    coltitle=black,
    fonttitle=\bfseries\small,
    fontupper=\footnotesize,
    boxrule=0.5pt,
    arc=2.5pt,
    left=6pt, right=6pt, top=5pt, bottom=5pt,
    title={(a) Prompt Template for Multi-Event Segmentation and Concise Instruction (Gemini 3.1 Pro)}
]
\textbf{System Prompt:}
Watch the continuous video carefully and segment it into contiguous, non-overlapping semantic events covering the full duration from beginning to end, pairing each event with a concise objective English description $T^{(m)}$.
\vspace{1mm}

\textbf{Guidelines:}
\begin{itemize}
    \setlength{\itemsep}{0.5pt}\setlength{\parskip}{0pt}\setlength{\topsep}{1pt}
    \item Prefer medium-grained semantic phases where the main activity focus, interaction pattern, or scene purpose clearly shifts or completes.
    \item Do not split tiny motion details, brief gestures, gaze shifts, or posture adjustments within an ongoing activity; merge short transitional beats into the neighboring main event.
    \item Output contiguous one-decimal timestamps ($t_0 = 0.0$, $t_{m-1} < t_m$), which are subsequently quantized to $0.5\text{s}$ intervals while enforcing a $2.0\text{s}$ minimum segment duration.
\end{itemize}
\vspace{1mm}

\textbf{Output Format:}\\
\texttt{[\{"event": "$T^{(1)}$", "start\_time": "$t_0$", "end\_time": "$t_1$"\}, \{"event": "$T^{(2)}$", "start\_time": "$t_1$", "end\_time": "$t_2$"\}, $\dots$]}
\end{tcolorbox}

\vspace{1mm}
\noindent
\begin{tcolorbox}[
    enhanced,
    breakable,
    skin first=enhanced,
    skin middle=enhanced,
    skin last=enhanced,
    pad at break*=4pt,
    colback=black!2,
    colframe=black!55,
    colbacktitle=black!12,
    coltitle=black,
    fonttitle=\bfseries\small,
    fontupper=\footnotesize,
    boxrule=0.5pt,
    arc=2.5pt,
    left=6pt, right=6pt, top=5pt, bottom=5pt,
    title={(b) Prompt Template for Initial and Chained Event-Level Annotation (Qwen3.6-27B)}
]
\textbf{Part 1 — Initial Event Annotation ($m=1$)}\\[0.5mm]
\textbf{Input:} \texttt{[Segment $1$ video]}\\
\textbf{Guidelines:}
\begin{itemize}
    \setlength{\itemsep}{0.5pt}\setlength{\parskip}{0pt}\setlength{\topsep}{1pt}
    \item \textbf{Structured Schema:} Generate a single-paragraph factual caption $\widetilde{T}^{(1)}$ grounded strictly in visible evidence, ordered as \texttt{[Overall]} $\rightarrow$ \texttt{[Subject $k$]} ($1$--$3$ primary salient subjects) $\rightarrow$ \texttt{[Environment]} $\rightarrow$ \texttt{[Camera]}, with one sentence per tag.
    \item \textbf{Subject Grounding:} Each \texttt{[Subject $k$]} must represent a single distinct person, animal, or salient object (mentioning minor auxiliary entities only in \texttt{[Overall]}). State a clear identity anchor (\eg, specific category, clothing, position; avoiding ambiguous pronouns) first, followed by the dominant action and interaction second.
    \item \textbf{Environment \& Camera:} Describe background layout, lighting, and atmosphere in \texttt{[Environment]}, and specify shot scale, angle, and explicit camera motion (\eg, static, handheld, tracking, panning, tilting, or zooming) in \texttt{[Camera]}.
\end{itemize}
\textbf{Output Format ($m=1$):} \texttt{[\{"caption": "$\widetilde{T}^{(1)}$"\}]}

\vspace{1.5mm}
{\color{black!25}\hrule height 0.4pt}
\vspace{1.5mm}

\textbf{Part 2 — Subsequent Chained Annotation ($m\ge2$)}\\[0.5mm]
\textbf{Input:} \texttt{PREVIOUS}: \texttt{[Segment $m-1$ video]} $\;\mid\;$ \texttt{CURRENT}: \texttt{[Segment $m$ video]} $\;\mid\;$ \texttt{PREVIOUS\_CAPTION}: $\widetilde{T}^{(m-1)}$ $\;\mid\;$ \texttt{NEXT\_SUBJECT\_ID}\\
\textbf{Guidelines:}
\begin{itemize}
    \setlength{\itemsep}{0.5pt}\setlength{\parskip}{0pt}\setlength{\topsep}{1pt}
    \item \textbf{Persistent Subject Registry:} Treat every inherited \texttt{[Subject $k$]} as a permanent physical identity ID across the video (up to $4$ active subject sections per segment). Never group multiple entities or renumber existing IDs; assign consecutive IDs from \texttt{NEXT\_SUBJECT\_ID} only when genuinely new salient subjects enter, and update \texttt{next\_subject\_id} accordingly.
    \item \textbf{Subject Presence, Exit, and Reappearance:} Describe subjects that remain visible or complete their exit during segment $m$. Omit subjects that already completed their exit in earlier segments and remain absent throughout segment $m$, without renumbering remaining IDs (\eg, \texttt{[Subject 1]} $\rightarrow$ \texttt{[Subject 3]}). Reappearing subjects retain their original IDs.
    \item \textbf{Context-Dependent Event Change $R^{(m)}$ (\texttt{change}):} Ordered as \texttt{[Subject $k$]} $\rightarrow$ \texttt{[Environment]} $\rightarrow$ \texttt{[Camera]} (one sentence per tag, without \texttt{[Overall]}). Capture before-to-after state transitions from the end of segment $m-1$ into segment $m$, or note ongoing continuity (\eg, ``continues to'', ``remains'') when unchanged.
    \item \textbf{Self-Contained Caption $\widetilde{T}^{(m)}$ (\texttt{caption}):} Ordered as \texttt{[Overall]} $\rightarrow$ visible \texttt{[Subject $k$]} $\rightarrow$ \texttt{[Environment]} $\rightarrow$ \texttt{[Camera]} (one sentence per tag; appearance anchor first, current action second). Inherit valid appearance and scene details from $\widetilde{T}^{(m-1)}$, but describe segment $m$ standalone without referencing prior segments or using continuity words such as ``still'', ``continues'', ``remains'', or ``as before''.
\end{itemize}
\textbf{Output Format ($m\ge2$):} \texttt{[\{"change": "$R^{(m)}$", "caption": "$\widetilde{T}^{(m)}$", "next\_subject\_id": <int>\}]}
\end{tcolorbox}
\vspace{4pt}

%% file: figures/supp_data_verify_and_example.tex
\vspace{4pt}
\noindent
\begin{tcolorbox}[
    enhanced,
    breakable,
    skin first=enhanced,
    skin middle=enhanced,
    skin last=enhanced,
    pad at break*=4pt,
    colback=black!2,
    colframe=black!55,
    colbacktitle=black!12,
    coltitle=black,
    fonttitle=\bfseries\small,
    fontupper=\footnotesize,
    boxrule=0.5pt,
    arc=2.5pt,
    left=6pt, right=6pt, top=5pt, bottom=5pt,
    title={(a) Prompt Template for Global Quality Verification (Gemini 3.1 Pro)}
]
\textbf{System Prompt:}
You are a conservative expert reviewer of structured multi-event video annotations. Given the complete video, fixed event boundaries, and the full annotation chain $\{\widetilde{T}^{(m)}, R^{(m)}\}_{m=1}^M$, audit cross-segment identity persistence and material factual accuracy.
\vspace{1mm}

\textbf{Sequential Review Rules:}
\begin{itemize}
    \setlength{\itemsep}{0.5pt}\setlength{\parskip}{0pt}\setlength{\topsep}{1pt}
    \item \textbf{Stage 1 (Persistent Identity Check):} Verify whether every \texttt{[Subject $k$]} refers to the exact same physical entity across all segments. Report \texttt{subject\_id\_remap} (with segment-wise ID mapping) if a reappearing subject receives a duplicate ID or two IDs swap, or \texttt{complex\_identity\_error} if multiple entities are conflated.
    \item \textbf{Stage 2 (Materiality-Gated Content Check):} When identities are valid, report \texttt{content\_findings} only when clear visual evidence reveals a material factual conflict in dominant appearance, primary action or transition, environment, camera motion, or absent-subject hallucination (including redundant post-exit mentions). Ignore close synonyms or minor stylistic differences, and output atomic section-level replacements or \texttt{complex\_segment\_rewrite} when manual rewriting is needed.
\end{itemize}
\vspace{1mm}

\textbf{Output Format:}\\
\texttt{\{"summary": "<factual audit summary>", "identity\_findings": [$\dots$], "content\_findings": [$\dots$]\}}
\end{tcolorbox}

\vspace{1mm}
\noindent
\begin{tcolorbox}[
    enhanced,
    breakable,
    skin first=enhanced,
    skin middle=enhanced,
    skin last=enhanced,
    pad at break*=4pt,
    colback=black!2,
    colframe=black!55,
    colbacktitle=black!12,
    coltitle=black,
    fonttitle=\bfseries\small,
    fontupper=\footnotesize,
    boxrule=0.5pt,
    arc=2.5pt,
    left=6pt, right=6pt, top=5pt, bottom=5pt,
    title={(b) Curated Multi-Event Annotation Example from \datasetname{} (4 Events, $28.0\text{s}$)}
]
\textbf{Event 1 ($0.0$--$7.5\text{s}$)} $\;\mid\;$ \textbf{Concise Instruction $T^{(1)}$:} ``A man drives a black car into a suburban intersection, stops, and opens the door.''\\
\textbf{Structured Caption $\widetilde{T}^{(1)}$:}
\texttt{[Overall]} A man in a black car drives into a suburban intersection and comes to a stop.
\texttt{[Subject 1]} A man wearing a dark jacket sits in the driver's seat of a black car, holding the steering wheel before opening the driver's door.
\texttt{[Environment]} A quiet residential neighborhood features a white house, bare trees, a red stop sign, and a red fire hydrant under overcast lighting.
\texttt{[Camera]} The camera pans right in a low-angle medium shot to follow the car as it enters and stops.
\vspace{1.2mm}

\textbf{Event 2 ($7.5$--$14.0\text{s}$)} $\;\mid\;$ \textbf{Concise Instruction $T^{(2)}$:} ``He steps out of the car, closes the door, and walks toward the back.''\\
\textbf{Structured Caption $\widetilde{T}^{(2)}$:}
\texttt{[Overall]} A man steps out of a parked black car and walks toward its rear in a suburban neighborhood.
\texttt{[Subject 1]} The man, wearing a dark jacket, exits the black car, shuts the driver's door, and walks toward the rear of the vehicle.
\texttt{[Environment]} A quiet residential neighborhood features a white house, bare trees, a red stop sign, and a red fire hydrant under overcast lighting.
\texttt{[Camera]} The camera pans right in a low-angle medium shot following the man along the side of the car.\\
\textbf{Event Change $R^{(2)}$:}
\texttt{[Subject 1]} The man in the dark jacket steps out from the open driver's door of the black car, closes the door, and walks toward the rear.
\texttt{[Environment]} The overcast residential neighborhood with the white house, bare trees, stop sign, and fire hydrant remains unchanged.
\texttt{[Camera]} The camera continues panning right in a low-angle medium shot to track the man's movement.
\vspace{1.2mm}

\textbf{Event 3 ($14.0$--$21.0\text{s}$)} $\;\mid\;$ \textbf{Concise Instruction $T^{(3)}$:} ``He opens the trunk and takes out a green duffel bag.''\\
\textbf{Structured Caption $\widetilde{T}^{(3)}$:}
\texttt{[Overall]} A man in a dark jacket retrieves a green duffel bag from the trunk of a black car.
\texttt{[Subject 1]} The man, wearing a dark jacket, stands at the rear of the black car, opens the trunk lid, and lifts out a green duffel bag.
\texttt{[Environment]} A quiet residential neighborhood features a white house, bare trees, and a red stop sign under overcast lighting.
\texttt{[Camera]} The camera pans slightly right and holds a low-angle medium shot framing the rear of the car.\\
\textbf{Event Change $R^{(3)}$:}
\texttt{[Subject 1]} Reaching the rear of the black car, the man in the dark jacket opens the trunk and lifts a green duffel bag out of the compartment.
\texttt{[Environment]} The overcast suburban street setting remains consistent.
\texttt{[Camera]} The camera pans right to center on the rear trunk and settles into a steady low-angle medium shot.
\vspace{1.2mm}

\textbf{Event 4 ($21.0$--$28.0\text{s}$)} $\;\mid\;$ \textbf{Concise Instruction $T^{(4)}$:} ``Carrying the bag on his back, he glances back and walks away.''\\
\textbf{Structured Caption $\widetilde{T}^{(4)}$:}
\texttt{[Overall]} A man carrying a green duffel bag glances over his shoulder and walks away from a parked black car.
\texttt{[Subject 1]} The man, wearing a dark jacket with a green duffel bag slung over his back, looks back over his shoulder with a startled expression before walking out of the frame.
\texttt{[Environment]} A quiet residential neighborhood features a white house, bare trees, a red stop sign, and a red fire hydrant under overcast lighting.
\texttt{[Camera]} The camera pans left to follow the walking man and then holds a static medium shot on the parked car.\\
\textbf{Event Change $R^{(4)}$:}
\texttt{[Subject 1]} Now carrying the green duffel bag on his back, the man in the dark jacket turns his head to look over his shoulder before walking away from the car and exiting left.
\texttt{[Environment]} The overcast suburban neighborhood setting remains unchanged.
\texttt{[Camera]} The camera transitions from framing the trunk to panning left with the departing man before holding static on the parked car.
\end{tcolorbox}
\vspace{4pt}